# GUIDE: computationally augmented consensus for trustworthy clinical practice guidelines

**Coauthors:** Yuanfan Lin[1†]; Jie Di[1†]; Xujin He[1†]; Zhenzhe Lin[1†]; Jialing Chen[1]; Ruiyang Li[1]; Changyu Zheng[2]; Yuanrui Shang[1]; Weiqin Peng[1]; Jingxi Qin[2]; Xiong Wang[2]; Dongxin Ye[2]; Yining Huang[1]; Guoqiang Li[1]; Andi Xu[1]; Sijia Yu[2]; Ruixin Wang[1]; Yahan Yang[1]; Xiaohang Wu[1]; Duoru Lin[1]; Jingjing Chen[1]; Zhenzhen Liu[1]; Wenben Chen[1*]; Haotian Lin[1,3,4,5*]

**Affiliations:**

[1]Zhongshan Ophthalmic Center, Sun Yat-sen University, WHO Collaborating Centre for Eye Care and Vision, State Key Laboratory of Ophthalmology, Guangdong Provincial Key Laboratory of Ophthalmology and Visual Science, Guangdong Provincial Clinical Research Center for Ocular Diseases, Guangzhou, China;

[2]Zhongshan School of Medicine, Sun Yat-sen University, Guangzhou, China;

[3]Hainan Eye Hospital and Key Laboratory of Ophthalmology, Zhongshan Ophthalmic Center, Sun Yat-sen University, Haikou, Hainan, China;

[4]Department of Genetics and Biomedical Informatics, Zhongshan School of Medicine, Institute for Frontier Interdisciplinary Research in Health Sciences and Technology, Sun Yat-sen University, Guangzhou, China;

[5]Lead contact.

†These authors contributed equally to this work.

***Corresponding authors:**

Professor Haotian Lin, M.D., Ph.D., Email: linht5@mail.sysu.edu.cn. Zhongshan Ophthalmic Center, Sun Yat-sen University, 7# Jinsui Road, Guangzhou, China, 510623. Telephone: +86-20-87330493. Fax: +86-20-87333271.

Associate Professor Wenben Chen, M.D., Ph.D., Email: chenwb56@mail.sysu.edu.cn. Zhongshan Ophthalmic Center, Sun Yat-sen University, 7# Jinsui Road, Guangzhou, China, 510623. Telephone: +86-20-66648760. Fax: +86-20-87333271.

## Abstract

Maintaining up-to-date clinical practice guidelines (CPGs) is essential for evidence-based medicine (EBM), yet the current process is constrained by the rapid expansion of evidence and the complexity of expert consensus formation. To address this challenge, we developed GUIDE, a multi-agent copilot system that enables computationally augmented consensus for human-AI collaborative clinical guideline generation and updating, assigning labor-intensive tasks to specialized AI agents while preserving critical judgment and final arbitration for human experts. In a benchmark comprising 12 datasets derived from NICE ophthalmology-specific guidelines, GUIDE showed strong performance across evidence retrieval, assessment, and synthesis, achieving 88.85% literature recall, 86.06% alignment in certainty-of-evidence ratings, and 91.67% agreement on recommendations, only requiring 162.17 minutes per dataset on average (including 28.68% human intervention). Performance remained stable in additional 20 datasets spanning cardiology, endocrinology, gastroenterology, and respiratory medicine, indicating its cross-specialty generalizability. We then conducted an ablation analysis to validate the importance of human-AI collaboration mechanism, as removing expert intervention reduced performance in judgment-intensive steps such as Risk-of-Bias assessment. Furthermore, a simulated end-to-end evaluation using six post-cutoff datasets demonstrated that GUIDE maintained robust performance, supporting its ability to generalize to genuinely unseen evidence within a continuous guideline-updating workflow. Finally, prospective real-world validation in a live guideline updating process showed that GUIDE substantially alleviated expert workload across evidence selection, extraction, and synthesis. The GUIDE-assisted update was judged more clinically acceptable than the original guideline recommendation issued a decade earlier (mean score, 3.73 vs. 1.53; agreement, 66.7% vs. 13.3%) and achieved outcomes comparable to those of human-only updating (P = 0.588), while requiring substantially less time and expert effort. Together, our findings support GUIDE as a practical framework for timely, trustworthy, and expert-centric clinical guideline generation, and validate computationally augmented consensus as a feasible paradigm for enhanced guideline development.

## Introduction

Clinical practice guidelines (CPGs) constitute central instruments of evidence-based medicine (EBM), designed to integrate the best available evidence with clinical expertise to support informed healthcare decision-making [1]. By translating rapidly evolving knowledge into structured evidence bodies and actionable practice recommendations, CPGs play a critical role in reducing unwarranted practice variation, improving care delivery quality, and enhancing the overall efficiency and equity of healthcare systems [2]. Their value, however, depends on their timeliness. Recommendations that are not updated in line with emerging evidence can weaken their clinical validity and potentially lead to suboptimal care [3]. Conventional guideline workflows are increasingly unable to keep pace with the exponential growth of biomedical knowledge [4]. With typical update cycles exceeding two years [5,6], there is an inherent delay in translating new evidence into practice recommendations. This strain is exacerbated by the annual addition of over one million citations and a growing burden of secondary research [7], the scale and complexity of which far exceed current capacities for efficient evidence synthesis.

Yet the difficulty of guideline generation or updating extends beyond scale alone. More importantly, it is a consensus-building process that requires experts to collectively interpret uncertain evidence, weigh methodological limitations, balance benefits and harms, and reconcile clinical, economic, and societal considerations into accountable consensus. For each clinical question, experts must determine study eligibility, interpret heterogeneous findings and conclusions to estimate certainty of evidence, and finally translate evidence into recommendations in broad medical and socioeconomic context [8]. Many of these steps cannot be fully delineated by deterministic rules but depend on interpretation under uncertainty, contextual trade-offs and deliberation among domain experts. These tasks are not merely technical operations, but judgment-intensive processes that rely on time-consuming expert deliberation and consensus formation to ensure the trustworthiness of clinical guidelines. Consequently, continued reliance on traditional updating workflows may widen the gap between the rapid growth of new evidence and the responsiveness of existing guideline systems, ultimately limiting their effectiveness [9].

Recent studies highlight the potential of large language models (LLMs) to streamline literature screening and evidence extraction in medical guideline development [10,11], with systems like Quicker extending automation to end-to-end recommendation generation [12]. Despite these technical strides, real-world clinical integration remains challenging. Crucial elements such as transparent evidence scrutiny, the incorporation of expert clinical judgment, and accountable consensus formation under uncertainty are still inadequately addressed [13]. As probabilistic models, LLMs remain unreliable in tasks requiring implicit inference and quantitative reasoning. For example, they may fail to detect methodological bias indicated by missing information, extract numerical outcomes dispersed across a report, or assess the overall credibility of a trial [14,15]. Human experts are better equipped to evaluate methodological nuances and implementation trade-offs, but their capacity is increasingly strained by the rapidly expanding evidence base. This creates an urgent need for expert-facing AI systems that enable real-time human-AI collaboration [16]. Such systems should support computationally augmented consensus: LLMs enhance the speed, breadth, and consistency of evidence processing, while experts retain responsibility for resolving uncertainty and arbitrating high-stakes decisions, consistent with the emerging paradigm of AI-integrated medicine [17]. This also necessitates a well-defined human-AI collaborative framework that clarifies what tasks AI can perform, when expert judgment is required, and how responsibility should be allocated across both intermediate steps and final recommendations [18].

Here, we introduce GUIDE (**G**uideline **U**pdate **I**ntelligence **D**ecision **E**ngine), an LLM-based copilot framework for collaborative clinical guideline generation, built upon a human-AI partnership model derived from existing research and Delphi-like experts panel consultations [19]. The design strategically allocates roles to optimize the strengths of both AI and human experts throughout the guideline development process. While AI agents execute high-volume, rule-based tasks such as clinical question decomposition, literature screening, data extraction, and evidence synthesis, human experts maintain ultimate accountability, overseeing the process and contributing medical insights to ensure methodological rigor. This human-AI collaboration is implemented in two key modes: experts interact reactively, addressing flagged issues raised by AI reviewers, and proactively, using their professional judgment to fine-tune system outputs. To evaluate GUIDE, we first established a benchmark comprising 12 datasets derived from NICE (National Institute for Health and Care Excellence) ophthalmology-specific guidelines, confirming the system's foundational ability to perform key tasks in guideline generation. We then conducted an ablation study to quantify the impact of human involvement, removing reviewer mechanisms and expert participation, revealing the critical role of human judgment in enhancing system performance. To assess the framework's generalizability, we extended testing across multiple medical specialties and clinical question complexities on 20 datasets, demonstrating its broad applicability. Finally, we conducted two critical experiments to assess GUIDE's real-world applicability. We performed temporal hold-out simulations, applying GUIDE to six datasets released after the LLM's knowledge cutoff, to test the system's ability to integrate new evidence into an end-to-end guideline development workflow in real time. Furthermore, in a prospective real-world validation, we explored how GUIDE integrates emerging evidence into an existing guideline on intraocular lenses (IOLs) selection for cataract surgery patients. Our findings demonstrate that trustworthy guideline generation can be facilitated through GUIDE, a framework for computationally augmented consensus that systematically couples AI-mediated evidence processing with expert deliberation, thereby preserving expert oversight, methodological rigor, and clinical accountability while substantially reducing human workload.

## Results

### Development of the GUIDE for clinical guideline generation

GUIDE is a multi-agent copilot system to facilitate clinical guideline generation by integrating automated efficiency with clinical rigor through a structured human-AI partnership (Supplementary Video 1). The foundational logic of the system is anchored in a set of collaboration principles established via a Delphi-like consensus among a multidisciplinary expert panel. Centered on a "human-in-the-loop" architecture, GUIDE delegates high-volume, rule-based tasks to AI agents while ensuring that critical professional judgment and final validation remain firmly under human oversight. This collaborative design is implemented by experts via two modes: a proactive intervention approach to ensure clinical accountability in critical junctures and a condition-triggered review mechanism for monitoring automated tasks at predefined checkpoints (Supplementary Table 1). Throughout the guideline generation workflow, this allocation is tailored to the demand for professional expertise and risk profiles of each specific step. While AI agents primarily support relatively structured and standardized processes such as search strategy formulation and data extraction, human experts lead the interpretive and value-sensitive stages: activities including PICO (Population, Intervention, Comparison, and Outcome) construction, evidence quality assessment, and the final synthesis of recommendations are reserved for proactive human intervention to ensure clinical rigor and consensus (Supplementary Table 2).

To translate these principles into executable system logic, the system architecture incorporates specialized agents that operationalize the human-AI interface. Specifically, the proactive intervention approach is coordinated by an assistant agent that parses natural-language expert instructions into executable commands for specialized agents. The condition-triggered review mechanism is implemented through a rule-based reviewer agent that monitors system outputs to trigger reactive human intervention when suspected anomalies arise. By integrating these agent-driven functions, GUIDE establishes a dynamic division of labor that balances computational power with the rigorous accountability required for real-world evidence synthesis.

The framework operates through a seven-step sequential workflow spanning from question decomposition to clinical recommendation generation (Fig. 1a, Supplementary Video 1). The operation process begins with PICO construction, where GUIDE parses the clinical question and generates a structured PICO draft that can be directly used for evidence mapping. Given that the decomposed PICO factors determine the recommendation's clinical scope, experts must verify them and make necessary amendments at this step. Based on the finalized PICO, GUIDE generates a structured review protocol detailing the review type, eligibility criteria and synthesis framework for expert validation and adjustment before retrieval (Supplementary Video 1). Following protocol confirmation, two search agents driven by different LLMs formulate search strategies and perform parallel retrieval independently, while the AI reviewer monitors signals of potentially unreliable retrieval, such as marked discrepancies in literature volume or limited overlap between two search outcomes and elicits human intervention. The assistant agent facilitates experts in revising the strategy by translating natural-language instructions into search strings, obviating the need to directly modify complex syntax (Supplementary Fig. 1). Subsequently, two screening agents conduct title-and-abstract and full-text eligibility assessments sequentially, presenting rationales and confidence levels for each inclusion or exclusion decision (Supplementary Fig. 2). At the full-text screening stage, experts may be brought in when the AI reviewer detects any disagreement in two parallel rounds of screening and judgment. For included studies, two data extraction agents extract textual and numerical data and generate formats corresponding to the PICO framework (Supplementary Fig. 3). As previously described, any discrepancy in two parallel extractions identified by AI Reviewer is referred to experts for verification against the original text. Comparable outcomes from verified study-level data were merged into evidence bodies for synthesis and certainty assessment, with pooled effect sizes calculated by the agent where appropriate. The quality assessment agents then assess these evidence bodies using the widely adopted Grading of Recommendations Assessment, Development and Evaluation (GRADE) methods, and experts audit agents' evaluations where divergent grading arises. Experts can either convey their insights via the assistant agent or directly override automated quality assessment results. Finally, the evidence synthesis and recommendation module, implemented as a Delphi-like multi-agent module, integrates the processed evidence into Evidence-to-Recommendation (EtR) materials and draft recommendation statements, which are then submitted to the expert panel for final deliberation and consensus decision-making (Supplementary Fig. 4). As a copilot system that extends expert capabilities, GUIDE synergizes automated efficiency with stringent clinical governance (Fig. 1b). Its real-time interactive interface allows experts to seamlessly integrate their insights into the workflow, enhancing both the precision and the clinical accountability of the final outputs.

**GUIDE's benchmark performance**

To evaluate the baseline performance of GUIDE, we conducted a benchmark test using 12 specialized datasets derived from four ophthalmology guidelines published by NICE. NICE was selected as its guidelines development process is highly transparent and traceable. Each of these 12 datasets encompasses the complete evidence-based workflow, spanning from initial clinical question formulation to the development of final recommendations (Supplementary Table 3). To isolate module performance and mitigate error propagation, we tested the framework using a modular approach, in which each step was independently conducted and evaluated using original NICE workflow materials (Supplementary Table 4 and 5).

The collaborative workflow for each dataset involved a senior specialist with over ten years of clinical experience in the respective domain. Under this configuration, GUIDE performed effectively, enhancing efficiency across the entire process. In the PICO construction phase, the agent near-instantaneously generated structured drafts in a mean of 0.37 minutes per question. By having expert verifying its interpretation, the system achieved a mean semantic similarity score of 0.83 (ranging from 0.80 to 0.86). Across 12 benchmark tasks, the retrieval module identified 25,744 records in total, which were reduced to 1,369 after title-and-abstract screening and 135 final inclusions after full-text screening. High recall was maintained throughout the pipeline, with mean recall values of 98.81% (95% Confidence Interval [CI], 97.42%–100.00%), 93.57% (95% CI, 90.07%–96.96%), and 88.85% (95% CI, 85.22%–92.91%) for retrieval, title-and-abstract screening, and full-text screening, respectively. Precision improved progressively from 0.47% (95% CI, 0.33%-0.64%) at retrieval to 8.11% (95% CI, 6.02%–10.70%) after title-and-abstract screening and 84.60% (95% CI, 78.28%–90.61%) after full-text screening (Fig. 2a). Specificity and negative predictive value (NPV) were 98.56% (95% CI, 98.09%–99.05%) and 98.81% (95% CI, 98.27%–99.32%), respectively. In retrieval stage, the AI reviewer correctly triggered human intervention in 4 of 12 datasets when retrieval count discrepancies were detected. At the full-text screening stage, discrepant screening decisions were identified for 22 articles (accounting for 16.30% of all full-text articles screened), all of which were referred to experts for arbitration. Among them, 21 studies were confirmed to have inclusion/exclusion decisions consistent with the original guidelines. The literature search step required an average of 9.26 minutes per dataset in total, comprising 7.95 minutes of automated system processing and 1.31 minutes of human intervention, whereas the literature screening step required 76.20 minutes per dataset in total, comprising 60.40 minutes of automated system processing and 15.80 minutes of human intervention.

The data extraction process demonstrated high efficiency and precision, requiring an average of 28.20 minutes per dataset, including 15.89 minutes of automated system processing and 12.31 minutes of human rechecking. Compared with the conventional approach, in which manual extraction of data from a single article typically requires more than 40 minutes [20], this workflow substantially improved extraction efficiency while maintaining high accuracy of 98.02%. Although the system handled the majority of the workload, 22.69% of the extraction tasks required expert verification, with numerical values accounting for 88.9% of the verified items and textual data accounting for the remaining 11.1%. Further analysis shows that the overall error rate of 1.98% was entirely due to omissions of numerical data. Because the reviewer agent was designed to flag only mismatches for expert correction, these omissions did not trigger alerts and therefore remained undetected. Subsequent assessment of the 300 profiles averaged 10.24 minutes per dataset, consisting of 5.14 minutes of automated processing and 5.10 minutes of expert intervention. The framework showed strong agreement with the reference standard, achieving 82.80% (95% CI, 79.57%–86.76%) for risk-of-bias assessment, 90.16% (95% CI, 85.48%–94.33%) for indirectness, 88.84% (95% CI, 83.47%–94.08%) for inconsistency, 92.24% (95% CI, 87.90%–96.35%) for imprecision, and 86.06% (95% CI, 81.38%–90.96%) for final certainty grading. (Fig. 2b). Finally, the Delphi-like multi-agent module integrated these processed findings into draft recommendation statements within 37.90 minutes per question, achieving a consistency rate of 91.67% with the recommendations reported in the original guidelines (Fig. 2c, Supplementary Table 6). The average time expenditure for both GUIDE and human experts within each step is illustrated in Fig. 2d. Overall, GUIDE accounted for 71.32% of the total time (115.66 minutes), while expert oversight and refinement only comprised the remaining 28.68% (46.51 minutes).

**Ablation analysis of expert and multi-agent contributions to performance**

To quantify performance gains attributable to expert involvement and the multi-agent system, we conducted an ablation analysis by excluding records from the AI reviewer, AI assistant and human expert in the benchmark test, retaining only outputs corresponding to a fully automated workflow. As the AI reviewer responsible for identifying discrepancies across parallel agents was not considered, we restricted the analysis to a single GPT-4.1-driven agent at each step to ensure workflow consistency and comparability. We observed that the automated system alone performed well on relatively straightforward tasks, achieving a PICO semantic similarity of 0.79 and a mean literature retrieval recall of 92.58% (Fig. 3a). Across the 12 tasks, the retrieval agent identified 29,121 records in total, 1,645 records after title-and-abstract screening and 155 final inclusions, reflecting relatively broad eligibility criteria in the absence of expert input. Notably, final-stage recall decreased from 88.85% (95% CI, 85.22%–92.91%) in the full configuration to 79.80% (95% CI, 74.15%–86.13%) in the ablation analysis (P = 0.008), while final-stage precision fell even more pronounced from 84.60% (95% CI, 78.28%–90.61%) to 66.09% (95% CI, 58.92%–73.61%) (P = 0.010) (Fig. 3b). Specificity (96.49% (95% CI, 95.36%–97.60%)) and NPV (98.24% (95% CI, 97.43%–98.97%)) were largely preserved at high levels. This sharp decline indicates that without monitoring provided by the AI reviewer and human expert, single automated agent struggles to formulate and navigate comprehensive inclusion/exclusion criteria, leading to an influx of confounding information, especially irrelevant articles, that may ultimately dilute the quality of evidence synthesis and recommendation value. Data extraction accuracy dropped from 98.02% to 85.30% (P = 0.046), while the omission rate rose from 1.98% to 14.7% (P = 0.027). During the evidence assessment phase, consistency in risk of bias judgments declined significantly from 82.80% (95% CI, 79.57%-86.76%) to 69.54% (95% CI, 61.15%-77.92%) (P = 0.002), with a corresponding decline in consistency for overall certainty-of-evidence ratings from 86.06% (95% CI, 81.38%-90.96%) to 68.78% (95% CI, 62.72%-74.83%) (P < 0.001) (Fig. 3c). Detailed erroneous analysis reveals that agents fell to detect latent methodological flows that required deductive reasoning to identify the absence of evidence, such as lack randomization details. In such cases, the agent frequently defaulted to a "low risk" judgment for allocation concealment based solely on the presence of the term "sealed," whereas experts correctly interpreted the absence of these details as indicative of potential bias. In the recommendation generation step, the proportion of drafted recommendation rated as "exactly consistent" with the reference standards also decreased from 83.33% to 50.00% (Fig. 3d, Supplementary Table 6). Taken together, these findings show that LLMs may execute many structured components of guideline updating efficiently, but the reviewer-guided human-AI collaborative mechanism is critical for maintaining clinically interpretable recommendation generation in the most judgment-intensive stages.

**Evaluation of GUIDE generalizability across clinical specialties**

We further evaluated the generalizability of GUIDE across clinical specialties by testing GUIDE on 20 datasets derived from NICE guidelines in cardiology, endocrinology, gastroenterology, and respiratory medicine, with five datasets from each specialty (Supplementary Table 7). A senior domain specialist with over ten years of clinical experience was recruited for each field to function as a collaborative partner. Across these specialties, GUIDE maintained robust performance despite differences in domain-specific terminology and clinical context. The system achieved an average PICO similarity of 0.85 and literature retrieval recall of 98.88% (95% CI, 97.71%–99.81%) (Fig. 4a). The system's high-throughput efficiency remained consistent during study selection, where agents processed a total of 38,317 records and 1,668 full-text literature. Despite the heterogeneity of cross-disciplinary literature, the collaborative framework sustained high performance, with a full-text screening recall of 88.83% (95% CI, 85.45%–92.26%) and precision of 80.11% (95% CI, 74.33%–85.90%) (Fig. 4a). The final screening yielded a specificity of 97.24% (95% CI, 96.27%–98.14%) and an NPV of 98.61% (95% CI, 98.14%–99.07%). This step involved 51.79 minutes of agent processing and only 10.41 minutes of human review. In data extraction from 177 included studies, the overall data extraction accuracy was 81.13%, with an error rate caused by omission of 18.87%. Textual data extraction reached an accuracy of 89.91%, whereas numerical data extraction achieved an accuracy of 78.81%, consistent with previous findings that numerical data handing remains a challenge for LLMs. For evidence certainty assessment, the system achieved 82.23% (95% CI, 79.09%-85.55%) agreement in overall quality with the reference standard, while requiring a mean processing time of 12.47 minutes per article (8.2 minutes of expert recheck) (Fig. 4b). Finally, the multi-agent module synthesized these findings and, through the Delphi-like consensus process, generated draft recommendation statements within 44.0 minutes per dataset, achieving 90.00% agreement in expert evaluations (Fig. 4c, Supplementary Table 6). These findings indicate that GUIDE generalizes well across multiple medical specialties and provides a broadly applicable framework for guideline updating. The time expenditure for both AI agents and human experts within each step is illustrated in Fig. 4d.

**End-to-end evaluation of GUIDE performance on post-cutoff datasets**

To eliminate data leakage risks and assess the holistic performance of the framework, we curated six datasets from NICE guidelines published between April 2025 to March 2026 (details in Methods), simulating an end-to-end prospective experiment within a continuous workflow (Supplementary Table 8). Unlike modular tests in the benchmark, this end-to-end evaluation required GUIDE to process raw outputs from previous components, a design wherein the framework must withstand real-world error propagation to demonstrate its true operational integrity.

The end-to-end pipeline achieved a PICO semantic similarity of 0.84 and maintained strong performance through literature screening, with a final screening recall of 87.75% (95% CI, 81.05%–94.44%), precision of 82.17% (95% CI, 73.49%–91.06%), specificity of 97.15% (95% CI, 95.95%–98.49% ) and NPV of 97.21% (95% CI, 94.63%–99.35%) (Fig. 5a). While minor shifts in the evidence pool limited direct comparison of intermediate metrics, such as evidence quality assessment with original NICE reference, the core clinical reasoning remained consistent. Subject-matter expert evaluation confirmed a complete (100.00%) directional agreement between GUIDE's outputs and recommendations in the original guidelines (Fig. 5b, Supplementary Table 6). By successfully interpreting unseen evidence to generate clinically sound recommendations, GUIDE proves it moves beyond reproducing embedded content, highlighting its potential as a resilient tool for the continuous evolution of clinical practices.

**Prospective case study of GUIDE for cataract guideline updating**

We conducted a prospective validation study of the update of the NICE 2017 guideline Cataracts in adults: management (NG77), using its original review question, "What is the optimal strategy to facilitate simultaneous distance and near vision following cataract surgery?", as a case study. Two senior cataract specialists, each with over twenty years of clinical expertise, participated in the entire guideline-updating process. For steps requiring human judgment or intervention, the two experts independently assessed the issue and resolved disagreements by consensus. A pre-designated expert then operated GUIDE interface, implementing only the agreed actions.

In defining the PICO framework, the "Population" and "Comparator" criteria remained largely identical to the original version. Notably, the AI system incorporated Extended Depth of Focus (EDOF) IOLs, a category of growing significance in recent clinical research, into the "Intervention" criteria. However, consistent with previous observations regarding the limitations of automated systems in capturing socioeconomic nuances, the system omitted resource utilization and costs from the "Outcomes" criteria, which were manually added by experts (Supplementary Information 1).

Initial search strategies generated by the system yielded an unmanageable 9,740 records, indicating an overly broad search scope. Upon experts' intervention to restrict the inclusion to Randomized Controlled Trials (RCTs) and systematic reviews (following the original guideline's methods), the strategies were refined and resulting in 1,111 records (1,107 after de-duplication). Following parallel title-and-abstract screening, 56 articles underwent full-text assessment, yielding 23 new inclusions published after the original guideline's cutoff date. Furthermore, 28 studies identified from systematic review searches were subjected to automated full-text screening, from which four additional studies were included. These processes culminated in a total of 27 eligible studies comprising the updated body of evidence (Supplementary Information 2 and 3). The AI reviewer identified only six (6/56, 11%) discrepancies during the second screening phase. These were resolved through expert review, which confirmed that all six should be excluded based on minor misalignments with the predefined PICO scope. Remarkably, filtering the 1,107 retrieved records into the final 27 inclusions studies took only 55 minutes, including 11 minutes of human adjudication.

Data extraction was complicated by the relative abundance and non-standardized outcome measures. For example, the outcome "visual function" exhibited substantial heterogeneity across studies, encompassing multiple assessment instruments, diverse testing conditions, and varied data formats. These included binary outcomes (e.g., spectacle independence), continuous scale scores (e.g., vision-related quality of life questionnaires), continuous measurements (e.g., contrast sensitivity), and performance-based metrics (e.g., reading-related tests). Despite significant variations in narrative phrasing, the data extraction agents robustly captured categorical data, including study design, intervention details, and complications. Conversely, while numerical data were often presented in more structured tables, the AI reviewer flagged several omissions and inconsistencies. For these anomalies, experts verified and corrected them against the original article. Ultimately, seven IOLs comparisons and 21 distinct outcome categories were successfully extracted into standardized data tables within 38 minutes, including 15 minutes of expert involvement. The agent automatically grouped study-level data into comparison- and outcome-specific evidence bodies, calculated pooled effect sizes using appropriate synthesis methods, and organized the results into structured outputs for downstream quality assessment.

For evidence quality assessment, the system performance revealed distinct patterns across different Risk-of-Bias domains. Randomization and blinding domains demonstrated high inter-rater consistency between the two assessment agents, with only a single disagreement identified (3.7% discrepancy rate). In contrast, selection and reporting biases were predominantly rated as "unclear" due to insufficient reporting in the primary studies; only two studies provided adequate detail to warrant a "low risk" rating (No discrepancy in these two domains). Attrition bias emerged as the most contentious domain, with seven inconsistencies (25.9%) flagged by AI reviewer requiring further adjudication. These discrepancies were resolved by experts viewing relevant excerpts from source articles. Following this rigorous reconciliation process, the overall risk of bias for all included study was appraised as serious. The system showed more robust performance applying GRADE criteria for intervention studies. Out of 291 total quality rating across evidence bodies, discrepancies occurred only in the indirectness domain (6 cases, 2.1%), and 100% agreement was observed for imprecision and inconsistency owing to more objective and quantitatively defined criteria. By contrast, these identified discrepancies in indirectness typically involved sophisticated PICO interpretations that demanded contextual understanding. All discrepancies were carefully reviewed by the experts, who provided the final adjudicated ratings (Supplementary Information 4). The automated assessment process was completed within 47 minutes, while expert took approximately 25 minutes to finish judgement.

Upon completion of the evidence appraisal, a virtual Delphi panel comprising an AI committee chair, two ophthalmic practice agents, a healthcare provider representative agent, a commissioner agent, and a virtual patient together with two human cataract specialists, deliberated on the findings from their respective perspectives to formulate structured recommendations. From PICO formulation to recommendation development, the entire process took 192 minutes, including 65 minutes of expert involvement. By constructing a comprehensive EtR content, the virtual panel integrated evidence quality and emerging trends to balance clinical efficacy with cost and safety trade-offs (Supplementary Information 5). This evidence review supported a transition from the original "do not offer" statement to a more supportive recommendation for multifocal and EDOF IOLs. Notably, the updated recommendation adopted "consider" rather than more definitive phrasing, reflecting a shift in recommendation strength. This indicates that while these newer designs show clear clinical promise, the final choice should remain highly individualized and dependent on the patient's specific values and preferences (Supplementary Information 6).

In parallel, the human-only group followed the same prespecified guideline-updating workflow, encompassing PICO refinement, protocol development, literature searching and screening, data extraction and evidence synthesis, evidence quality assessment, and completion of the EtR framework. As in the GUIDE-assisted group, the experts incorporated EDOF IOLs into the intervention framework on the basis of contemporary clinical practice and emerging evidence (Supplementary Information 7). The literature search retrieved a comparable 1,012 records. Title-and-abstract screening required a total of 33 hours, followed by 21 hours for full-text eligibility assessment. Of the retrieved records, 37 articles proceeded to full-text assessment, of which 25 ultimately met the eligibility criteria (Supplementary Information 8 and 9). Compared with the GUIDE-assisted group, fewer articles proceeded to full-text assessment, indicating greater screening specificity at this stage. Data extraction and quantitative synthesis was one of the most time-consuming components of the human-only workflow, requiring 76 hours to construct a structured, long-format data extraction table. The same seven IOL comparison categories and 20 distinct outcome categories were extracted, and pooled effect estimates were subsequently calculated where appropriate. Evidence quality assessment represented another labor-intensive step, requiring another eight hours for independent appraisal and consensus resolution of disagreements. The distribution of evidence ratings and the reasons for downgrading were broadly consistent with those observed in the GUIDE-assisted group (Supplementary Information 10). An eight-member guideline committee then reviewed the GRADE evidence profiles and relevant benefits, harms, resource use, costs factors to formulate the draft recommendation over three days (Supplementary Information 11). Although phrased slightly differently, the updated recommendation similarly shifted from a strong recommendation against multifocal IOLs to a more conditional recommendation to consider their use, consistent with the conclusion of the GUIDE-assisted group. Overall, updating this review question through the human-only workflow required 23 days from initiation to completion with a largely consistent practice recommendation (Supplementary Information 12).

After both group's completion of their updating process, three recommendations (the original 2017 NICE statement, the GUIDE-assist drafted version and human-only developed version) were presented to an external panel of 15 ophthalmic experts to evaluate its clinical acceptability.

Experts were blinded with recommendation origins and asked to rate their level of agreement using a five-point Likert scale ranging from "strongly disagree" to "strongly agree". Agreement was defined as a rating of either "agree" (4) or "strongly agree" (5) (Supplementary Information 13).

The mean scores were 1.53 ± 1.06 for the original 2017 NICE statement, 3.73 ± 0.96 for the GUIDE-assist version, and 3.87 ± 1.36 for the human-only version. Both the GUIDE-assist and human-only versions received significantly higher agreement scores than the original 2017 NICE statement (adjusted P=0.009 and P=0.015), whereas no significant difference was observed between the GUIDE-assist and human-only versions (adjusted P=0.588). Agreement was 13.3%, 66.7%, and 73.3% for the original 2017 NICE, GUIDE-assist, and human-only recommendations, respectively, with a significant overall difference across groups (Cochran's Q=11.23, P=0.0036), although pairwise comparisons did not remain significant after correction, potentially due to limited power from the small sample size.

## Discussion

In this study, we developed and validated GUIDE, a human-AI collaborative framework for clinical guideline generation in realistic settings. The GUIDE framework supports the end-to-end process from evidence identification to recommendation drafting by leveraging the complementary strengths of LLM agents and human expert through a stepwise workflow. The value of GUIDE stems not only from its functional automation but from its principled integration into the practical workflows of guideline synthesis. Recognizing specific demands of each phase, a tailored approach to human-AI collaboration was designed. In PICO construction, for instance, the priority is defining a clinically sharp scope and selecting outcomes that balance benefits with harms—a process where proactive human oversight is vital to ensure relevance and interpretability. Tasks like search formulation, screening, and data extraction are defined by the need for technical precision and high-volume processing. These are ideal for condition-triggered review, where LLM agents manage the heavy lifting of repetitive operations, while experts intervene only when the system flags uncertainty or inconsistency. Finally, higher-order synthesis and final deliberation demand holistic judgment such as integrating risk, cost, and patient values, that only experts can provide. This configuration ensures that the workflow achieves maximal operational efficiency without compromising the rigor of the clinical evidence synthesis. Across benchmark evaluation, ablation analysis, temporal hold-out testing, and prospective real-world validation, GUIDE proved to be a robust collaborative partner, assisting experts in constructing, appraising, and synthesizing evidence with high fidelity shielding experts from the burden of manual labor.

Previous studies have shown that LLMs can support individual components of the process, such as literature screening, evidence extraction, and quality assessment, and more recent efforts such as Quicker have begun to integrate these components into end-to-end workflows [21-25]. Yet, procedural continuity alone does not by itself fulfill the rigorous demands of formal guideline development [26-28]. The essential challenge extends beyond task execution to the synthesis of recommendations that possess both clinical interpretability and expert consensus [26]. GUIDE was designed to address this gap by deeply embedding expert oversight throughout the entire workflow. Through a reviewer-triggered and copilot-assisted collaboration mechanism, experts engage both reactively to resolve automated anomalies and proactively to provide guidance at critical decision junctures. This dual-mode integration preserves computational speed while ensuring that expert discretion remains central to guidelines' integrity. These features distinguish GUIDE from earlier systems and support its role as a practical framework for bringing LLM capabilities into real-world production.

Our findings demonstrate that the proposed design performs robustly in module testing, with its efficacy further validated through ablation studies. In the most labor-intensive stages of guideline generation such as study selection and data extraction, the system achieved significant time savings by automating preliminary tasks. Furthermore, for tasks demanding holistic judgment and nuanced interpretation, such as risk of bias assessment and the formulation of final recommendations, experts leveraged their clinical insights to instruct the AI agents or revise their generations, ensuring the outputs were finely calibrated with real-world considerations. These results position GUIDE as more than a technical tool for accelerating guideline production, instead representing a form of computationally augmented consensus, in which AI-mediated evidence processing is systematically coupled with expert deliberation. By redistributing labor across the workflow, GUIDE reduces the amount of expert time required and redirects scarce expert cognitive resources away from low-value repetitive tasks toward critical judgment, arbitration, and consensus-building. This reframing is particularly important in guideline development, where the ultimate value of efficiency lies in preserving expert attention for decisions that most directly shape clinical validity.

Importantly, we further demonstrate the practical feasibility of this paradigm through prospective validation in real-world guideline updating scenarios. Unlike retrospective benchmarks, this setting more closely reflects the complexity, variability, and uncertainty encountered in real guideline development. In this setting, the indispensable role of human experts was repeatedly reinforced. Beyond correcting anomalies and discrepancies arising from automated processes, experts remained essential for tasks requiring arbitration and contextual judgment, including defining the research scope, assessing evidence quality, interpreting outcomes, and considering implementation feasibility in light of both treatment effects and broader practical concerns. Such tasks require clinical experience, normative reasoning and the ability to balance competing considerations that remain beyond the reach of current LLM-based automation. At the same time, the copilot system substantially reduced the burden on experts by supporting high-volume tasks such as evidence identification, screening, extraction, and preliminary synthesis [29-31]. The human-only arm further highlighted the substantial time and workforce required for conventional guideline updating. Against this benchmark, the GUIDE-assisted workflow achieved comparable evidence synthesis and recommendation updating with markedly greater efficiency. The optimal model, therefore, is not full automation but expert-AI collaboration, in which LLMs provide scalability and analytical support, while experts ensure methodological rigor, clinical relevance, and contextual appropriateness [32-36].

Several limitations of this study warrant consideration. First, while our benchmark covers multiple specialties, its scale and diversity remain modest. Testing the framework against more complex or controversial clinical questions is essential to confirm its generalizability. Second, the current workflow still lacks granular automated processing in evidence synthesis. Specifically, it cannot yet autonomously extract primary studies from retrieved systematic reviews, leaving the final screening of these nested references to human experts. Third, the system focuses primarily on clinical outcomes and lacks integrated modules for economic evaluation and modeling. As a result, evidence and perspective concerning cost-effectiveness and resource allocation still requires manual synthesis by expert panels. Furthermore, certain high-level tasks such as risk-of-bias assessment and final recommendation formulation remain sensitive to the reasoning nuances of LLMs, occasionally showing variability compared to human consensus. Future iterations should incorporate more explicit domain-specific criteria to bridge this gap. Additionally, the reviewer agent's intervention logic currently relies on static, predefined triggers; transitioning to adaptive, learning-based strategies could more effectively balance operational efficiency with quality control. Finally, although prospectively validated, broader multi-center trials are necessary to evaluate the system's robustness and long-term clinical impact across diverse organizational contexts.

In conclusion, this study demonstrates that while LLMs alone are insufficient to fully automate clinical guideline development, their integration within a structured, human-supervised workflow can substantially enhance both efficiency and reliability. By shifting the focus from automation to collaboration, our work provides a practical and scalable framework for next-generation guideline development. More broadly, this paradigm may represent a generalizable approach for integrating AI into other high-stakes, knowledge-intensive processes in medicine, where both computational efficiency and expert judgment are indispensable.

## Methods

### Overall study design and datasets

To evaluate the performance of GUIDE in clinical guideline generation, we constructed datasets for different evaluation purposes from 13 publicly available clinical guidelines published by NICE between October 26, 2017, and March 31, 2026 (for details of the dataset contents, see Supplementary Table 5). NICE guidelines were selected because of their highly transparent documentation, which provide traceable search strategies, explicit inclusion and exclusion criteria, and detailed evidence profiles, with each dataset corresponding to the full set of original materials for a single review question. We first established 12 datasets derived from four ophthalmology guidelines covering different subspecialties for benchmark testing and ablation study (Supplementary Table 3). To assess cross-disciplinary generalizability, we further incorporated 20 datasets derived from 6 guidelines across four clinical specialties: cardiology, endocrinology, gastroenterology, and respiratory medicine (Supplementary Table 7). To rule out the possibility that the strong performance of GUIDE was driven by data leakage, we constructed six additional datasets based on four guidelines published after the training cut-off dates of the LLMs used in this study and conducted end-to-end simulated prospective evaluations (Supplementary Table 8). For each dataset, the corresponding original materials, including PICO elements, search strategies, study selection records, data extraction tables, evidence quality assessments, and final recommendations, were extracted as ground truth for modular evaluation (Supplementary Table 4). In addition, a real-world prospective study was designed by identifying a clinical question from a 2017 ophthalmology guideline (*NG77: Cataracts in adults: management*) in which the recommendation had been removed for further consideration, thereby enabling a practical evaluation of the system in a real guideline updating scenario.

**Development of human-AI collaboration principles**

To establish a framework for human-AI collaboration in medical guideline development, a multidisciplinary panel of 10 experts was convened (four ophthalmologists with expertise in evidence-based ophthalmic practice and clinical decision-making, three EBM methodologists specializing in systematic reviews, evidence appraisal, and guideline methodology, and three specialists with experience in medical AI, LLMs, and clinical decision-support systems). We balanced the panel settings across clinical, methodological, and technical domains to ensure both scientific rigor and operational feasibility. We employed a three-round modified Delphi-like process to define and refine human-AI collaboration principle [19]. In Round one, panelists independently reviewed the traditional medical guideline development workflow and identified initial areas of AI involvement regarding efficiency and quality. Round two focused on refining role boundaries by evaluating anonymized feedback from the previous stage, with specific emphasis on accountability, and human intervention points. Round three involved a focused review to resolve remaining contested items and finalize the consensus-based suggestions. Throughout the process, response anonymization was strictly maintained to mitigate hierarchical bias and facilitate expert convergence. The resulting suggestions provided valuable guidance for the design of our system (Supplementary Table 1 and 2).

**Engineering design of GUIDE framework**

We developed a user-friendly framework for human-AI collaboration in clinical guideline generation (Supplementary Video 1). The framework comprises multiple LLM-based agents, each configured to perform a specific step in the guideline development workflow, an AI reviewer (ChatGPT-4.1) audits outputs and identifies when human intervention is required, an AI assistant (ChatGPT-4.1) enables experts to interact with the agents through natural-language instructions, and medical experts who contribute when expert judgment is required. For different task requirements, we customized standardized prompt templates for each agent, including role definitions, detailed task instructions, and specifications for output content and format. During prompt development, the templates were iteratively refined and debugged using five additional datasets (derived from *NG242: Diabetic retinopathy: management and monitoring*, *NG82: Age-related macular degeneration: diagnosis and management, NG98: Hearing loss in adults: assessment and management, NG257: Fertility problems: assessment and treatment and NG250: Pneumonia: diagnosis and management*) that were not included in the subsequent evaluations. These datasets were used for multiple rounds of development and small-scale testing to ensure the reliability of task execution.

The framework was designed to follow the core workflow of guideline development, spanning evidence retrieval, evidence assessment, and final recommendation formulation. The process initiates with PICO construction, where a ChatGPT-4.1-driven agent decomposes each clinical question into structured PICO components. Upon expert validation of these components, the agent then generates a draft review protocol outlining key methodological parameters, such as study objectives, review type, eligibility criteria, and the planned evidence synthesis framework (Supplementary Video 1). This protocol undergoes further expert refinement prior to literature retrieval. To accommodate complex clinical questions, the system's interactive interface allows experts to manually append additional contextual elements or broader considerations.

From literature retrieval to evidence assessment, we adopt a dual-agent parallel design, in which two independent agents using different LLM models (ChatGPT-4.1 and DeepSeek-R1) operate concurrently, with human intervention triggered under predefined conditions. In the literature retrieval stage, two agents independently generate search strings based on the review protocol. These queries are used to search major biomedical literature databases (PubMed via NCBI E-utilities, and Scopus via the Elsevier API), for retrieving article metadata, including titles, abstracts, author lists, publication dates, and journal information. The AI reviewer judges the consistency by previewing and comparing the sets of articles fetched by the different search strings generated by two agents, and triggers expert intervention when substantial discrepancies are detected, defined as either a relative difference of more than 30% in the total number of retrieved records or insufficient overlap between the two retrieval sets (indicated by a Jaccard similarity coefficient of less than 0.5). When either condition is met, human experts are alerted and can either directly revise the search strategy or provide natural-language instructions for optimization. These instructions are then translated by the assistant agent into executable search syntax, allowing refinement of the retrieval strategy without requiring experts to manually manipulate complex query logic. To maximize recall, the results from both retrieval rounds are merged to form the candidate set for downstream screening. To balance recall and screening efficiency, study selection is conducted in two stages: title-and-abstract screening followed by full-text screening. In the first stage, the two independent agents evaluate the title and abstract of each retrieved record against inclusion and exclusion criteria derived from the PICO elements, and each article is classified as either Include or Exclude. The union of included records retained by the two agents is then carried forward to full-text screening. For full text acquisition, GUIDE integrates open access repositories, including Semantic Scholar, OpenAlex, and Crossref to locate freely available PDF links. For copyright-protected full texts obtained through authorized channels, GUIDE also supports manual PDF upload. GUIDE supports built-in optical character recognition (OCR) to extract machine readable text, eliminating manual transcription. Full-text eligibility assessment is likewise performed independently by the two agents and monitored by the AI Reviewer agent. When discordant decisions occur, the reviewer agent triggers expert intervention. Screening results are displayed in a tabular interface, allowing human experts to inspect the AI-generated decisions and rationales for each article and, when necessary, revise the classification. The final set of included studies constitutes the body of evidence.

The extracted text was independently analyzed by the dual-agents for data extraction, including study design, population characteristics, intervention and comparator details, sample size for each comparison group, outcome measures, and corresponding result values. These extracted fields from each article were organized into structured evidence tables. When discrepancies arose for the same data element, the AI reviewer initiated expert intervention, and the experts verified the original text and corrected the extracted information as needed. Under the GRADE framework, each clinical outcome corresponded to a body of evidence, and its quality assessment incorporated both effect estimates and certainty of evidence. To estimate effects, study-level data were automatically categorized into comparison- and outcome-level evidence bodies. Clinical experts evaluated the extracted data for conceptual and clinical homogeneity to determine whether quantitative synthesis was appropriate. Where necessary, they refined the automated collections by excluding or segregating studies that inappropriate for pooling. For each eligible evidence body, the system selected an appropriate pre-specified meta-analytic approach according to the outcome type, effect measure, review design, and completeness of the available numerical data. The data required to calculate pooled effect estimates were then transformed into structured tabular inputs, including study identifiers, intervention and comparator groups, outcome data, and relevant variance information where available. These inputs were passed to R scripts through a Python-based computational utility, which invoked the execution of the statistical analyses via system commands. After computation, the pipeline extracted the pooled effect estimates and associated statistical outputs and reformatted the results into structured evidence tables suitable for subsequent GRADE assessment. Certainty was evaluated across Risk of bias, Indirectness, Imprecision, and Inconsistency. Risk of bias was assessed at the study level for each outcome using methodological assessment tools matched to study design. For included studies, the corresponding instruments were applied, including the Cochrane Risk of Bias tool (RoB or RoB 2), ROBINS-I, and the QUADAS-2 checklist [37-40]. Indirectness was assessed according to the degree of alignment between the study population, intervention, comparator, and outcomes and the target clinical question defined by the PICO framework. Imprecision was evaluated primarily on the basis of the width of confidence intervals and sample size. Inconsistency was assessed at the body-of-evidence level by examining the variability of results across studies reporting the same outcome. These domain-specific assessments are then synthesized and assign an overall certainty rating for each outcome according to GRADE upgrading and downgrading rules. When the two parallel agents produced discordant certainty ratings, the AI reviewer triggered expert adjudication.

The final evidence synthesis was performed by a Delphi-like multi-agent module (all agents in this module were driven by ChatGPT-4.1 model) that integrated the accumulated evidence and quality assessment into comprehensive EtR materials and preliminary recommendation draft. To convene the virtual committee, human experts dynamically selected AI participants from eight predefined specialist roles: Clinical Practice Expert, Research Methodology Expert, Research Ethics and Quality Expert, Evidence Synthesis Expert, Health Economics Expert, Patient and Public Representative, Health Policy and Implementation Expert, and Biostatistician. Beyond these established roles, experts might introduce novel roles by providing brief professional backgrounds and role definition, these inputs were subsequently refined by the AI Committee Chair into optimized system prompts. Once the panel was constituted based on the specific decision context, the AI Committee Chair provided each panelist with the complete evidence base. Each AI panelist then generated an independent, role-specific assessment. Finally, the AI Committee Chair synthesized these evaluations, identified points of convergence and divergence, and determined whether the predefined consensus criteria had been met. Once this consensus criterion was met, the deliberation process was concluded. If consensus was not reached, the Chair generated targeted feedback for each panelist and initiated a subsequent round of deliberation. Throughout this process, human experts observed the discussion in real time, reviewed each panelist's decision and reasoning, and could provide steering instructions between rounds to ensure that the deliberation remained methodologically rigorous and clinically grounded. Notably, in our retrospective experimental settings, human experts did not participate in this final deliberation step to avoid potential bias from their prior knowledge of the original guideline recommendations.

**Implementation of modular test**

To assess the system's performance at each stage and mitigate error propagation, we adopted a modular testing approach. This approach was applied across the benchmark test, ablation study, and cross-specialty generalization test. In all three settings, individual components were evaluated in isolation by initializing intermediate results from the gold-standard guidelines, enabling a direct comparison between the system's outputs and the reference material. In the benchmark and cross-specialty generalization tests, each dataset was paired with a senior domain specialist who had over ten years of clinical experience in the relevant field and served as the human collaborative partner.

Specifically, clinical questions derived from guidelines were provided as inputs for PICO construction. The expert-verified PICO outputs were compared against the reference using Bidirectional Encoder Representations from Transformers Score (BERTScore) to evaluate semantic similarity. For literature retrieval, original PICO elements and eligibility criteria derived from protocol utilized as search inputs. To ensure a rigorous evaluation of the system's recall, the temporal upper bound of the literature retrieval period was manually set to the publication date of the source guideline. Following the search strategy reported in the source guideline, a two-stage screening process (title-and-abstract screening followed by full-text screening) was conducted. Screening performance was evaluated using precision (positive predictive value [PPV]), recall (sensitivity), negative predictive value (NPV), and specificity. These metrics were calculated using the following formulas:

$$Precision\left(PPV\right)=\frac{TP}{TP+FP},\quad Recall\,(Sensitivity)=\frac{TP}{TP+FN}$$

$$Negative\ predictive\ value\left(NPV\right)=\frac{TN}{TN+FN},\quad Specificity=\frac{TN}{TN+FP}$$

Here, TP denotes studies selected in agreement with the reference, FP denotes studies selected but not included in the reference, FN denotes studies included in the reference but not selected, and TN denotes studies excluded in agreement with the reference. For data extraction, full-text PDFs of the included studies and the corresponding PICO frameworks were provided. The accuracy of the extraction was validated against the data points specified in the GRADE tables. Quality assessment was performed based on the extracted outcomes and the effect sizes calculated for each PICO element. Following the GRADE methodology, risk of bias, inconsistency, indirectness, and imprecision were assessed individually to determine the overall certainty of the evidence. Finally, after consensus reached by Delphi-like deliberation, the synthesized recommendations were evaluated into three levels of consistency (exactly consistent, partially consistent, inconsistent) relative to the original guidelines (Supplementary Table 6).

**Implementation of end-to-end experiments using post-cutoff datasets**

Before applying GUIDE in real-world guideline updating practice, to rule out the potential risk that its strong performance was attributable to data leakage arising from prior LLMs exposure to guidelines content during training, we constructed datasets from guidelines published after the release dates of the LLMs used in GUIDE (ChatGPT-4.1 (April 14, 2025) and DeepSeek-R1 (January 20, 2025))and used them for an end-to-end simulated prospective evaluation. As previously described, each dataset was paired with a senior domain specialist who had over ten years of clinical experience in the corresponding field and served as the human collaborative partner. Unlike the modular evaluation where steps were initialized with gold-standard inputs, this simulation required each step to process the direct generations of its predecessor. The process commenced with the clinical questions as the sole input, from which the determined PICO frameworks were passed to the retrieval stage. The retrieval literature pool was then fed into a two-stage screening process. All subsequent stages, including data extraction and evidence quality assessment, were performed based exclusively on the studies selected in the preceding steps, and the final recommendations were generated relying entirely on the evidence synthesized within this specific run.

**Design and implementation of the prospective case study**

We prospectively evaluated the GUIDE human–AI collaborative framework in a real-world case study based on the NICE 2017 guideline *Cataracts in adults: management (NG77)*. By targeting the review question "What is the optimal strategy to facilitate simultaneous distance and near vision following cataract surgery?" -- a topic that had been officially prioritized for updating on the NICE website. This case study aimed to demonstrate the system's ability to integrate modern clinical data into a formal updating process. Two senior cataract specialists, each with over twenty years of clinical expertise, participated in the entire guideline-updating process. For steps requiring human judgment or intervention, the two experts independently assessed the issue and discussed any disagreements until a consensus decision was reached. One of the two experts was pre-designated before the experiment as the system operator and was responsible for all operations performed within GUIDE, implementing only the actions agreed upon by both experts.

In parallel with the GUIDE human-AI collaborative team, we convened an independent human-only guideline-updating committee. Its evidence review group comprised two additional senior cataract specialists (each with over twenty years of clinical experience) and one technical analyst. This group was responsible for evidence retrieval, quality assessment, and data processing, ultimately producing the bodies of evidence and their corresponding quality ratings. Throughout the process, members were prohibited from using GUIDE system. The core decision-making committee comprised one committee chair, four cataract specialists (including two experts participated in the preceding evidence synthesis steps), one healthcare provider representative, one commissioner, and one lay member. After reviewing the bodies of evidence, the committee used a modified three-round Delphi process to complete an EtR table and formulate a draft recommendation addressing the same review question.

To validate the clinical relevance of the recommendations generated through the two updating approaches, an external consensus process was initiated with a broader cohort of 15 senior physicians with over 10 years of clinical experience in cataract care. The objective of this survey was to evaluate the acceptability of three recommendation statements: the original 2017 recommendation, the recommendation drafted through the GUIDE human-AI collaborative process, and the recommendation developed through the human-only process. The external reviewers were not involved in any part of either updating process and were fully blinded to the source of the three recommendation statements. To minimize potential preferential evaluation based on presentation sequence, the three statements were presented in a randomized order across reviewers. Using a five-point Likert scale ranging from “strongly disagree” (1) to “strongly agree” (5), the external reviewers independently assessed each recommendation. These ratting were dichotomized as approval (a rating of 4 or 5) or non-approval (a rating of 1–3).

**Statistical analysis**

In the benchmark test, ablation analysis, generalization test and end-to-end experiments, system performance was evaluated against guideline-derived gold standards. For clinical question parsing, semantic similarity between generated and reference PICO elements was assessed using BERTScore. Retrieval strategy generation and literature screening were evaluated using precision (PPV), recall (sensitivity), NPV, and specificity. For evidence extraction and quality assessment, categorical outputs were evaluated using classification accuracy, whereas the accuracy of quantitative results was assessed by their deviation from reference values. Results are reported as means with 95% CIs, calculated using a non-parametric bootstrap approach with 1,000,000 resamples. Paired t-tests were used for approximately normally distributed differences, and Wilcoxon signed-rank tests were used otherwise. Approval for three recommendation statements was reported as the number and proportion of approving reviewers. Overall differences in approval were assessed using Cochran's Q test. If the omnibus test was significant, pairwise comparisons were performed using exact McNemar tests with Holm adjustment. All tests were two-sided, with $P < 0.05$ considered statistically significant. All statistical analyses were performed in R (v.4.2.3).

**Data availability**

The publicly available guideline documents used in this study include: Diabetic retinopathy: management and monitoring (NICE guideline Reference number:NG242), Cataracts in adults: management (NICE guideline Reference number:NG77), Age-related macular degeneration (NICE guideline Reference number:NG82), Glaucoma: diagnosis and management (NICE guideline Reference number:NG81), Chronic heart failure in adults (NICE guideline Reference number: NG106), Adrenal insufficiency: identification and management (NICE guideline Reference number: NG243), Cirrhosis in over 16s: assessment and management (NICE guideline Reference number: NG50), Barrett's oesophagus and stage 1 adenocarcinoma (NICE guideline Reference number: NG231), Pneumonia: diagnosis and management (NICE guideline Reference number: NG250), Asthma: diagnosis and monitoring of asthma in adults, children and young people (NICE guideline Reference number: NG80), Falls: assessment and prevention in older people and people 50 and over at higher risk (NICE guideline Reference number: NG249), Fertility problems: assessment and treatment (NICE guideline Reference number: NG257), Suspected sepsis in people aged 16 or over: recognition, assessment and early management (NICE guideline Reference number: NG253), Hearing loss in adults: assessment and management (NICE guideline Reference number: NG98). All of these guideline materials can be found in https://www.nice.org.uk/guidance.

**Code availability**

GUIDE is an LLM-based copilot system designed to support expert-facing clinical practice guideline generation and updating. We are not making the source code of GUIDE publicly available at this stage owing to the safety implications of unmonitored use of such a system in medical guideline workflows. Inappropriate deployment outside expert-supervised settings could lead to misuse in evidence appraisal, recommendation formulation, or clinical decision-support contexts beyond the validated scope of this study. In the interest of responsible innovation, we will continue to work with clinical experts, guideline methodologists, institutional partners, and relevant stakeholders to further validate GUIDE and explore safe onward uses. To facilitate peer review, we have provided the source code, testing dataset, and demonstration video as supplementary materials for the editors and expert reviewers in the submission system. The project overview and deployment instructions are available through GitHub (https://github.com/KimHe-RGB/GUIDE_INTRO). Additional details may be made available by the corresponding author upon reasonable request, subject to institutional policies and approval procedures.

**Acknowledgments**

We gratefully acknowledge the Guangdong Basic Research Center of Excellence for Major Blinding Eye Diseases Prevention and Treatment for its support of this study. We also sincerely thank all participants for their valuable contributions. We gratefully acknowledge Zhenjun, Tu, Yunyu Lian, Luyu Liu, Kechen Wang, Yulin Zheng, and Qingqing Zhao for their valuable feedback on the design and use of the GUIDE copilot system. This study was funded by the National Natural Science Foundation of China (92368205 and 82441003 to H.L.; 82401313 to R.L.), Guangdong Basic and Applied Basic Research Foundation (2024A1515012001 to W.C.; 2025A1515011616 to R.L.), China Postdoctoral Science Foundation (2023M744033 to R.L.), the China Postdoctoral Science Foundation Special Fund (R.L.), the Postdoctoral Fellowship Program of CPSF (GZB20230896 to R.L.), Guangdong Basic Research Center of Excellence for Major Blinding Eye Diseases Prevention and Treatment (W.C.), the Science and Technology Development Fund, Macau SAR (File No. 0010/2023/AKP to H.L.), Basic scientific research projects of Sun Yat-sen University (23ykcxqt002 to H.L.), the Science and Technology Planning Projects of Guangdong Province (2018B010109008 to H.L.), Hainan Province Clinical Medical Center (H.L.). “3510” Medical Innovation Leadership Program, Fundamental Research Funds for the Central Universities, Sun Yat-sen University (20263510YXON012 to W.C.).

**Author Contributions**

H.L. and W.C. contributed to the concept of the study and designed the research; Y.L, X.H., J.D., Z.L., J.C., R.L., C.Z., Y.S., W.P., J.Q., X.W., D.Y., G.L., A.X., and S.Y. collected the data; W.C., Y.L, X.H., J.D., Z.L., J.C., W.P., and G.L. conducted the study; Y.L, X.H., J.D., and J.Q. analyzed the data; W.C., Y.L, X.H., J.D., Z.L., and C.Z. drafted the manuscript; H.L., W.C., Y.Lin, X.H., J.D., Z.L., R.L., C.Z., Y.S., J.Q., and S.Y. critically revised the manuscript; H.L., X.H., Z.L., and W.P. performed the technical review; and all authors discussed the results and provided comments regarding the manuscript.

**Competing interests**

The authors declare no conflicts of interest.

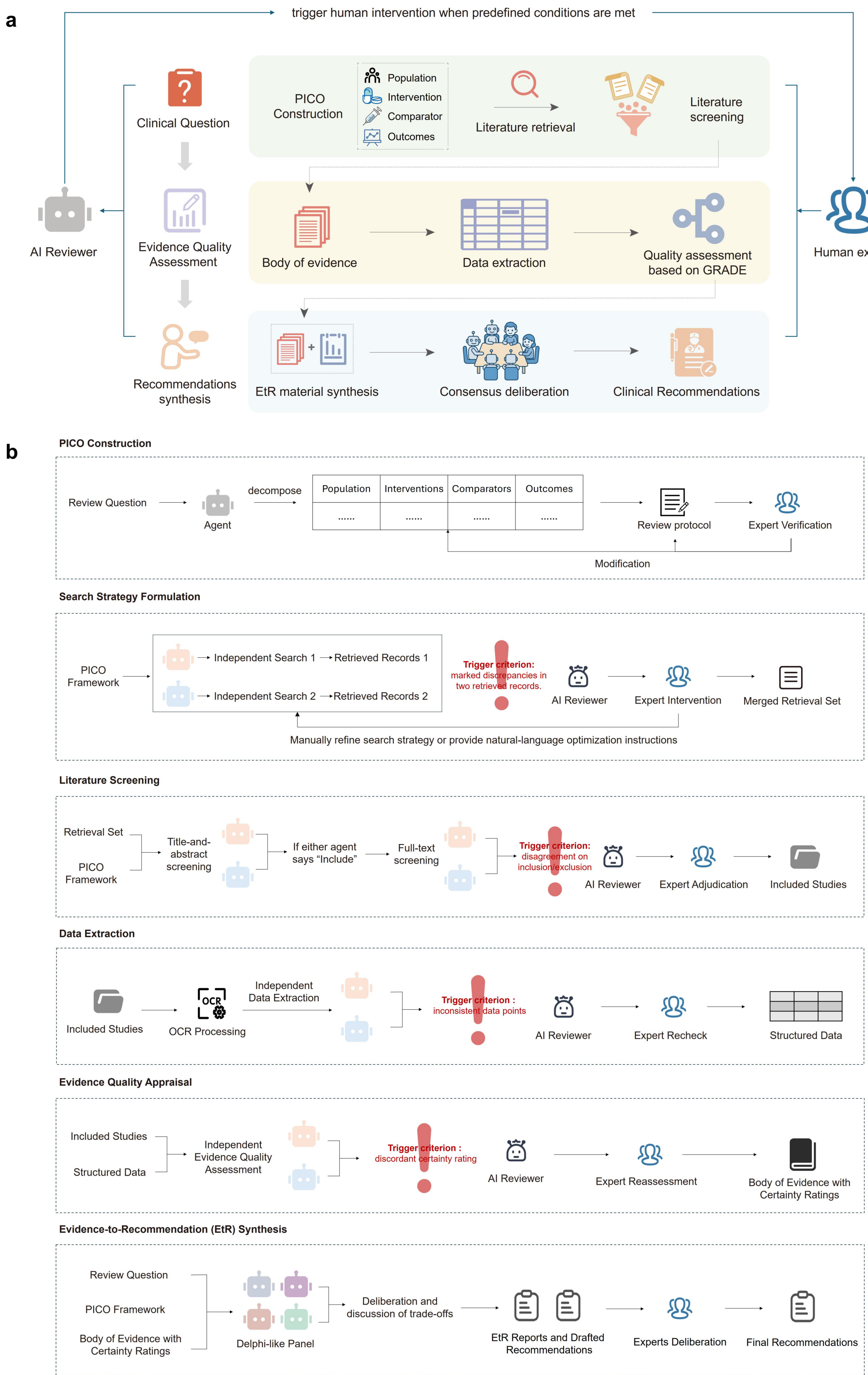
a
trigger human intervention when predefined conditions are met
Clinical Question
Evidence Quality Assessment
Recommendations synthesis
AI Reviewer
Human expert
PICO Construction
Population
Intervention
Comparator
Outcomes
Literature retrieval
Literature screening
Body of evidence
Data extraction
Quality assessment based on GRADE
EtR material synthesis
Consensus deliberation
Clinical Recommendations
b
PICO Construction
Review Question
Agent
decompose
Population
Interventions
Comparators
Outcomes
Review protocol
Expert Verification
Modification
Search Strategy Formulation
PICO Framework
Independent Search 1
Retrieved Records 1
Independent Search 2
Retrieved Records 2
Trigger criterion: marked discrepancies in two retrieved records.
AI Reviewer
Expert Intervention
Merged Retrieval Set
Manually refine search strategy or provide natural-language optimization instructions
Literature Screening
Retrieval Set
PICO Framework
Title-and-abstract screening
If either agent says "Include"
Full-text screening
Trigger criterion: disagreement on inclusion/exclusion
AI Reviewer
Expert Adjudication
Included Studies
Data Extraction
Included Studies
OCR
OCR Processing
Independent Data Extraction
Trigger criterion : inconsistent data points
AI Reviewer
Expert Recheck
Structured Data
Evidence Quality Appraisal
Included Studies
Structured Data
Independent Evidence Quality Assessment
Trigger criterion : discordant certainty rating
AI Reviewer
Expert Reassessment
Body of Evidence with Certainty Ratings
Evidence-to-Recommendation (EtR) Synthesis
Review Question
PICO Framework
Body of Evidence with Certainty Ratings
Delphi-like Panel
Deliberation and discussion of trade-offs
EtR Reports and Drafted Recommendations
Experts Deliberation
Final Recommendations

**Fig.1 | Clinical guideline updating workflow and human-GUIDE collaboration flowchart. a,** Overview of the GUIDE workflow under a human-in-the-loop copilot model. Starting from a clinical question, GUIDE supports PICO construction, literature retrieval and screening, evidence assembly, data extraction, GRADE-based quality assessment, evidence-to-recommendation material synthesis, multi-agent consensus, and generation of clinical recommendations. Throughout the process, AI agents perform structured evidence-synthesis tasks, whereas human experts retain supervisory authority and intervene when predefined conditions are met. **b,** Detailed workflow showing the trigger criterion for expert involvement. GUIDE decomposes the review question into PICO elements, coordinates independent agents for search, screening, data extraction and evidence quality appraisal, and synthesizes evidence-to-recommendation materials through a Delphi-like multi-agent process. At each key stage, an AI Reviewer monitors predefined signals of uncertainty or inconsistency, such as discordant retrieval results, screening disagreements, inconsistent extracted data or divergent certainty ratings, and triggers expert verification, adjudication or reassessment. This condition-triggered oversight enables GUIDE to combine automated efficiency with transparent and accountable human governance.

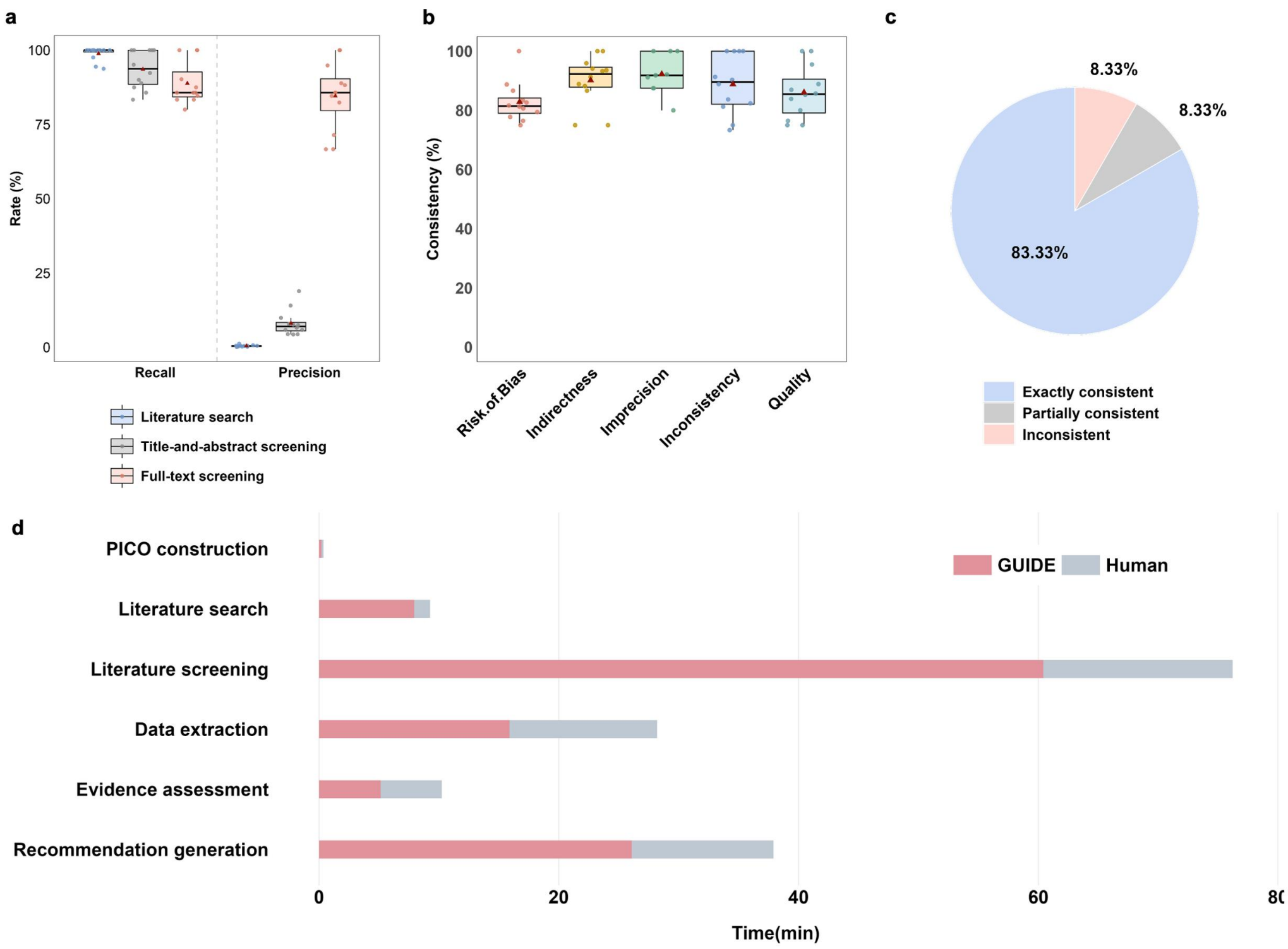

**Fig.2 | Performance of the human-GUIDE collaborative framework in the benchmark evaluation using 12 ophthalmology datasets. a,** Recall and precision of GUIDE across literature retrieval, title-and-abstract screening, and full-text screening. GUIDE achieved high recall across all stages, while precision progressively increased across the three steps. **b,** Agreement between GUIDE and human reference standards in evidence assessment, including risk of bias, indirectness, imprecision, inconsistency, and overall certainty grading. In each box plot, the triangle represents the mean, the central line denotes the median, the box indicates the interquartile range, and the whiskers extend to values within 1.5× the interquartile range. **c,** Consistency between GUIDE's recommendation outputs and original recommendations assessed by domain expert. **d,** Time expenditure of GUIDE copilot system and human experts across the guideline development workflow. GUIDE contributed the majority of processing time, while expert involvement was concentrated on oversight, arbitration, and refinement.

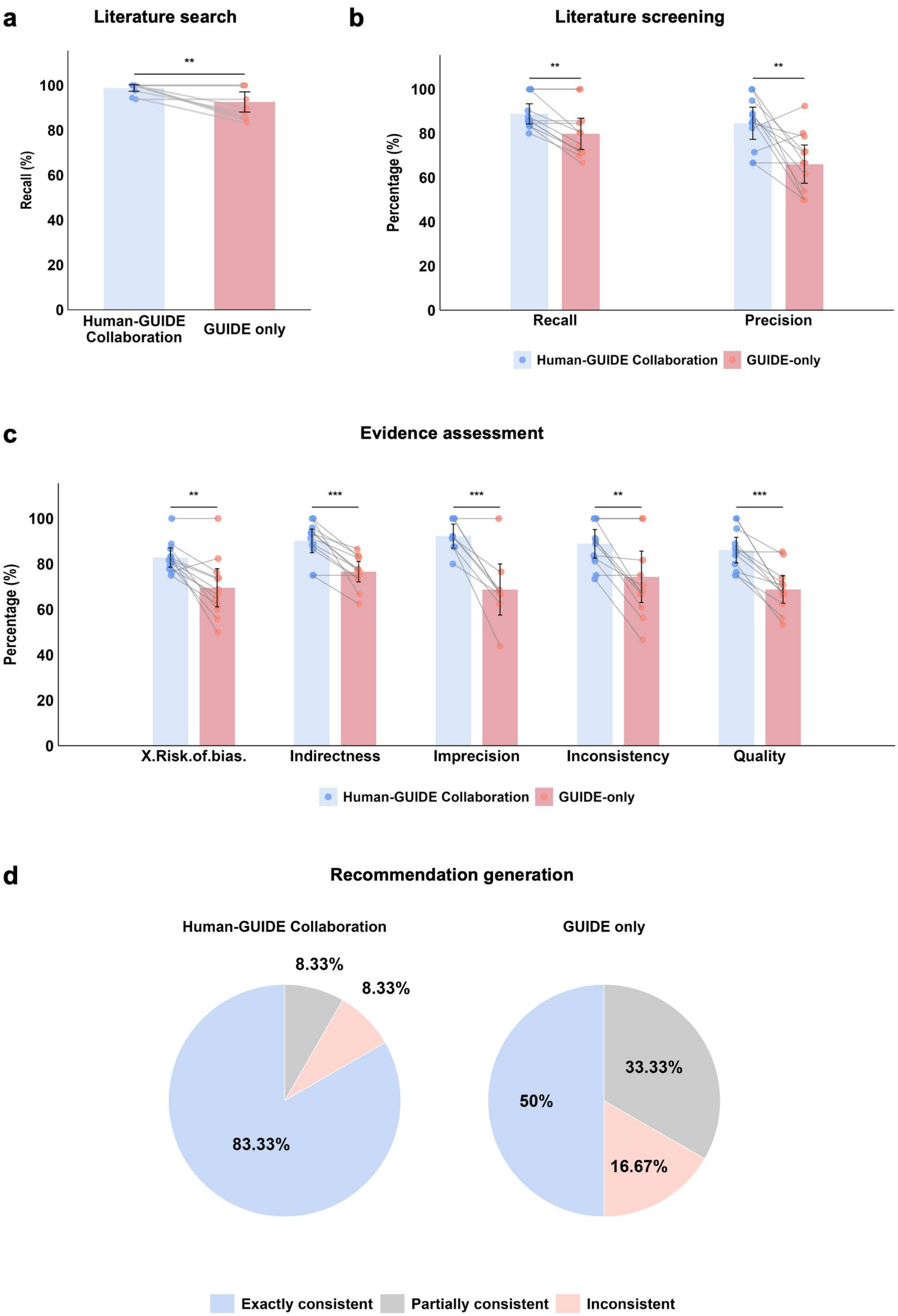
a
Literature search
**
Recall (%)
0
20
40
60
80
100
Human-GUIDE Collaboration
GUIDE only
b
Literature screening
**
**
Percentage (%)
0
20
40
60
80
100
Recall
Precision
Human-GUIDE Collaboration
GUIDE-only
c
Evidence assessment
**
***
***
**
***
Percentage (%)
0
20
40
60
80
100
X.Risk.of.bias.
Indirectness
Imprecision
Inconsistency
Quality
Human-GUIDE Collaboration
GUIDE-only
d
Recommendation generation
Human-GUIDE Collaboration
8.33%
8.33%
83.33%
GUIDE only
50%
33.33%
16.67%
Exactly consistent
Partially consistent
Inconsistent

**Fig. 3 | Ablation tests comparing human-GUIDE collaboration versus GUIDE alone: a** The recall rate of the literature search. **b** The recall and accuracy rates of literature screening. **c**–**d** The consistency between the GUIDE and the NICE guidelines for the evidence assessment (RoB, indirectness, imprecision, inconsistency, and total evidence quality evaluations), and recommendation generation modules (n = 12). The error bars represent the 95% confidence intervals (CIs). Statistical significance is indicated using paired comparisons, with ns denoting not significant, ** denoting $P < 0.01$ and *** denoting $P < 0.001$.

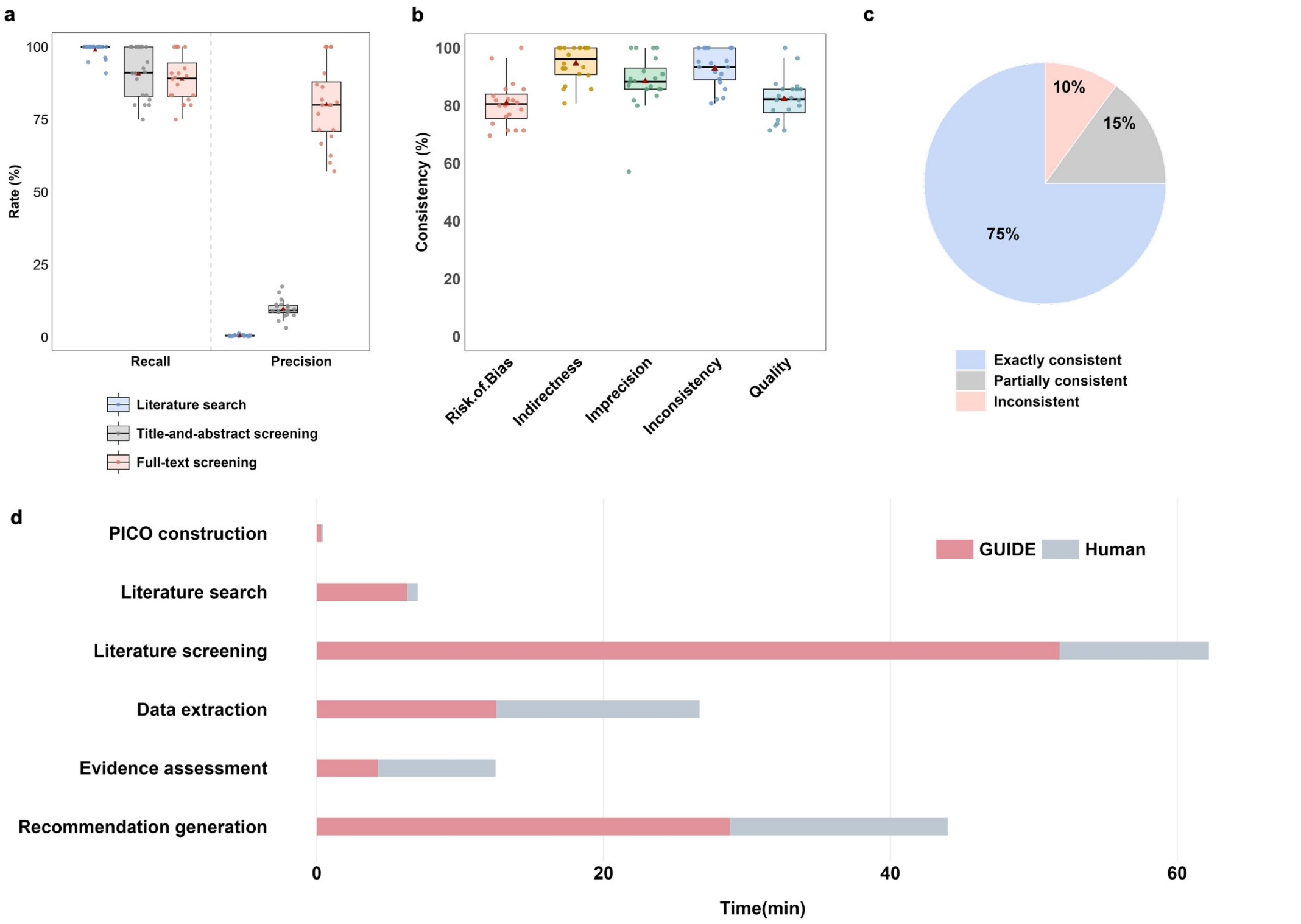

**Fig.4 | Performance of the human-GUIDE collaborative framework in the generalization test using 20 datasets derived from four different specialties (cardiology, endocrinology, gastroenterology, and respiratory medicine). a,** Recall and precision of GUIDE across literature retrieval, title-and-abstract screening, and full-text screening. **b,** Consistency of GUIDE-assisted evidence assessment across risk of bias, indirectness, imprecision, inconsistency, and evidence quality. **c,** Consistency between GUIDE's recommendation outputs and original recommendations assessed by domain expert. In each box plot, the triangle represents the mean, the central line denotes the median, the box indicates the interquartile range, and the whiskers extend to values within 1.5× the interquartile range. **d,** Time expenditure across the guideline-updating workflow, comparing GUIDE processing time with human expert involvement for PICO construction, literature search, literature screening, data extraction, evidence assessment, and recommendation generation.

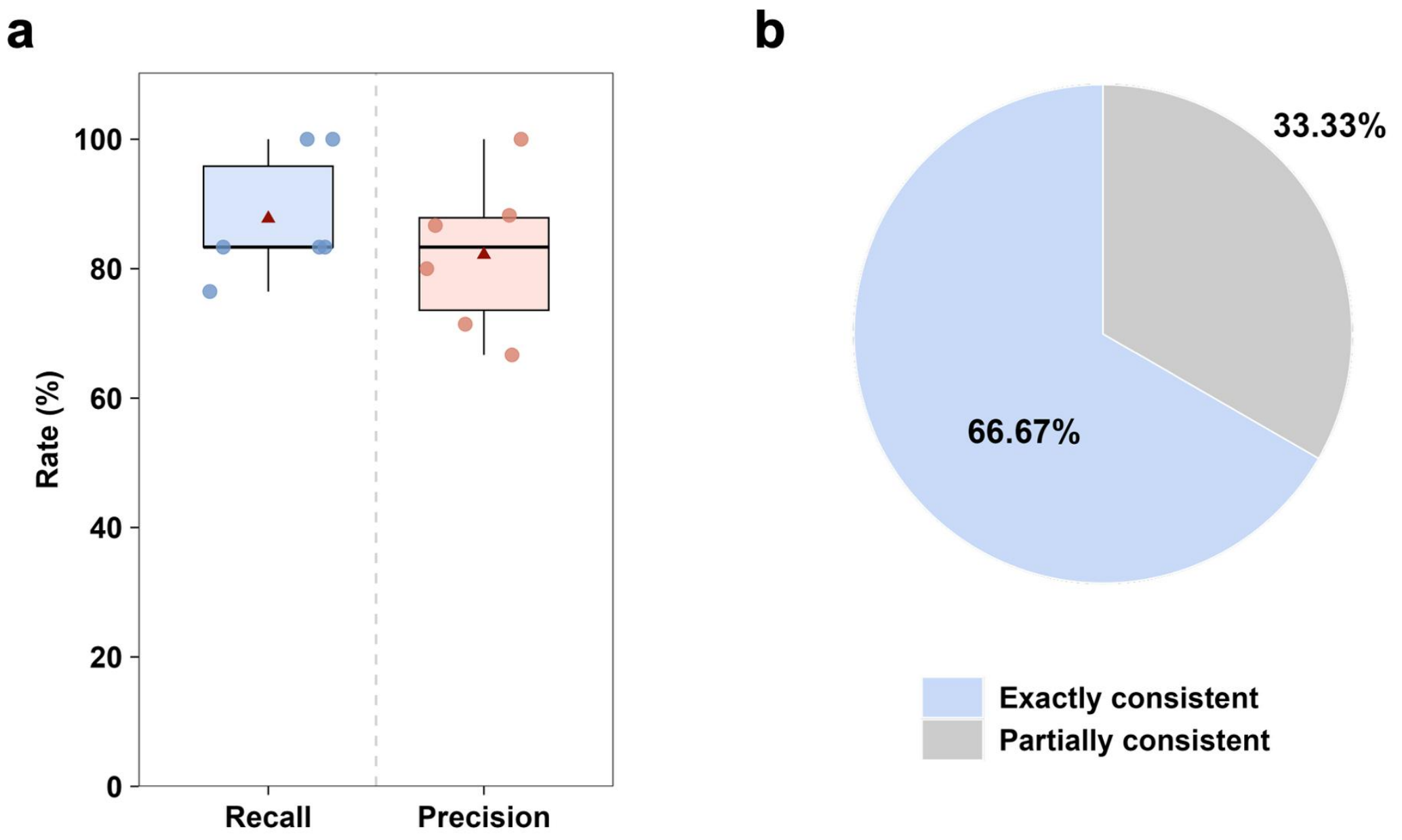

**Fig.5 | human-GUIDE's end-to-end performance in simulated prospective experiments. a,** Screening performance across six post-cutoff NICE guideline datasets, showing consistently high recall and precision despite potential upstream error propagation. In each box plot, the central line denotes the median, the box indicates the interquartile range, the whiskers extend to values within 1.5× the interquartile range, and the triangle represents the mean. **b,** Assessment of drafted recommendations in the final step, showing 100.00% directional agreement with the original NICE recommendations.

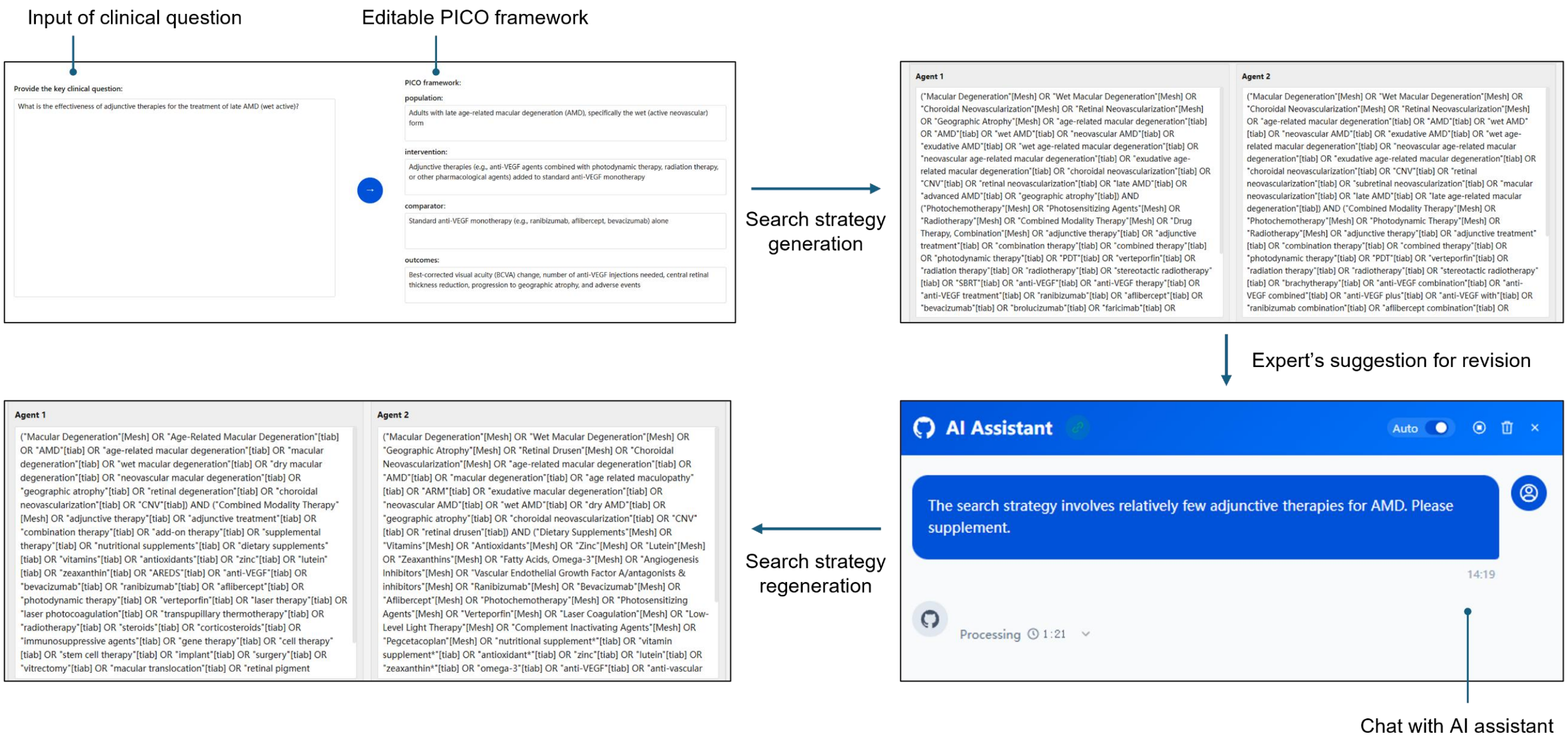

**Supplementary Fig. 1 | Example of search strategy revision through expert's instructions to the AI assistant.** After GUIDE constructs the structured PICO framework from the clinical question, two independent LLM-based search agents generate retrieval strategies in parallel. The AI Reviewer compares the resulting strategies and retrieval outputs to identify potential unreliability, such as low overlap or marked discrepancies in retrieval volume. When revision is needed, experts provide natural-language instructions through the assistant agent, which translates these instructions into updated Boolean search strings, enabling strategy refinement without direct manual editing of complex syntax.

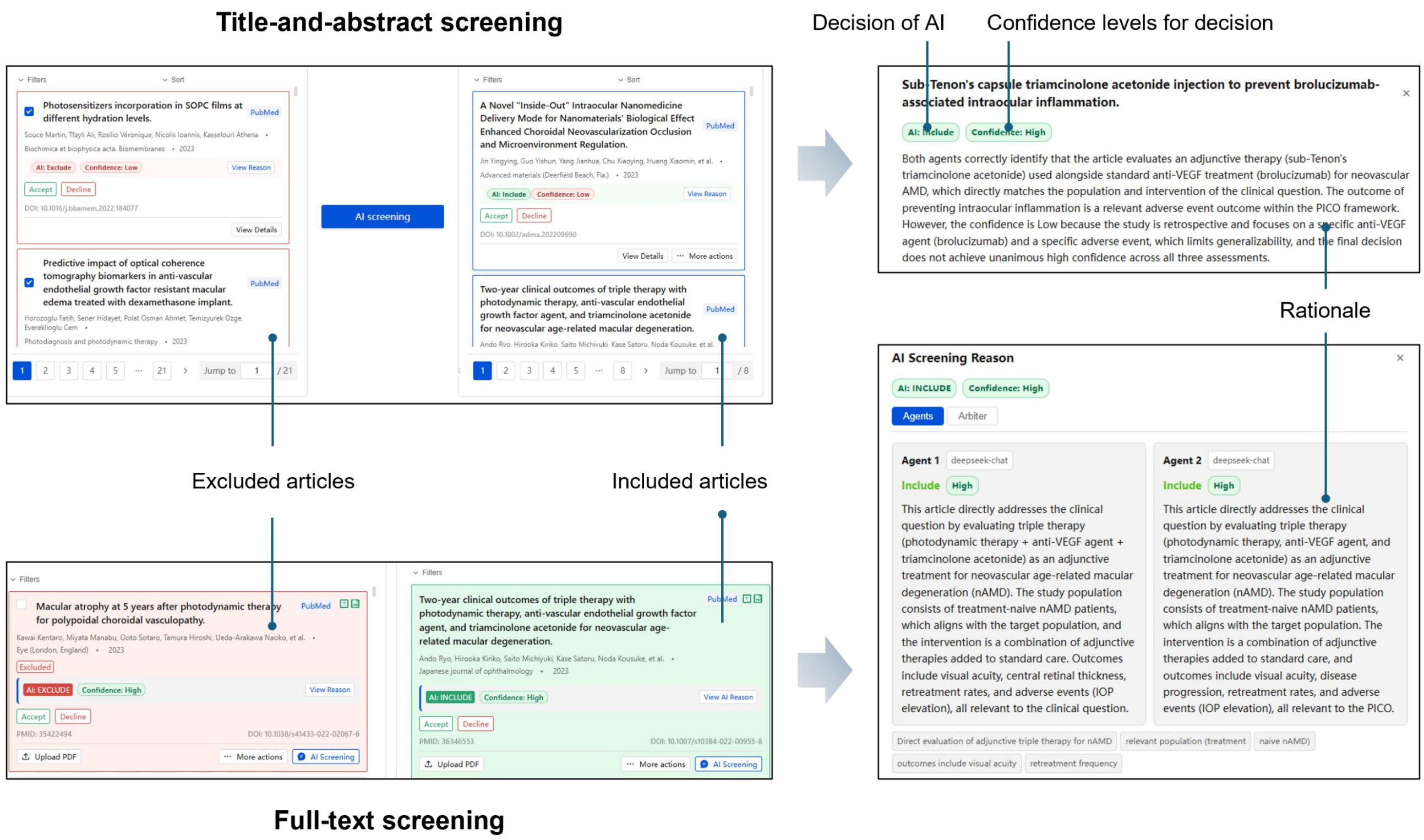

**Supplementary Fig. 2 | GUIDE-assisted literature screening workflow and rationale display.** The interface supports sequential title-and-abstract and full-text screening by parallel screening agents. Retrieved records are independently assessed for eligibility, with inclusion or exclusion labels, confidence levels, and concise rationales displayed for expert review. When agent outputs are consistent, the system presents a summarized screening justification; when discrepancies arise, detailed agent-specific reasoning is shown to facilitate human adjudication and correction.

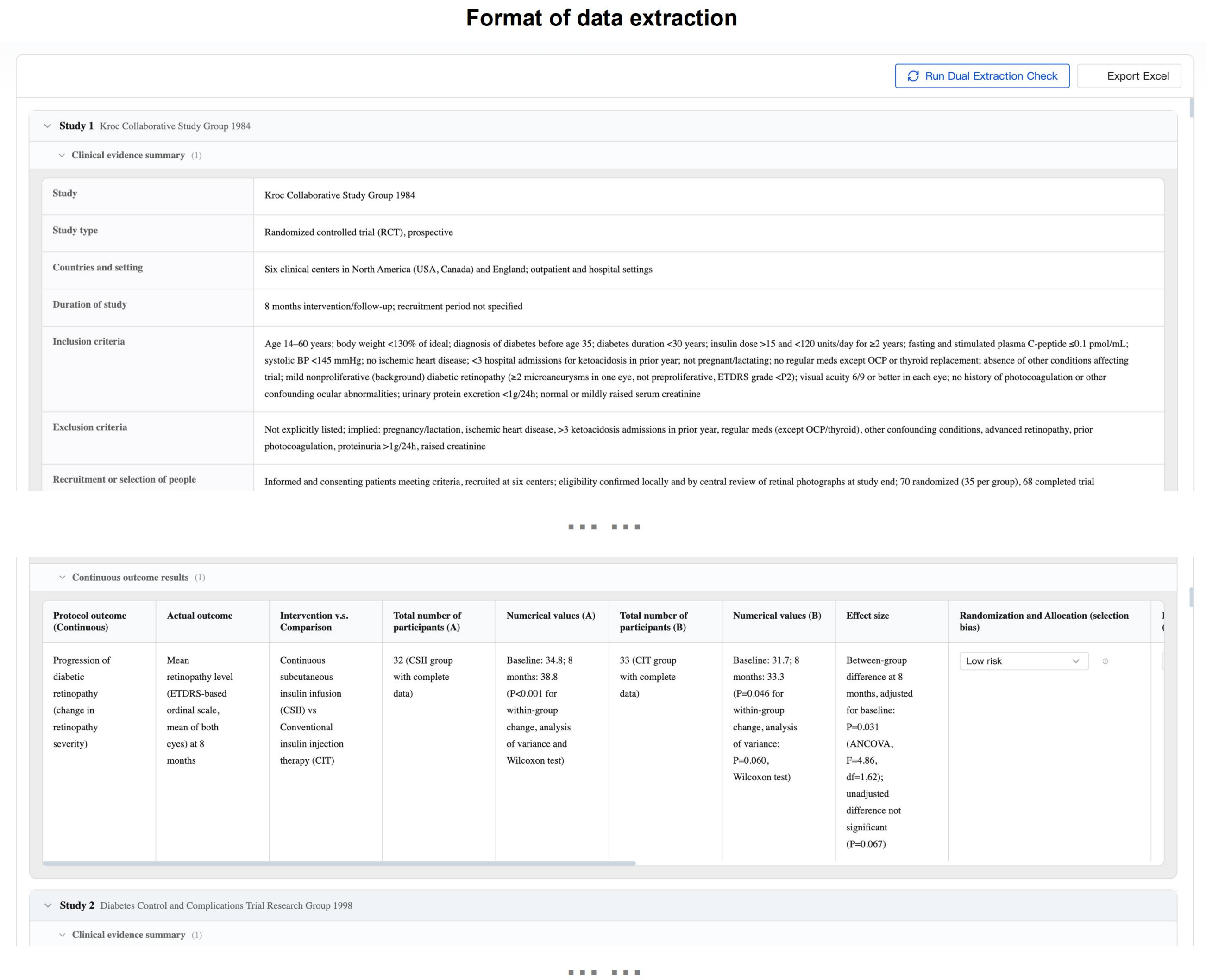

**Supplementary Fig. 3 | Data extraction interface aligned with the PICO framework.**

For studies included after full-text screening, GUIDE supports structured extraction of study-level textual data and outcome-level numerical results. Experts can edit or re-extract individual fields to verify discrepancies against the original full text and ensure consistency with the PICO-defined evidence framework.

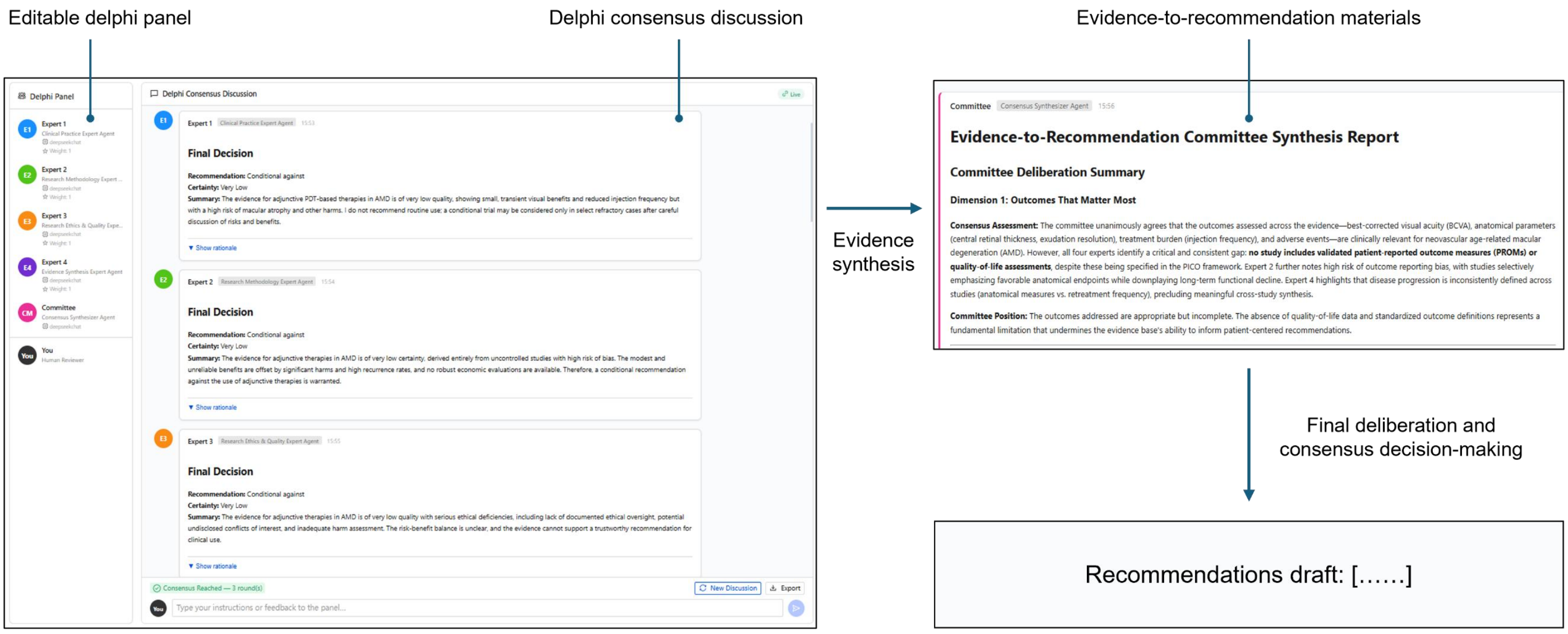

**Supplementary Fig. 4 | Delphi-like multi-agent evidence synthesis and recommendation drafting.** Processed evidence is submitted to a Delphi-like consensus module, in which human experts can select and assemble a virtual panel with different role specifications. Panel members independently review the evidence, provide judgments, and state final recommendation decisions with supporting rationales. The committee agent synthesizes these inputs into an evidence-to-recommendation report and draft recommendation, which experts can further revise and finalize through deliberation and consensus.

**Supplementary Table 1. Proposed suggestions for human-AI collaboration**

| Category | Suggestion | Description |
|---|---|---|
| **Core Principles** | Suggestion 1 | The collaboration should follow a "human-in-the-loop" model: AI executes rule-based, bulk processing and preliminary drafting, whereas human experts provide professional judgement and final decision-making. This collaboration operates within a dynamic framework that iteratively optimizes the human-AI division of labor in alignment with the characteristics and complexity of tasks. |
| **Proposed Human-AI Collaboration Modes** | Suggestion 2 | Proactive human-intervention mode: Applicable to scenarios where AI capabilities are not yet mature or decision-making is critical and requires nuanced judgment. |
| | Suggestion 3 | Condition-triggered human review mode: Applicable to scenarios where AI performance is generally robust, but potential anomalies or uncertainties may arise, triggering human review according to predefined criteria. |

**Supplementary Table 2. Task characteristics across the guideline generation workflow and corresponding preferable human-AI collaboration modes**

| Step | Description | Challenge | Preferable collaboration mode |
|---|---|---|---|
| **PICO Construction** | Formulating the clinical question into PICO components (Population, Intervention, Comparison, Outcome). | Defining a focused scope and selecting clinical outcomes that reflect both risks and benefits. | Proactive human-intervention mode |
| **Search Strategy Formulation** | Developing database-specific search strings based on the PICO framework. | Mapping and expanding core entities into search terms, formulating search strategies in accordance with specific syntax rules. | Condition-triggered human review mode |
| **Literature Screening** | Selecting relevant studies from search results based on inclusion/exclusion criteria. | Processing large volumes of literature following inclusion/exclusion criteria. | Condition-triggered human review mode |
| **Data Extraction** | Collecting key variables and statistical data for analysis and quality assessment. | Handling inconsistent data formats and accurately extracting from primary studies. | Condition-triggered human review mode |
| **Evidence Quality Assessment** | Assessing the certainty of evidence and risk of bias using the GRADE criteria. | Detecting potential bias in both the presence and absence of information, making a holistic judgment tailored to the unique conditions of studies. | Proactive human-intervention mode |
| **Evidence-to-Recommendation (EtR) Synthesis** | Synthesizing clinical evidence into the Evidence-to-Decision report. | Accounting for practical factors like cost, feasibility, and patient values alongside | Proactive human-intervention mode |
| | | clinical effects. | |
| **Recommendation Formulation** | Reconciling expert opinions to reach a final, unified recommendation. | Resolving expert disagreements to produce a clear and deliberate clinical statement. | Proactive human-intervention mode |

**Supplementary Table 3. Overview of the benchmark dataset**

| Specialty | NICE Guideline | Publication Date | Review question |
| --- | --- | --- | --- |
| Ophthalmology | NG242: Diabetic retinopathy: management and monitoring | 13-Aug-24 | What is the effectiveness of different thresholds or criteria for starting treatment for non-proliferative diabetic retinopathy, proliferative diabetic retinopathy, and diabetic macular oedema? |
| | | | What is the effect of intensive treatments to rapidly lower blood glucose levels on progression of diabetic retinopathy and diabetic macular oedema? |
| | | | In people who are about to undergo or who have undergone cataract surgery, what is the effectiveness of treatments (before, during or after surgery) for managing:<br>• non-proliferative diabetic retinopathy<br>• proliferative diabetic retinopathy<br>• diabetic macular oedema? |
| | NG77: Cataracts in adults: management | 26-Oct-17 | What is the optimal strategy to address pre-existing regular astigmatism in people undergoing cataract surgery? |
| | | | Are tinted lenses effective in preventing the progression of age-related macular degeneration compared with colourless lenses in cataract surgery? |
| | | | What is the effectiveness of sedation as an adjunct to local anaesthesia during cataract surgery? |
| | | | What is the effectiveness of interventions to |

| | | |
|---|---|---|
| | | increase pupil size to improve visual outcomes and reduce complications during phacoemulsification cataract surgery? |
| | | What is the optimal type and administration of anesthesia for cataract surgery? |
| NG82: Age-related macular degeneration: diagnosis and management | 23-Jan-18 | What is the effectiveness of psychological therapies for AMD? |
| | | What strategies and tools are useful for self-monitoring for people with AMD? |
| | | What is the effectiveness of adjunctive therapies for the treatment of late AMD (wet active)? |
| NG81: Glaucoma: diagnosis and management | 01-Nov-17 | What is the effectiveness and cost-effectiveness of selective laser trabeculoplasty (SLT) as a first line treatment compared with intraocular pressure lowering eyedrops in ocular hypertension (OHT) or chronic open-angle glaucoma (COAG) adult patients? |

**Supplementary Table 4. Reference Answers and Evaluation Metrics Used in Module Testing**

| Step | Input | GUIDE Output | Reference Answers (from guideline) | Evaluation Metrics |
|---|---|---|---|---|
| **Clinical Question Parsing** | Natural language clinical question (from source guidelines) | Structured PICO elements | PICO definitions from the review protocol | BERTScore |
| **Retrieval Strategy Generation** | PICO representation | Search strategy (databases, filters, study types, keywords) | Search protocol | Recall |
| **Study Selection** | Retrieved candidate studies | Included study list | Included studies reported in the guideline | Recall; Precision |
| **Evidence Extraction** | Included studies, PICO | Structured study-level data (categorical) and quantitative results | Evidence tables with summary of effectiveness evidence | Type classification accuracy; Numerical accuracy |
| **Evidence Quality Assessment** | PICO, Synthesized evidence table | Quality ratings | Quality assessments reported in evidence tables and methods | Classification accuracy |
| **Recommendation Generation** | PICO, Evidence and quality assessment | Guideline-style recommendation | Committee final recommendation | Customized evaluation rubric |

**Supplementary Table 5. Description of the Dataset Composition**

| **Clinical Question Definition** | | | |
|---|---|---|---|
| **Data Field** | **Description** | **Example** | **Data Type** |
| **Clinical Question** | The primary clinical question addressed by the guideline | "What is the effectiveness of different thresholds for initiating treatment in diabetic retinopathy?" | String |
| **Population** | Target patient groups defined in the review (P in PICO) | Non-proliferative DR; proliferative DR; DMO | List[String] |
| **Intervention** | Interventions being evaluated (I in PICO) | Early laser; anti-VEGF; vitrectomy | List[String] |
| **Comparator** | Comparator conditions (C in PICO) | Deferred treatment; standard threshold | List[String] |
| **Outcomes** | Clinical outcomes of interest (O in PICO) | Visual acuity; retinal thickness; quality of life | List[String] |
| **Literature Retrieval and Screening** | | | |
| **Data Field** | **Description** | **Example** | **Data Type** |
| **Retrieved Records** | Bibliographic records identified during the initial literature search | A list of candidate studies returned by database search | List[Record] |
| **Retrieved Record Metadata** | Citation-level metadata for each retrieved record | Title; authors; publication year; journal; database source | Record Object |
| **Screened-in Records** | Records retained after title/abstract screening for full-text assessment | Studies judged potentially eligible after initial screening | List[Record] |
| **Full-text Assessed Records** | Records undergoing full-text eligibility assessment | Candidate studies reviewed in full text | List[Record] |
| **Included Studies** | Final studies included in evidence synthesis | Baker et al., 2019; Elman et al., 2015 | List[Record] |
| **Study-Level Evidence Extraction** | | | |
| **Data Field** | **Description** | **Example** | **Data Type** |
| **Study Identifier** | Unique study reference | Baker 2019 | String |
| **Study Design** | Study methodology | Randomized controlled trial | Categorical |
| **Population Characteristics** | Baseline patient characteristics | DMO with good visual acuity | Text |
| **Intervention** | Description of intervention | Prompt intravitreal anti-VEGF | Text |
| **Details** | | | |
| **Comparator Details** | Description of comparator | Observation with deferred treatment | Text |
| **Outcome Name** | Specific clinical endpoint | Loss of ≥15 letters in BCVA | String |
| **Sample Size (Outcome-Level)** | Number of observations contributing to the estimate | 420 eyes | Integer |
| **Effect Type** | Statistical measure used | Risk Ratio | Categorical |
| **Effect Size** | Estimated effect value | 0.92 | Float |
| **Confidence Interval** | Lower and upper bounds of estimate | [0.82, 1.03] | Tuple |
| **Evidence Quality Assessment** | | | |
| **Data Field** | **Description** | **Example** | **Data Type** |
| **Risk of Bias Assessment Tool** | Tool used for quality assessment | RoB2; ROBINS-I | Categorical |
| **Risk of Bias** | Downgrade due to study limitations | Downgraded (serious limitations) | Categorical |
| **Inconsistency** | Variability across study results | No serious inconsistency | Categorical |
| **Indirectness** | Applicability of evidence to target population | No serious indirectness | Categorical |
| **Imprecision** | Uncertainty in effect estimates | Serious imprecision | Categorical |
| **Other Considerations** | Additional upgrading/downgrading factors | None | String |
| **Final Certainty Rating** | Overall interpretation of effect | Favours early treatment | Categorical |
| **Recommendation-Level Extraction** | | | |
| **Data Field** | **Description** | **Example** | **Data Type** |
| **Evidence Summary** | Narrative summary of findings | Early laser reduces risk of severe visual loss | Text |
| **Benefit Summary** | Summary of benefits | Early treatment reduces progression | Text |
| **Harm Summary** | Summary of harms or risks | Limited or unclear harms | Text |
| **Clinical Judgment** | Expert interpretation beyond data | Early panretinal photocoagulation aligns with clinical practice | Text |
| **Resource Considerations** | Cost or feasibility considerations | Increased resource use for early treatment | Text |
| **Recommendation Statement** | Final summarized recommendation | Offer PRP at diagnosis of proliferative DR | Text |

**Supplementary Table 6. Evaluation criteria for the consistency of recommendations**

| Consistency Level | Description |
|---|---|
| **Exactly Consistent** | The semantic meaning of core elements (e.g., recommendation direction, target population, and interventions) completely matches the original guideline, with no substantive differences. |
| **Partially Consistent** | The overall recommendation direction is consistent, but there are differences or omissions regarding the specific scope of the target population, intervention details, or implementation conditions. |
| **Inconsistent** | The recommendation directions are contradictory, or the system output has no corresponding statement in the original guideline. |

Note: The overall agreement rate is calculated as: Agreement Rate = (Exactly Consistent Recommendations + Partially Consistent Recommendations) / Total Recommendations.

**Supplementary Table 7. Overview of the generalization test dataset**

| Specialty | NICE Guideline | Publication Date | Review question |
|---|---|---|---|
| Cardiovascular | NG106: Chronic heart failure in adults | 3-Sep-25 | What is the clinical and cost effectiveness of mineralocorticoid receptor antagonists in people with heart failure with preserved ejection fraction? |
| | | | What is the clinical and cost effectiveness of salt and/or fluid restriction in people with heart failure? |
| | | | What is the clinical and cost effectiveness of iron supplementation in people with chronic heart failure and iron deficiency? |
| | | | What is the clinical and cost effectiveness of pharmacological interventions for heart failure in people with heart failure who also have chronic kidney disease? |
| Endocrine system | NG243: Adrenal insufficiency: identification and management | 28-Aug-24 | What is the clinical and cost effectiveness of glucocorticoids for the routine management of secondary and tertiary adrenal insufficiency? |
| | | | What is the clinical and cost effectiveness of pharmacological treatments for the routine management of primary adrenal insufficiency? |
| | | | In people at risk of adrenal insufficiency |
| | | | because of prolonged corticosteroid use, what is the best way to manage corticosteroid withdrawal when corticosteroids are no longer needed to control disease activity? |
| Gastrointestinal | NG50: Cirrhosis in over 16s: assessment and management | 8-Sep-23 | What is the clinical and cost-effectiveness of non-selective beta-blockers for the primary prevention of bleeding in people with oesophageal varices due to cirrhosis? |
| | | | What is the clinical and cost-effectiveness of endoscopic band ligation for the primary prevention of bleeding in people with oesophageal varices due to cirrhosis? |
| | | | What is the most clinical and cost-effective prophylactic antibiotic for the primary prevention of bacterial infections in people with cirrhosis and upper gastrointestinal bleeding? |
| | | | What is the clinical and cost-effectiveness of antibiotics compared with placebo for the primary prevention of spontaneous bacterial peritonitis (SBP) in people with cirrhosis and ascites? |
| | | | What is the most clinically and cost-effective intervention for the first-line treatment of an episode of acute hepatic encephalopathy in people with cirrhosis? |
| | | | What is the clinical and cost-effectiveness of transjugular intrahepatic portosystemic shunt (TIPS) compared with large-volume |
| | | | paracentesis (LVP) with albumin in the management of diuretic-resistant ascites due to cirrhosis? |
| | NG231: Barrett’s oesophagus and stage 1 adenocarcinoma | 8-Feb-23 | For adults with Barrett’s oesophagus, what is the clinical and cost effectiveness of pharmacological interventions (such as antacids, aspirin, H2 receptor antagonists, proton pump inhibitors) in reducing progression to dysplasia or cancer? |
| Respiratory system | NG250: Pneumonia: diagnosis and management | 2-Sep-25 | What is the clinical and cost effectiveness of care outside of the acute hospital setting (for example using intermediate care, hospital at home, or virtual wards) for people with community-acquired pneumonia who would otherwise be admitted to hospital? |
| | | | Are shorter durations of antibiotic treatment effective and cost-effective at treating babies, children and young people with community-acquired pneumonia compared to longer treatment courses? |
| | | | In adults with community-acquired pneumonia, what is the clinical and cost effectiveness of corticosteroid treatment in addition to antibiotic treatment, compared with antibiotic treatment alone? |
| | | | The clinical and cost-effectiveness of corticosteroid treatment in addition to antibiotic treatment compared with antibiotic treatment |
| | | | alone in children with CAP or HAP? |
| | NG80: Asthma: diagnosis and monitoring of asthma in adults, children and young people | 29-Nov-17 | In people with asthma, what is the clinical and cost-effectiveness of using measures of pulmonary function assessing asthma control (for example, spirometry and peak expiratory flow) to monitor asthma? |
| | | | In people with asthma, what is the clinical and cost-effectiveness of using indirect challenge tests with mannitol or direct challenge tests with histamine or methacholine for monitoring asthma control? |

**Supplementary Table 8. Overview of the simulate end-to-end prospective test dataset**

| NICE Guideline | Publication Date | Review question |
|---|---|---|
| NG249: Falls: assessment and prevention in older people and people 50 and over at higher risk | 29-Apr-25 | What are the most effective methods for maximising participation, adherence and continuation of falls prevention interventions? |
| | | What are the most clinically and cost-effective methods for falls prevention in older people in residential care settings? |
| NG257: Fertility problems: assessment and treatment | 31-Mar-26 | What is the clinical and cost effectiveness of tests for endometrial receptivity (including gene expression analysis and microbiological analysis) as a treatment add-on for people undergoing fertility treatment? |
| | | What is the clinical and cost effectiveness of treating DNA fragmentation (identified from screening ejaculated sperm) on reproductive outcomes for people with male factor fertility problems? |
| NG253: Suspected sepsis in people aged 16 or over: recognition, assessment and early management | 19-Nov-25 | What is the most clinical and cost-effective intravenous fluid for resuscitation of people aged 16 or over with suspected sepsis? |
| | | In people aged 16 or over with suspected sepsis, how safe is the peripheral administration of intravenous vasopressors compared to central line administration? |

**Supplementary Information 1. PICO criteria (prospective study, GUIDE-assisted group)**

| | Generation by GUIDE | Revision by human expert |
|---|---|---|
| **Population** | Patients undergoing cataract surgery | Adults (18 years and older) undergoing phacoemulsification cataract surgery and intraocular lens (IOL) implantation in one or both eyes |
| **Interventions** | Any type of intraocular lenses to facilitate simultaneous distance and near vision (e.g., multifocal intraocular lenses, monovision with monofocal lenses, extended depth of focus lenses) | 1. Any type of non-accommodating intraocular lenses to facilitate simultaneous distance and near vision (e.g., multifocal intraocular lenses, monovision with monofocal lenses, extended depth of focus lenses)<br>2. Standard monofocal intraocular lenses (providing clear distance vision only, requiring reading glasses for near vision) |
| **Comparators** | Standard monofocal intraocular lenses (providing clear distance vision only, requiring reading glasses for near vision) | All listed interventions vs. each other |
| **Outcomes** | ● Visual acuity at distance and near<br>● Spectacle independence<br>● Contrast sensitivity<br>● Patient satisfaction<br>● Quality of life<br>● Incidence of visual disturbances (e.g., glare, halos)<br>● Need for additional corrective procedures | ● Visual acuity at distance and near<br>● Spectacle independence<br>● Contrast sensitivity<br>● Patient satisfaction<br>● Quality of life<br>● Incidence of visual disturbances (e.g., glare, halos)<br>● Need for additional corrective procedures<br>● Resource use and costs |

**Supplementary Information 2. Summary of included studies (prospective study, GUIDE-assisted group)**

| Study & location | Sample size | Methods |
|---|---|---|
| Cochener (2018)<br>France | 120 eyes<br>(60 patients) | Prospective randomized comparative study of bilateral AcrySof IQ PanOptix and FineVision Micro F trifocal IOLs versus the TECNIS Symfony EDOF IOL. |
| Vandekerckhove (2026)<br>Switzerland | 210 eyes<br>(105 patients) | Single-center prospective randomized double-blind trial comparing bilateral Isopure, Eyhance, and Vivity presbyopia-correcting IOLs with mini-monovision (−0.50 D). |
| McCabe (2022)<br>USA | 435 eyes<br>(218 patients) | Prospective multicenter randomized parallel controlled, patient- and assessor-masked trial comparing bilateral DFT015 (Vivity) nondiffractive EDOF IOLs with SN60WF monofocal IOLs. |
| Turhan (2021) / Sevik (2022)<br>Turkey | 58 eyes<br>(29 patients) | Prospective randomized comparative study of bilateral TECNIS ZKB00 low-add bifocal IOLs versus TECNIS Symfony EDOF IOLs, both using mini-monovision (−0.50 D). |
| Chang (2023)<br>USA | 272 patients | Prospective multicenter bilateral randomized, three-way-masked comparative trial of TECNIS Synergy ZFR00V hybrid IOLs versus TECNIS ZCB00 monofocal IOLs. |
| Zeilinger (2025)<br>Austria | 80 eyes<br>(40 patients) | Single-center randomized double-masked controlled trial comparing bilateral RayOne EMV advanced monofocal IOLs with Vivity nondiffractive EDOF IOLs using mini-monovision. |
| Han (2025)<br>China | 100 eyes<br>(100 patients) | Prospective randomized case-control study comparing an aspheric monofocal IOL with a multifocal IOL in monocular cataract surgery. |
| Li (2023)<br>China | 99 eyes<br>(99 patients) | Prospective randomized double-masked controlled trial comparing unilateral AT LISA toric 939M/MP trifocal IOLs with AT LISA toric 909M bifocal IOLs. |
| Webers (2020)<br>Netherlands | 54 eyes<br>(27 patients) | Prospective randomized controlled trial comparing bilateral TECNIS Symfony EDOF IOLs with micromonovision versus bilateral AT LISA tri 839MP trifocal IOLs. |
| Reinhard (2021)<br>Germany, Belgium, | 211 patients | Prospective multicenter randomized three-arm trial comparing bilateral AT LARA and TECNIS Symfony EDOF IOLs with CT ASPHINA monofocal IOLs. |
| Portugal, Spain, France, and Italy | | |
| Hida (2024) Brazil | 116 eyes (58 patients) | Prospective interventional randomized parallel study comparing bilateral TECNIS Synergy ZFR00V implantation with a blended strategy (Eyhance ICB00 in the dominant eye and Synergy in the nondominant eye). |
| Sandoval (2020) USA | 76 patients | Prospective unmasked randomized study of bilateral toric TECNIS Symfony EDOF implantation targeting emmetropia in both eyes versus −0.50 D mini-monovision in the nondominant eye. |
| Pradeep (2026) India | 70 patients | Prospective randomized comparative study of bilateral PanOptix trifocal versus Vivity EDOF IOL implantation. |
| Gil (2022) Spain | 152 patients | Prospective double-masked comparative study in which six bilateral presbyopia-correcting IOL groups were block-randomized 1:1:1:1:1:1 (SV25T0, AT LISA 809M, AT LISA tri 839MP, TECNIS ZKB00, TECNIS ZLB00, and TECNIS Symfony ZXR00), with a nonrandomized bilateral TECNIS ZA9003 monofocal comparator group. |
| Messias (2021) Brazil | 66 eyes (33 patients) | Prospective randomized study comparing bilateral SN6AD1 diffractive multifocal IOLs with SN60WF aspheric monofocal IOLs. |
| Labiris (2024) Greece | 90 patients | Prospective randomized comparative study of premium monovision (PanOptix in the nondominant eye and Vivity in the dominant eye) versus bilateral PanOptix or bilateral Vivity implantation. |
| Bala (2022) Australia, Canada, Spain, and UK | 270 patients | Prospective multicenter randomized parallel controlled, patient- and assessor-masked trial comparing bilateral DFT015 (Vivity) nondiffractive EDOF IOLs with SN60WF monofocal IOLs. |
| Skov (2026) Denmark | 180 patients | Single-center randomized controlled pilot trial comparing four bilateral strategies: partial-range-of-focus narrow IOLs targeting emmetropia, mini-monovision (−1.25 D), monovision (−2.50 D), and partial-range-of-focus extended IOLs. |
| Corbett / Black (2024) Australia and New | 118 patients | Prospective multicenter randomized subject- and evaluator-masked trial comparing bilateral TECNIS PureSee ZEN00V refractive EDOF IOLs with Eyhance ICB00 enhanced monofocal IOLs. |
| Zealand | | |
| Pantanelli (2023)<br>USA | 51 patients | Prospective single-center double-arm randomized, patient- and assessor-masked trial comparing bilateral Vivity nondiffractive EDOF IOLs with enVista neutral aspheric monofocal IOLs targeted for emmetropia. |
| Gil (2020)<br>Spain | 116 patients | Prospective randomized double-masked study comparing six bilateral presbyopia-correcting IOLs (SV25T0, AT LISA 809M, AT LISA tri 839MP, TECNIS ZKB00, TECNIS ZLB00, and TECNIS Symfony). |
| Guarro (2022)<br>Spain | 88 patients | Prospective randomized double-blind study comparing bilateral Vivity, AT LARA, and TECNIS Symfony EDOF IOLs with SN60WF monofocal IOLs. |
| Law (2020)<br>UK | 90 patients | Prospective randomized double-masked study comparing bilateral Bi-Flex 677MY multifocal IOLs with Bi-Flex 677AB monofocal IOLs. |
| Chang (2022)<br>USA | 299 patients | Prospective multicenter bilateral randomized, patient- and evaluator-masked trial comparing TECNIS Symfony ZXR00 EDOF IOLs with TECNIS ZCB00 monofocal IOLs. |
| Palkovits (2026)<br>Austria | 35 patients | Prospective randomized single-center study comparing bilateral Acunex Vario AN6V EDOF IOLs with Acunex AN6 monofocal IOLs. |

**Supplementary Information 3. Excluded studies and reasons (prospective study, GUIDE-assisted group)**

| **Study** | **Reason for exclusion** |
|---|---|
| Savini G, Schiano-Lomoriello D, Balducci N, Barboni P. Visual performance of a new extended depth-of-focus intraocular lens compared to a distance-dominant diffractive multifocal intraocular lens. *Journal of Refractive Surgery*. 2018;34(4):228-235. | Retrospective comparative study. |
| Escandón-García S, Ribeiro FJ, McAlinden C, Queirós A, González-Méijome JM. Through-focus vision performance and light disturbances of 3 new intraocular lenses for presbyopia correction. *Journal of Ophthalmology*. 2018;2018:6165493. | Nonrandomized case series. |
| Mencucci R, Favuzza E, Caporossi O, Savastano A, Rizzo S. Comparative analysis of visual outcomes, reading skills, contrast sensitivity, and patient satisfaction with two models of trifocal diffractive intraocular lenses and an extended range of vision intraocular lens. *Graefe's Archive for Clinical and Experimental Ophthalmology*. 2018;256(10):1913-1922. | Nonrandomized case series. |
| Pedrotti E, Carones F, Aiello F, Mastropasqua R, Bruni E, Bonacci E, et al. Comparative analysis of visual outcomes with 4 intraocular lenses: monofocal, multifocal, and extended range of vision. *Journal of Cataract & Refractive Surgery*. 2018;44(2):156-167. | Participants were not randomly allocated to the IOL groups.* |
| Fernandes MC, Nunomura CY, Messias A. Visual performance and photic disturbances with diffractive and nondiffractive EDOF intraocular lenses using mini-monovision: randomized trial. Journal of Cataract & Refractive Surgery. 2024;50(2):153–159. | Both groups received EDOF IOLs. |
| de Medeiros AL, Jones Saraiva F, Iguma CI, Kniggendorf DV, Alves G, Chaves MAPD, et al. Comparison of visual outcomes after bilateral implantation of two intraocular lenses with distinct diffractive optics. *Clinical Ophthalmology*. 2019;13:1657-1663. | Nonrandomized case series. |
| Rodov L, Reitblat O, Levy A, Assia EI, Kleinmann G. Visual outcomes | Retrospective review. |
| and patient satisfaction for trifocal, extended depth of focus and monofocal intraocular lenses. *Journal of Refractive Surgery*. 2019;35(7):434-440. | |
| Wanten JC, Bauer NJC, Berendschot TTJM, van den Biggelaar FJHM, Nuijts RMMA. Two novel extended depth-of-focus intraocular lenses targeted for mini-monovision: A prospective randomized controlled trial. American Journal of Ophthalmology. 2025;276:286–296. | Both groups received EDOF IOLs.* |
| Singh B, Sharma S, Dadia S, Bharti N, Bharti S. Comparative evaluation of visual outcomes after bilateral implantation of a diffractive trifocal intraocular lens and an extended depth of focus intraocular lens. *Eye & Contact Lens*. 2020;46(5):314-318. | Nonrandomized comparative study. |
| Paik DW, Park JS, Yang CM, Lim DH, Chung TY. Comparing the visual outcome, visual quality, and satisfaction among three types of multi-focal intraocular lenses. *Scientific Reports*. 2020;10:14832. | Nonrandomized comparative study. |
| Farvardin M, Johari M, Attarzade A, Rahat F, Farvardin R, Farvardin Z. Comparison between bilateral implantation of a trifocal intraocular lens (Alcon AcrySof IQ PanOptix) and extended depth of focus lens (Tecnis Symfony ZXR00 lens). *International Ophthalmology*. 2021;41(2):567-573. | Nonrandomized comparative study. |
| Hayashi K, Yoshida M, Sasaki H, Hirata A. Binocular visual function of myopic pseudophakic monovision. Japanese Journal of Ophthalmology. 2018;62:357–364. | Intervention did not meet predefined scope.* |
| Chao CC, Lin HY, Lee CY, Mai ELC, Lian IB, Chang CK. Difference in quality of vision outcome among extended depth of focus, bifocal, and monofocal intraocular lens implantation. *Healthcare*. 2022;10(6):1000. | Retrospective comparative study. |
| Zhu M, Fan W, Zhang G. Stereopsis and visual acuity: bilateral trifocal versus blended extended depth of focus and diffractive bifocal intraocular lenses. *Frontiers in Medicine*. 2022;9:1042101. | Nonrandomized comparative study. |
| Nagy ZZ, Popper-Sachetti A, Kiss HJ. Comparison of visual and refractive outcomes between hydrophilic and hydrophobic trifocal intraocular lenses | Both IOLs were trifocal lenses. |
| sharing the same optical design. *Journal of Cataract & Refractive Surgery*. 2019;45(5):553-561. | |
| Dick HB, Ang RE, Corbett D, Hoffmann P, Tetz M, Villarrubia A, Palomino C, Castillo-Gomez A, Tsai L, Thomas EK, Janakiraman P. Comparison of 3-month visual outcomes of a new multifocal intraocular lens vs a trifocal intraocular lens. Journal of Cataract & Refractive Surgery. 2022;48(11):1270–1276. | Cataract-only patients outcome data were not extractable.* |
| Nagy ZZ, Dorman P, Szalczer S, Kiss H. Prospective Comparison of Clinical Outcomes After Bilateral Implantation of Diffractive Trifocal Extended Depth-of-Focus and Diffractive Trifocal Hydrophobic Intraocular Lenses. Journal of Refractive Surgery. 2025;41(3):e102–e113. | Cataract-only patients outcome data were not extractable.* |
| Vasavada V, Vasavada SA, Shastri L, Nath V, Vasavada AR, Srivastava S. Visual outcomes comparing emmetropia vs mini-monovision after bilateral implantation of a nondiffractive extended vision intraocular lens: randomized trial. Journal of Cataract & Refractive Surgery. 2024;50(8):799–804. | Cataract-only patients outcome data were not extractable.* |
| Martínez de Carneros-Llorente A, Martínez de Carneros A, Martínez de Carneros-Llorente P, Jiménez-Alfaro I. Comparison of visual quality and subjective outcomes among 3 trifocal intraocular lenses and 1 bifocal intraocular lens. *Journal of Cataract & Refractive Surgery*. 2019;45(5):587-594. | All IOLs belonged to the multifocal class. |
| Asena BS. Visual and refractive outcomes, spectacle independence, and visual disturbances after cataract or refractive lens exchange surgery: comparison of 2 trifocal intraocular lenses. *Journal of Cataract & Refractive Surgery*. 2019;45(11):1539-1546. | Both study groups received trifocal IOLs. |
| Wendelstein JA, Vallotton K, Ziörjen A, Müller M, Riaz KM, Pantanelli SM, Langenbucher A, Seiler TG. Reading Speed and Visual Acuity in Photopic and Mesopic Conditions After Bilateral Implantation of Diffractive Multifocal Intraocular Lenses. American Journal of | Both groups received diffractive multifocal IOLs. |
| Ophthalmology. 2026;281:516–525. | |
| Schojai M, Schultz T, Jerke C, Böcker J, Dick HB. Visual performance comparison of 2 extended depth-of-focus intraocular lenses. Journal of Cataract & Refractive Surgery. 2020;46(3):388–393. | Both groups evaluated EDOF-based IOLs. |
| Ferreira TB, Ribeiro FJ. Prospective comparison of clinical performance and subjective outcomes between two diffractive trifocal intraocular lenses in bilateral cataract surgery. *Journal of Refractive Surgery*. 2019;35(7):418-425. | Both study groups received trifocal IOLs. |
| Ribeiro FJ, Ferreira TB. Comparison of visual and refractive outcomes of 2 trifocal intraocular lenses. *Journal of Cataract & Refractive Surgery*. 2020;46(5):694-699. | Both study groups received diffractive trifocal IOLs. |
| Xun Y, Wan W, Jiang L, Hu K. Crossed versus conventional pseudophakic monovision for high myopic eyes: a prospective, randomized pilot study. BMC Ophthalmology. 2020;20:447. | Enrolled high-myopic cataract patients. |
| Ang RE, Picache GCS, Rivera MCR, Lopez LRL, Cruz EM. A comparative evaluation of visual, refractive, and patient-reported outcomes of three extended depth of focus (EDOF) intraocular lenses. *Clinical Ophthalmology*. 2020;14:2339-2351. | Retrospective cohort study. |
| Hienert J, Stjepanek K, Hirnschall N, Ruiss M, Zwickl H, Findl O. Visual performance of two diffractive trifocal intraocular lenses: a randomized trial. *Journal of Refractive Surgery*. 2021;37(7):460-465. | Both study groups received diffractive trifocal IOLs. |
| Kılıç A, Özpınar A, Tanrıverdi C. Comparison of visual outcomes of two trifocal IOLs. *Journal of Refractive Surgery*. 2023;39(8):524-530. | Both study groups received diffractive trifocal IOLs. |
| De Rojas JO, Sandoval HP, Potvin R, Solomon KD. Comparison of visual and subjective outcomes between two EDOF intraocular lenses, with emphasis on mesopic performance. *Journal of Cataract & Refractive Surgery*. 2026;52(6):574-581. | Both study groups received EDOF IOLs. |

Note: The 6 articles marked with asterisks were excluded after review by human experts.

## Supplementary Information 4. GRADE tables (prospective study, GUIDE-assisted group)

### (1) Multifocal vs Monofocal

#### Visual acuity

| No. of studies | Design | Risk of bias | Inconsistency | Indirectness | Imprecision | No. of participants | Effect size (95% CI) | Quality |
|---|---|---|---|---|---|---|---|---|
| Corrected distance visual acuity worse than 6/6 (lower favours multifocal) | | | | | | | | |
| 9 (Chang 2023; elMaghraby 1992; Steinert 1992; Percival 1993; Rossetti 1994; Haaskjold 1998; Kamlesh 2001; Leyland 2002; Sen 2004) | RCT | Serious | Serious | Not serious | Serious | 953 | RR 1.09 (0.75 to 1.58) | Very low |
| Uncorrected distance visual acuity worse than 6/6 (lower favours multifocal) | | | | | | | | |
| 9 (Chang 2023; elMaghraby 1992; Steinert 1992; Percival 1993; Rossetti 1994; Haaskjold 1998; Leyland 2002; Sen 2004; Jusufovic 2011) | RCT | Serious | Serious | Not serious | Not serious | 943 | RR 1.01 (0.89 to 1.14) | Low |
| Uncorrected near visual acuity worse than J3/J4 or an equivalent threshold (lower favours multifocal) | | | | | | | | |
| 9 (Chang 2023; elMaghraby 1992; Percival 1993; Rossetti 1994; Haaskjold 1998; Javitt 2000; Leyland 2002; Jusufovic 2011; Ji 2013) | RCT | Serious | Serious | Not serious | Not serious | 1,043 | RR 0.18 (0.08 to 0.41) | Low |
| Mean corrected distance visual acuity (logMAR) (lower favours multifocal) | | | | | | | | |
| 9 (Chang 2023; Messias 2021; Law 2020; Leyland 2002; Nijkamp 2004; Harman 2008; Palmer 2008; Rasp 2012; Peng 2012) | RCT | Serious | Not serious | Not serious | Not serious | 1,265 | MD 0.03 (0.01 to 0.04) | Moderate |
| Mean corrected intermediate visual acuity (logMAR) (lower favours multifocal) | | | | | | | | |
| 4 (Chang 2023; Law 2020; Maxwell 2017; Peng 2012) | RCT | Serious | Serious | Not serious | Serious | 866 | MD −0.12 (−0.29 to 0.05) | Very low |
| Mean corrected near visual acuity (logMAR) (lower favours multifocal) | | | | | | | | |
| 9 (Chang 2023; Law 2020; Javitt 2000; Leyland 2002; | RCT | Serious | Very serious | Not serious | Serious | 1,667 | MD −0.15 (−0.34 to 0.05) | Very low |
| Palmer 2008; Harman 2008; Rasp 2012; Peng 2012; Maxwell 2017) | | | | | | | | |
| Mean uncorrected distance visual acuity (logMAR) (lower favours multifocal) | | | | | | | | |
| 9 (Chang 2023; Messias 2021; Law 2020; Leyland 2002; Nijkamp 2004; Palmer 2008; Harman 2008; Peng 2012; Rasp 2012) | RCT | Serious | Not serious | Not serious | Not serious | 1,265 | MD 0.02 (−0.01 to 0.05) | Moderate |
| Mean uncorrected intermediate visual acuity (logMAR) (lower favours multifocal) | | | | | | | | |
| 3 (Chang 2023; Maxwell 2017; Peng 2012) | RCT | Serious | Serious | Not serious | Serious | 776 | MD −0.11 (−0.27 to 0.05) | Very low |
| Mean uncorrected near visual acuity (logMAR) (lower favours multifocal) | | | | | | | | |
| 9 (Chang 2023; Messias 2021; Law 2020; Javitt 2000; Leyland 2002; Harman 2008; Peng 2012; Rasp 2012; Maxwell 2017) | RCT | Serious | Serious | Not serious | Not serious | 1,559 | MD −0.25 (−0.39 to −0.12) | Low |

**Visual function**

| No. of studies | Design | Risk of bias | Inconsistency | Indirectness | Imprecision | No. of participants | Effect size (95% CI) | Quality |
|---|---|---|---|---|---|---|---|---|
| Any spectacle dependence (lower favours multifocal) | | | | | | | | |
| 15 (Chang 2023; Han 2025; Law 2020; Gil 2022; Steinert 1992; Percival 1993; Rossetti 1994; Haaskjold 1998; Javitt 2000; Leyland 2002; Harman 2008; Cillino 2008; Zhao 2010; Peng 2012; Maxwell 2017) | RCT | Serious | Serious | Not serious | Not serious | 1,899 | RR 0.52 (0.39 to 0.68) | Low |
| Contrast sensitivity (higher favours multifocal) | | | | | | | | |
| 4 (Harman 2008; Leyland 2002; Rossetti 1994; Sen 2004) | RCT | Serious | Not serious | Not serious | Serious | 288 | SMD −0.19 (−0.57 to 0.19) | Low |
| Visual function (higher favours multifocal) | | | | | | | | |
| 6 (Han 2025; Law 2020; Cillino 2008; Nijkamp 2004; | RCT | Serious | Serious | Not serious | Serious | 663 | SMD 0.78 (−0.09 to 1.65) | Very low |
| Sen 2004; Zhao 2010) | | | | | | | | |
| Patient satisfaction (number of satisfied participants) (higher favours multifocal) | | | | | | | | |
| 3 (Chang 2023; Hida 2024; Law 2020) | RCT | Serious | Serious | Not serious | Serious | 403 | RR 2.41 (0.47 to 12.46) | Low |
| Patient satisfaction (questionnaire score) (higher favours multifocal) | | | | | | | | |
| 6 (Cillino 2008; Nijkamp 2004; Peng 2012; Sen 2004; Steinert 1992; Zhao 2010) | RCT | Serious | Very serious | Not serious | Very serious | 643 | SMD 0.26 (−0.37 to 0.89) | Very low |

**Adverse outcomes**

| No. of studies | Design | Risk of bias | Inconsistency | Indirectness | Imprecision | No. of participants | Effect size (95% CI) | Quality |
|---|---|---|---|---|---|---|---|---|
| Glare (lower favours multifocal) | | | | | | | | |
| 12 (Chang 2023; Han 2025; Hida 2024; Gil 2022; Percival 1993; Rossetti 1994; Haaskjold 1998; Kamlesh 2001; Sen 2004; Cillino 2008; Harman 2008; Maxwell 2017) | RCT | Serious | Serious | Not serious | Serious | 1,405 | RR 1.34 (0.98 to 1.83) | Very low |
| Halos (lower favours multifocal) | | | | | | | | |
| 13 (Chang 2023; Han 2025; Hida 2024; Gil 2022; Law 2020; Cillino 2008; Haaskjold 1998; Kamlesh 2001; Maxwell 2017; Percival 1993; Rossetti 1994; Sen 2004; Zhao 2010) | RCT | Serious | Serious | Not serious | Not serious | 1,614 | RR 4.06 (2.19 to 7.55) | Low |
| Starbursts (lower favours multifocal) | | | | | | | | |
| 2 (Chang 2023; Han 2025) | RCT | Serious | Not serious | Not serious | Very serious | 362 | RR 6.40 (0.00 to 33991.66) | Very low |
| Nd:YAG posterior capsulotomy (lower favours multifocal) | | | | | | | | |
| 1 (Law 2020) | RCT | Serious | N/A | Not serious* | Very serious | 180 | RR 2.55 (0.68 to 9.55) | Very low |

## (2) EDOF vs Monofocal

**Visual acuity**

| No. of studies | Design | Risk of bias | Inconsistency | Indirectness | Imprecision | No. of participants | Effect size (95% CI) | Quality |
|---|---|---|---|---|---|---|---|---|
| Corrected distance visual acuity worse than 6/6 (lower favours EDOF) | | | | | | | | |
| 2 (Black 2024; Chang 2022) | RCT | Serious | Serious | Not serious | Very serious | 415 | RR 0.80 (0.00, 5055.39) | Very low |
| Corrected intermediate visual acuity meeting the study-defined threshold (higher favours EDOF) | | | | | | | | |
| 2 (McCabe 2022; Pantanelli 2023) | RCT | Serious | Not serious | Not serious | Not serious | 269 | RR 2.91 (0.43, 19.82) | Moderate |
| Mean corrected distance visual acuity (logMAR) (lower favours EDOF) | | | | | | | | |
| 5 (Reinhard 2021; Bala 2022; Corbett 2024; Black 2024; Pantanelli 2023) | RCT | Serious | Not serious | Not serious | Not serious | 764 | MD 0.02 (−0.01, 0.05) | Moderate |
| Mean corrected intermediate visual acuity (logMAR) (lower favours EDOF) | | | | | | | | |
| 4 (Reinhard 2021; Bala 2022; Corbett 2024; Pantanelli 2023) | RCT | Serious | Serious | Not serious | Serious | 644 | MD −0.15 (−0.28, −0.02) | Very low |
| Mean corrected near visual acuity (logMAR) (lower favours EDOF) | | | | | | | | |
| 4 (Reinhard 2021; Bala 2022; Corbett 2024; Pantanelli 2023) | RCT | Serious | Serious | Not serious | Not serious | 644 | MD −0.13 (−0.24, −0.01) | Low |
| Mean uncorrected distance visual acuity (logMAR) (lower favours EDOF) | | | | | | | | |
| 6 (McCabe 2022; Reinhard 2021; Bala 2022; Skov 2026; Black 2024; Pantanelli 2023) | RCT | Serious | Not serious | Not serious | Not serious | 893 | MD 0.03 (−0.00, 0.07) | Moderate |
| Mean uncorrected intermediate visual acuity (logMAR) (lower favours EDOF) | | | | | | | | |
| 5 (McCabe 2022; Reinhard 2021; Bala 2022; Skov 2026; Pantanelli 2023) | RCT | Serious | Serious | Not serious | Serious | 833 | MD −0.12 (−0.21, −0.02) | Very low |
| Mean uncorrected near visual acuity (logMAR) (lower favours EDOF) | | | | | | | | |
| 5 (McCabe 2022; Reinhard 2021; Bala 2022; Skov 2026; Pantanelli 2023) | RCT | Serious | Serious | Not serious | Serious | 833 | MD −0.16 (−0.24, −0.08) | Very low |
| Uncorrected distance visual acuity worse than 6/6 (lower favours EDOF) | | | | | | | | |
| 2 (Black 2024; Chang | RCT | Serious | Not serious | Not serious | Serious | 355 | RR 1.15 (0.32, 4.12) | Low |
| 2022) | | | | | | | | |
| Uncorrected near visual acuity worse than J3/J4 or equivalent (lower favours EDOF) | | | | | | | | |
| 1 (Chang 2022) | RCT | Serious | N/A | Not serious | Not serious | 295 | RR 0.11 (0.05, 0.25) | Moderate |

**Visual function**

| No. of studies | Design | Risk of bias | Inconsistency | Indirectness | Imprecision | No. of participants | Effect size (95% CI) | Quality |
|---|---|---|---|---|---|---|---|---|
| Binocular Visual Function questionnaire score (scale direction not reported) (lower favours EDOF) | | | | | | | | |
| 1 (Skov 2026) | RCT | Serious | N/A | Not serious | Serious | 92 | MD −0.16 (−0.41, 0.09) | Low |
| Contrast sensitivity (higher favours EDOF) | | | | | | | | |
| 1 (Skov 2026) | RCT | Serious | N/A | Not serious | Not serious | 92 | MD −0.01 (−0.06, 0.04) | Moderate |
| Mean stereoacuity (arcsec) (lower favours EDOF) | | | | | | | | |
| 1 (Skov 2026) | RCT | Serious | N/A | Not serious | Very Serious | 92 | MD 29.0 (−15.4, 73.4) | Very Low |
| Light distortion index (lower favours EDOF) | | | | | | | | |
| 1 (Guarro 2022) | RCT | Serious | N/A | Not serious* | Serious | 88 | MD 9.33 (3.81, 14.85) | Low |
| Patient satisfaction (higher favours EDOF) | | | | | | | | |
| 2 (Skov 2026; Black 2024) | RCT | Serious | Serious | Not serious | Serious | 152 | RR 1.17 (0.09, 14.57) | Very low |
| Spectacle dependence, any (lower favours EDOF) | | | | | | | | |
| 8 (McCabe 2022; Bala 2022; Skov 2026; Black 2024; Pantanelli 2023; Chang 2022; Palkovits 2026; Gil 2022) | RCT | Serious | Serious | Not serious | Serious | 985 | RR 0.70 (0.59, 0.85) | Very low |

**Adverse outcomes**

| No. of studies | Design | Risk of bias | Inconsistency | Indirectness | Imprecision | No. of participants | Effect size (95% CI) | Quality |
|---|---|---|---|---|---|---|---|---|
| Blurred vision (lower favours EDOF) | | | | | | | | |
| 3 (McCabe 2022; Pantanelli 2023; Chang 2022) | RCT | Serious | Not serious | Not serious | Very serious | 565 | RR 0.50 (0.12, 2.06) | Very low |
| Double vision (lower favours EDOF) | | | | | | | | |
| 1 (McCabe 2022) | RCT | Serious | N/A | Not serious | Very serious | 219 | RR 2.13 (0.20, 23.17) | Very low |
| Glare (lower favours EDOF) | | | | | | | | |
| 9 (McCabe 2022; Bala 2022; | RCT | Serious | Not serious | Not serious | Serious | 1,133 | RR 0.93 (0.73, 1.19) | Low |
| Skov 2026; Corbett 2024; Pantanelli 2023; Guarro 2022; Chang 2022; Palkovits 2026; Gil 2022) | | | | | | | | |
| Halos (lower favours EDOF) | | | | | | | | |
| 10 (McCabe 2022; Gil 2022; Bala 2022; Skov 2026; Corbett 2024; Black 2024; Pantanelli 2023; Guarro 2022; Chang 2022; Palkovits 2026) | RCT | Serious | Serious | Not serious | Not serious | 1,192 | RR 2.08 (1.16, 3.72) | Low |
| Hazy vision (lower favours EDOF) | | | | | | | | |
| 2 (McCabe 2022; Pantanelli 2023) | RCT | Serious | Not serious | Not serious* | Very serious | 270 | RR 0.78 (0.01, 83.56) | Very low |
| Negative dysphotopsia (lower favours EDOF) | | | | | | | | |
| 1 (McCabe 2022) | RCT | Serious | N/A | Not serious | Very serious | 219 | RR 0.68 (0.27, 1.69) | Very low |
| Starbursts (lower favours EDOF) | | | | | | | | |
| 8 (McCabe 2022; Bala 2022; Skov 2026; Corbett 2024; Black 2024; Pantanelli 2023; Chang 2022; Palkovits 2026) | RCT | Serious | Serious | Not serious | Very serious | 1,055 | RR 0.97 (0.59, 1.60) | Very low |

### (3) Monovsion vs Monofocal

**Visual acuity**

| No. of studies | Design | Risk of bias | Inconsistency | Indirectness | Imprecision | No. of participants | Effect size (95% CI) | Quality |
|---|---|---|---|---|---|---|---|---|
| Mean uncorrected distance visual acuity (logMAR) (lower favours monovision) | | | | | | | | |
| 1 (Skov 2026) | RCT | Serious | N/A | Not serious | Serious | 137 | MD 0.08 (0.04 to 0.11) | Low |
| Mean uncorrected intermediate visual acuity (logMAR) (lower favours monovision) | | | | | | | | |
| 1 (Skov 2026) | RCT | Serious | N/A | Not serious | Serious | 137 | MD −0.07 (−0.11 to −0.02) | Low |
| Mean uncorrected near visual acuity (logMAR) (lower favours monovision) | | | | | | | | |
| 1 (Skov 2026) | RCT | Serious | N/A | Not serious | Not serious | 137 | MD −0.22 (−0.27 to −0.16) | Moderate |

**Visual function**

| No. of studies | Design | Risk of bias | Inconsistency | Indirectness | Imprecision | No. of participants | Effect size (95% CI) | Quality |
|---|---|---|---|---|---|---|---|---|
| Contrast sensitivity (Pelli–Robson logCS) (higher favours monovision) | | | | | | | | |
| 1 (Skov 2026) | RCT | Serious | N/A | Not serious | Not serious | 137 | MD −0.01 (−0.06 to 0.04) | Moderate |
| Patient satisfaction (higher favours monovision) | | | | | | | | |
| 1 (Skov 2026) | RCT | Serious | N/A | Not serious | Very serious | 137 | RR 1.00 (0.79 to 1.26) | Very low |
| Spectacle dependence (any) (lower favours monovision) | | | | | | | | |
| 1 (Skov 2026) | RCT | Serious | N/A | Not serious | Not serious | 137 | RR 0.72 (0.62 to 0.84) | Moderate |
| Visual function (Catquest-7SF) (higher favours monovision) | | | | | | | | |
| 1 (Skov 2026) | RCT | Serious | N/A | Not serious | Not serious | 137 | MD −0.01 (−0.26 to 0.23) | Moderate |

**Adverse events**

| No. of studies | Design | Risk of bias | Inconsistency | Indirectness | Imprecision | No. of participants | Effect size (95% CI) | Quality |
|---|---|---|---|---|---|---|---|---|
| Glare (lower favours monovision) | | | | | | | | |
| 1 (Skov 2026) | RCT | Serious | N/A | Not serious | Very serious | 137 | RR 0.98 (0.74 to 1.30) | Very low |
| Halos (lower favours monovision) | | | | | | | | |
| 1 (Skov 2026) | RCT | Serious | N/A | Not serious | Serious | 137 | RR 1.25 (0.78 to 2.01) | Low |
| Mean aniseikonia, % (lower favours monovision) | | | | | | | | |
| 1 (Skov 2026) | RCT | Serious | N/A | Not serious | Not serious* | 137 | MD 0.66 (0.11 to 1.22) | Moderate * |
| Starbursts (lower favours monovision) | | | | | | | | |
| 1 (Skov 2026) | RCT | Serious | N/A | Not serious | Very serious | 137 | RR 0.97 (0.64 to 1.49) | Very low |

### (4) Multifocal vs EDOF

**Visual acuity**

| No. of studies | Design | Risk of bias | Inconsistency | Indirectness | Imprecision | No. of participants | Effect size (95% CI) | Quality |
|---|---|---|---|---|---|---|---|---|
| Corrected distance visual acuity worse than 6/6 (lower favours multifocal) | | | | | | | | |
| 1 (Cochener 2018) | RCT | Serious | N/A | Not serious | Serious | 118 | RR 0.90 (0.28, 2.89) | Low |
| Mean corrected distance visual acuity (logMAR) (lower favours multifocal) | | | | | | | | |
| 2 (Akkaya Turhan 2021; Gil 2020) | RCT | Serious | Not serious | Not serious | Not serious | 145 | MD 0.03 (-0.22, 0.28) | Moderate |
| Mean corrected intermediate visual acuity (logMAR) (lower favours multifocal) | | | | | | | | |
| 2 (Akkaya Turhan 2021; Gil 2020) | RCT | Serious | Not serious | Not serious | Serious | 145 | MD 0.12 (-0.09, 0.34) | Low |
| Mean corrected near visual acuity (logMAR) (lower favours multifocal) | | | | | | | | |
| 2 (Akkaya Turhan 2021; Gil 2020) | RCT | Serious | Very serious | Not serious | Very serious | 145 | MD -0.02 (-1.20, 1.15) | Very low |
| Mean uncorrected distance visual acuity (decimal) (higher favours multifocal) | | | | | | | | |
| 1 (Cochener 2018) | RCT | Serious | N/A | Not serious | Not serious | 118 | MD 0.02 (-0.05, 0.10) | Moderate |
| Mean uncorrected distance visual acuity (logMAR) (lower favours multifocal) | | | | | | | | |
| 3 (Sevik 2022; | RCT | Serious | Not serious | Not serious | Not serious | 116 | MD -0.03 (-0.08, 0.03) | Moderate |
| Webers 2020; Labiris 2024) | | | | | | | | |
| Mean uncorrected intermediate visual acuity (logMAR) (lower favours multifocal) | | | | | | | | |
| 3 (Sevik 2022; Webers 2020; Labiris 2024) | RCT | Serious | Not serious | Not serious | Not serious | 116 | MD 0.03 (-0.04, 0.10) | Moderate |
| Mean uncorrected intermediate visual acuity (decimal) (higher favours multifocal) | | | | | | | | |
| 1 (Cochener 2018) | RCT | Serious | N/A | Not serious | Not serious | 118 | MD -0.05 (-0.11, 0.02) | Moderate |
| Mean uncorrected near visual acuity (logMAR) (lower favours multifocal) | | | | | | | | |
| 3 (Sevik 2022; Webers 2020; Labiris 2024) | RCT | Serious | Very serious | Not serious | Serious | 116 | MD -0.03 (-0.27, 0.20) | Very low |
| Mean uncorrected near visual acuity (decimal) (higher favours multifocal) | | | | | | | | |
| 1 (Cochener 2018) | RCT | Serious | N/A | Not serious | Serious | 118 | MD 0.09 (0.03, 0.16) | Low |
| Uncorrected distance visual acuity worse than 6/6 (lower favours multifocal) | | | | | | | | |
| 1 (Cochener 2018) | RCT | Serious | N/A | Not serious | Serious | 147 | RR 1.19 (0.76, 1.85) | Low |
| Uncorrected near visual acuity worse than J3/J4 or equivalent (lower favours multifocal) | | | | | | | | |
| 1 (Cochener 2018) | RCT | Serious | N/A | Not serious | Serious | 118 | RR 0.34 (0.15, 0.77) | Low |

**Visual function**

| No. of studies | Design | Risk of bias | Inconsistency | Indirectness | Imprecision | No. of participants | Effect size (95% CI) | Quality |
|---|---|---|---|---|---|---|---|---|
| Maximum reading speed, words/min (higher favours multifocal) | | | | | | | | |
| 1 (Akkaya Turhan 2021) | RCT | Serious | N/A | Not serious | Very serious | 29 | MD -5.00 (-26.87, 16.87) | Very low |
| Reading acuity (logMAR) (lower favours multifocal) | | | | | | | | |
| 1 (Akkaya Turhan 2021) | RCT | Serious | N/A | Not serious* | Not serious | 29 | MD 0.01 (-0.04, 0.06) | Moderate |
| Mean stereoacuity, arcseconds (lower favours multifocal) | | | | | | | | |
| 1 (Pradeep 2026) | RCT | Serious | N/A | Not serious* | Serious | 70 | MD 29.14 (7.27, 51.01) | Low |
| Visual function (unspecified instruments) (higher favours multifocal) | | | | | | | | |
| 3 (Sevik 2022; Webers 2020; Labiris 2024) | RCT | Serious | Not serious | Not serious | Very serious | 116 | SMD 0.00 (-0.80, 0.80) | Very low |
| Patient satisfaction (score, 0–100 scale) (higher favours multifocal) | | | | | | | | |
| 1 (Sevik 2022) | RCT | Serious | N/A | Not serious | Very serious | 29 | MD -1.71 (-10.67, 7.25) | Very low |
| Patient satisfaction (binary) (higher favours multifocal) | | | | | | | | |
| 1 (Webers 2020) | RCT | Serious | N/A | Not serious | Serious | 27 | RR 0.99 (0.80, 1.23) | Low |

**Adverse outcomes**

| No. of studies | Design | Risk of bias | Inconsistency | Indirectness | Imprecision | No. of participants | Effect size (95% CI) | Quality |
|---|---|---|---|---|---|---|---|---|
| Glare (lower favours multifocal) | | | | | | | | |
| 2 (Webers 2020; Gil 2022) | RCT | Serious | Not serious | Not serious | Serious | 137 | RR 1.16 (0.13, 10.63) | Low |
| Halos (lower favours multifocal) | | | | | | | | |
| 2 (Webers 2020; Gil 2022) | RCT | Serious | Serious | Not serious | Very serious | 138 | RR 1.05 (0.02, 64.58) | Very low |
| Overall photic phenomena (lower favours multifocal) | | | | | | | | |
| 1 (Sevik 2022) | RCT | Serious | N/A | Not serious | Very serious | 29 | RR 4.69 (0.24, 89.88) | Very low |

## (5) Multifocal vs Monovision

**Visual acuity**

| No. of studies | Design | Risk of bias | Inconsistency | Indirectness | Imprecision | No. of participants | Effect size (95% CI) | Quality |
|---|---|---|---|---|---|---|---|---|
| Mean uncorrected distance visual acuity (logMAR) (lower favours multifocal) | | | | | | | | |
| 2 (Labiris 2024; Wilkins 2013) | RCT | Serious | Not serious | Not serious | Not serious | 246 | MD 0.04 (-0.15, 0.23) | Moderate |
| Mean uncorrected intermediate visual acuity (logMAR) (lower favours multifocal) | | | | | | | | |
| 2 (Labiris 2024; Wilkins 2013) | RCT | Serious | Not serious | Not serious | Not serious | 241 | MD 0.06 (-0.14, 0.26) | Moderate |
| Mean uncorrected near visual acuity (logMAR) (lower favours multifocal) | | | | | | | | |
| 2 (Labiris 2024; Wilkins 2013) | RCT | Serious | Serious | Not serious | Not serious | 246 | MD −0.00 (−0.57, 0.56) | Low |

**Visual function**

| No. of studies | Design | Risk of bias | Inconsistency | Indirectness | Imprecision | No. of participants | Effect size (95% CI) | Quality |
|---|---|---|---|---|---|---|---|---|
| Visual function (lower favours multifocal) | | | | | | | | |
| 1 (Labiris 2024) | RCT | Serious | N/A | Serious | Serious | 60 | MD −3.81 (−8.64, 1.02) | Very low |
| Visual function – VF-14 (higher favours multifocal) | | | | | | | | |
| 1 (Labiris 2015) | RCT | Serious | N/A | Not serious | Not serious | 75 | MD −1.47 (−5.51, 2.57) | Moderate |
| Spectacle dependence – any (lower favours multifocal) | | | | | | | | |
| 2 (Wilkins 2013; Labiris 2015) | RCT | Serious | Not serious | Not serious | Not serious | 262 | RR 0.40 (0.06, 2.59) | Moderate |
| Contrast sensitivity – Pelli-Robson test (higher favours multifocal) | | | | | | | | |
| 2 (Wilkins 2013; Labiris 2015) | RCT | Serious | Not serious | Not serious | Not serious | 262 | MD −0.03 (−0.47, 0.41) | Moderate |

**Adverse events**

| No. of studies | Design | Risk of bias | Inconsistency | Indirectness | Imprecision | No. of participants | Effect size (95% CI) | Quality |
|---|---|---|---|---|---|---|---|---|
| Glare (lower favours multifocal) | | | | | | | | |
| 1 (Wilkins 2013) | RCT | Serious | N/A | Not serious | Not serious | 187 | RR 1.41 (1.14, 1.73) | Moderate |

### (6) EDOF vs Monovision

#### Visual acuity

| No. of studies | Design | Risk of bias | Inconsistency | Indirectness | Imprecision | No. of participants | Effect size (95% CI) | Quality |
|---|---|---|---|---|---|---|---|---|
| Corrected distance visual acuity worse than 6/6 (lower favours EDOF) | | | | | | | | |
| 1 (Vandekerckhove 2026) | RCT | Serious | N/A | Not serious | Serious | 105 | RR 5.22 (1.07, 25.55) | Low |
| Mean corrected distance visual acuity (logMAR) (lower favours EDOF) | | | | | | | | |
| 2 (Vandekerckhove 2026; Zeilinger 2025) | RCT | Serious | Not serious | Not serious | Serious | 185 | MD 0.04 (−0.15, 0.23) | Low |
| Mean corrected intermediate visual acuity (logMAR) (lower favours EDOF) | | | | | | | | |
| 2 (Vandekerckhove 2026; Zeilinger 2025) | RCT | Serious | Not serious | Not serious | Serious | 185 | MD −0.10 (−0.33, 0.12) | Low |
| Mean corrected near visual acuity (logMAR) (lower favours EDOF) | | | | | | | | |
| 2 (Vandekerckhove 2026; Zeilinger 2025) | RCT | Serious | Serious | Not serious | Serious | 185 | MD −0.09 (−0.95, 0.77) | Very low |
| Mean stereoacuity (lower favours EDOF) | | | | | | | | |
| 1 (Skov 2026) | RCT | Serious | N/A | Not serious | Serious | 131 | MD 10.39 (−34.01, 54.78) | Low |
| Mean uncorrected distance visual acuity (logMAR) (lower favours EDOF) | | | | | | | | |
| 5 (Vandekerckhove 2026; Zeilinger 2025; Sandoval 2020; Labiris 2024; Skov 2026) | RCT | Serious | Not serious | Not serious | Not serious | 452 | MD 0.03 (−0.02, 0.08) | Moderate |
| Mean uncorrected intermediate visual acuity (logMAR) (lower favours EDOF) | | | | | | | | |
| 5 (Vandekerckhove 2026; Zeilinger 2025; Sandoval 2020; Labiris 2024; Skov 2026) | RCT | Serious | Not serious | Not serious | Not serious | 452 | MD −0.01 (−0.06, 0.04) | Moderate |
| Mean uncorrected near visual acuity (logMAR) (lower favours EDOF) | | | | | | | | |
| 5 (Vandekerckhove 2026; Zeilinger 2025; Sandoval 2020; Labiris | RCT | Serious | Very serious | Not serious | Very serious | 452 | MD 0.01 (−0.17, 0.19) | Very low |
| 2024; Skov 2026) | | | | | | | | |
| Uncorrected distance visual acuity worse than 6/6 (lower favours EDOF) | | | | | | | | |
| 1 (Vandekerckhove 2026) | RCT | Serious | N/A | Not serious | Serious | 105 | RR 1.70 (0.92, 3.11) | Low |
| Uncorrected near visual acuity worse than J3/J4 or equivalent (lower favours EDOF) | | | | | | | | |
| 1 (Vandekerckhove 2026) | RCT | Serious | N/A | Not serious | Serious | 105 | RR 0.25 (0.08, 0.77) | Low |

**Visual function**

| No. of studies | Design | Risk of bias | Inconsistency | Indirectness | Imprecision | No. of participants | Effect size (95% CI) | Quality |
|---|---|---|---|---|---|---|---|---|
| Visual function (VF-14) (higher favours EDOF) | | | | | | | | |
| 1 (Labiris 2024) | RCT | Serious | N/A | Not serious | Not serious | 60 | MD −5.59 (−10.20, −0.98) | Moderate |
| Patient satisfaction (higher favours EDOF) | | | | | | | | |
| 1 (Skov 2026) | RCT | Serious | N/A | Not serious | Serious | 131 | RR 0.97 (0.76, 1.25) | Low |
| Spectacle dependence (any) (lower favours EDOF) | | | | | | | | |
| 2 (Vandekerckhove 2026; Skov 2026) | RCT | Serious | Not serious | Not serious | Serious | 312 | RR 1.07 (0.40, 2.87) | Low |
| Contrast sensitivity (higher favours EDOF) | | | | | | | | |
| 1 (Skov 2026) | RCT | Serious | N/A | Not serious | Not serious | 131 | MD 0.001 (−0.044, 0.045) | Moderate |

**Adverse outcomes**

| No. of studies | Design | Risk of bias | Inconsistency | Indirectness | Imprecision | No. of participants | Effect size (95% CI) | Quality |
|---|---|---|---|---|---|---|---|---|
| Glare (lower favours EDOF) | | | | | | | | |
| 1 (Skov 2026) | RCT | Serious | N/A | Not serious | Very serious | 131 | RR 1.08 (0.82, 1.43) | Very low |
| Halos (lower favours EDOF) | | | | | | | | |
| 1 (Skov 2026) | RCT | Serious | N/A | Not serious | Very serious | 131 | RR 0.91 (0.57, 1.44) | Very low |
| Starbursts (lower favours EDOF) | | | | | | | | |
| 1 (Skov 2026) | RCT | Serious | N/A | Not serious | Serious | 131 | RR 1.29 (0.87, 1.90) | Low |

## (7) Trifocal multifocal IOL vs Bifocal multifocal IOL

**Visual acuity**

| No of studies | Design | Risk of bias | Inconsistency | Indirectness | Imprecision | No of participants | Effect size (95% CI) | Quality |
|---|---|---|---|---|---|---|---|---|
| Mean corrected distance visual acuity (lower favours trifocal) | | | | | | | | |
| 3 (Li 2023, Gundersen 2016, Jonker 2015) | RCT | Serious | Not serious | Not serious | Not serious | 149 | MD 0.00 (-0.07, 0.08) | Moderate |
| Mean corrected intermediate visual acuity (lower favours trifocal) | | | | | | | | |
| 2 (Li 2023, Jonker 2015) | RCT | Serious | Not serious | Not serious | Serious | 127 | MD -0.03 (-0.30, 0.24) | Low |
| Mean corrected near visual acuity (lower favours trifocal) | | | | | | | | |
| 2 (Li 2023, Jonker 2015) | RCT | Serious | Not serious | Not serious | Not serious | 127 | MD 0.01 (-0.26, 0.29) | Moderate |
| Mean uncorrected distance visual acuity (logMAR) (lower favours trifocal) | | | | | | | | |
| 3 (Li 2023, Gundersen 2016, Jonker 2015) | RCT | Serious | Not serious | Not serious | Not serious | 149 | MD -0.01 (-0.10, 0.07) | Moderate |
| Mean uncorrected near visual acuity (lower favours trifocal) | | | | | | | | |
| 2 (Li 2023, Jonker 2015) | RCT | Serious | Not serious | Not serious | Serious | 127 | MD -0.00 (-0.41, 0.41) | Low |

**Visual function**

| No of studies | Design | Risk of bias | Inconsistency | Indirectness | Imprecision | No of participants | Effect size (95% CI) | Quality |
|---|---|---|---|---|---|---|---|---|
| Spectacle dependence – near (lower favours trifocal) | | | | | | | | |
| 1 (Jonker 2015) | RCT | Serious | N/A | Not serious | Very serious | 28 | RR 0.65 (0.18, 2.38) | Very low |

Note: The 6 quality indirectness grading and corresponding overall evidence certainty level marked with asterisks were determined by human experts.

**Supplementary Information 5. Evidence to recommendations (prospective study, GUIDE-assisted group)**

| | |
|---|---|
| **Relative value of different outcomes** | The committee agreed that the choice of intraocular lens should be assessed using outcomes that reflect the overall experience of vision rather than any single measure of visual acuity. Patient-reported visual function, quality of life and satisfaction were considered particularly important because they integrate the ability to perform everyday activities, dependence on spectacles and the acceptability of unwanted visual symptoms. The need for additional treatment, serious ocular complications and the financial consequences for patients and healthcare services were also considered important.<br>The committee recognised that the relative importance of these outcomes varies between stakeholders. Patients may place different values on spectacle independence, night-time visual quality and expenditure; families and caregivers may value the patient's functional independence; clinicians may prioritise predictable outcomes and the avoidance of difficult postoperative adaptation; and healthcare providers must also consider affordability, equity and the consequences of additional follow-up or treatment. |
| **Trade-off between benefits and harms** | The committee agreed that none of the evaluated options was consistently superior across all relevant outcomes. Each option represents a different balance between range of vision, optical quality, unwanted visual phenomena and predictability of the postoperative result.<br>Monofocal lenses generally provide predictable distance vision and are associated with fewer concerns about glare, halos and other optical disturbances. Their principal disadvantage is that patients are more likely to require spectacles for intermediate and near activities. Multifocal lenses generally offer the greatest potential for near vision and spectacle independence. These benefits must be balanced against a greater likelihood of halos, starbursts or other unwanted visual phenomena and, in some patients, a small reduction in the clarity of distance vision. They may remain troublesome for people who frequently drive at night, work in low-contrast environments or have particularly high expectations of optical quality. EDOF lenses may provide a broader and more continuous range of vision than monofocal lenses, particularly for intermediate activities, while generally retaining useful distance vision. However, they do not guarantee spectacle independence for near tasks, and |
| | the available direct evidence does not consistently demonstrate that they produce fewer visual disturbances than multifocal lenses. Monovision may improve functional intermediate and near vision without using presbyopia-correcting optics. However, it introduces a deliberate difference between the two eyes and may adversely affect binocular balance, depth perception or distance visual quality. Tolerance is individual and may be more predictable in people who have previously used and accepted monovision. The committee considered that monovision should be viewed as a refractive strategy requiring careful patient selection rather than as a direct equivalent of a particular lens design.<br>The head-to-head evidence did not identify a clear overall hierarchy among different strategies. Similarly, the evidence comparing trifocal and bifocal lenses did not demonstrate a sufficiently consistent overall advantage to support one design for all patients. The committee emphasised that the absence of a statistically significant difference should not be interpreted as evidence that the options are equivalent, particularly where studies were small or imprecise. The balance of benefits and harms is therefore strongly preference sensitive.<br>Overall, the committee considered that the evidence supports individualized selection and informed shared decision-making rather than routine preference for a single option. Patients should understand that improved range of vision and reduced spectacle dependence may be obtained at the cost of optical trade-offs, and that no option can guarantee complete spectacle independence or freedom from unwanted visual symptoms. |
| **Consideration of health benefits and resource use** | The committee noted that the acquisition cost of the lens, the amount reimbursed by a payer and the amount ultimately paid by the patient are different quantities. Nevertheless, they illustrate that lenses intended to reduce presbyopia and spectacle dependence commonly remain substantially more expensive than standard monofocal lenses, and that the additional cost may fall on the healthcare system or the individual patient depending on the country.<br>In China, national volume-based procurement has substantially reduced the acquisition prices of artificial lenses. The National Healthcare Security Administration reported an average reduction of approximately 60% in the 2023 national procurement round. The principal economic benefit of multifocal and other presbyopia-correcting lenses is the potential reduction in future spectacle use, together with convenience and possible gains in functional |
| | independence. These benefits accrue over time, whereas the additional lens or procedure cost is generally incurred at the time of surgery. The value of the additional expenditure is therefore likely to be greater for patients who place a high value on spectacle independence, have a longer expected duration of benefit and are willing to accept the associated visual trade-offs.<br>Published economic models illustrate this uncertainty. A United States patient-perspective model estimated that bilateral trifocal implantation produced 0.67 additional QALYs at an incremental lifetime patient cost of US$2,783, while reducing lifetime spectacle expenditure by US$2,593. However, the results were highly sensitive to the assumed burden of wearing glasses, patient out-of-pocket costs and postoperative spectacle dependence. A South Korean model estimated an additional 0.60 QALYs at an incremental patient cost of US$4,274, but the estimated net benefit decreased substantially when assumptions about spectacle use or willingness to pay were changed. Both studies were sponsored by the manufacturer of the evaluated lenses and were based on modelling assumptions, their conclusions should therefore be interpreted cautiously. Where advanced lenses are predominantly self-funded, routine reliance on ability to pay may increase inequity in access. Procurement-related price reductions improve affordability but do not, by themselves, establish that the additional benefits justify routine use for all patients.<br>The committee therefore considered that the price premium should be assessed against the benefits that are important to the individual patient and against local opportunity costs. The present evidence does not support assuming that a more expensive lens necessarily provides proportionately greater overall health benefit. |
| **Quality of evidence** | The committee noted that the certainty of the evidence was predominantly low, with moderate-certainty evidence available for only selected comparisons and outcomes. Important limitations included risk of bias, imprecision, small studies and limited direct comparisons between newer lens designs. Evidence on patient-reported quality of life, dissatisfaction, serious complications and the need for additional treatment was less complete than evidence on visual acuity or spectacle use.<br>The results may also be influenced by differences in lens generation, surgical technique, |
| | refractive targets, patient selection and outcome measurement. Consequently, the evidence is sufficient to describe the main trade-offs but not to establish a robust overall ranking of all available options. |
| **Other considerations** | The committee considered that acceptability would depend heavily on the quality of preoperative counselling and whether the postoperative result matched the patient's expectations. Patients should receive balanced information about the likely range of vision, possible continued need for spectacles, unwanted visual phenomena, uncertainty in the evidence and the complete personal cost of each option.<br>Implementation would require appropriate patient selection, access to reliable biometry, clinicians experienced with the relevant refractive strategies and systems for monitoring dissatisfaction, additional procedures and lens exchange. Providers should avoid presenting premium lenses as inherently superior and should ensure that financial incentives do not replace preference-sensitive shared decision-making.<br>Further research should prioritise adequately powered direct comparisons using standardised patient-reported outcomes, longer follow-up and consistent reporting of visual symptoms, additional treatment and costs. |

**Supplementary Information 6. Recommendations (prospective study, GUIDE-assisted group)**

| Question: What is the optimal strategy to facilitate simultaneous distance and near vision following cataract surgery? | |
|---|---|
| **Recommendations in 2017 NICE guideline (NG77: Cataracts in adults:management)** | **Recommendations updated by GUIDE** |
| Do not offer multifocal intraocular lenses for people having cataract surgery.<br>Offer monovision for use after cataract surgery to people who have either anisometropia or monovision preoperatively and would like to remain with it. | Consider multifocal intraocular lenses for people who want greater spectacle independence for distance, intermediate and near vision. When discussing this option:<br>● ensure the person's ocular health is suitable for multifocal intraocular lenses<br>● explain the potential benefits in uncorrected near and intermediate vision compared with standard monofocal intraocular lenses<br>● discuss the increased risk of visual disturbances (such as halos, glare and starbursts)<br>● take into account the person's visual needs and daily activities<br>● consider cost implications and the person's willingness to pay |

**Supplementary Information 7. PICO criteria (prospective study, human-only group)**

| | Generation by human expert |
|---|---|
| **Population** | Adults (18 years and older) undergoing phacoemulsification cataract surgery and intraocular lens (IOL) implantation in one or both eyes |
| **Interventions** | ● Non-accommodating intraocular lens designed to support both near and distance vision (e.g., multifocal lenses, extended-depth-of-focus lenses or monovision correction strategy)<br>● Standard monofocal intraocular lenses |
| **Comparators** | All listed interventions compared with one another |
| **Outcomes** | ● Distance, intermediate, and near visual acuity<br>● Patient satisfaction<br>● Spectacle independence<br>● Incidence of visual disturbances<br>● Quality of life<br>● Resource use and costs |

**Supplementary Information 8. Summary of included studies (prospective study, human-only group)**

| Study & location | Sample size | Methods |
|---|---|---|
| Cochener (2018)<br>France | 120 eyes<br>(60 patients) | Comparison of AcrySof IQ PanOptix and FineVision Micro F trifocal IOLs with the TECNIS Symfony EDOF IOL. |
| Gil (2020)<br>Spain | 232 eyes<br>(116 patients) | Comparison of AcrySof ReSTOR SV25T0, AT LISA 809M, TECNIS ZKB00 and ZLB00 bifocal IOLs and the AT LISA tri 839MP trifocal IOL with the TECNIS Symfony ZXR00 EDOF IOL. |
| Law (2020)<br>United Kingdom | 180 eyes<br>(90 patients) | Comparison of the Bi-Flex 677MY apodized diffractive multifocal IOL with the Bi-Flex 677AB aspheric aberration-neutral monofocal IOL. |
| Sandoval (2020)<br>USA | 152 eyes<br>(76 patients) | Comparison of bilateral TECNIS Symfony toric EDOF IOLs targeted for binocular emmetropia or mini-monovision. |
| Webers (2020)<br>The Netherlands | 54 eyes<br>(27 patients) | Comparison of the TECNIS Symfony (ZXR00) EDOF IOL using micromonovision with the AT LISA tri 839MP trifocal IOL. |
| Messias (2021)<br>Brazil | 66 eyes<br>(33 patients) | Comparison of the AcrySof ReSTOR SN6AD1 apodized diffractive multifocal IOL with the AcrySof IQ SN60WF aspheric monofocal IOL. |
| Reinhard (2021)<br>Germany, Belgium, Portugal, Spain, France, and Italy | 422 eyes<br>(211 patients) | Comparison of the AT LARA 829MP and TECNIS Symfony EDOF IOLs with the CT ASPHINA 409MP monofocal IOL. |
| Turhan (2021) / Sevik (2022)<br>Turkey | 58 eyes<br>(29 patients) | Comparison of the TECNIS +2.75 D (ZKB00) low-add bifocal IOL with the TECNIS Symfony (ZXR00) EDOF IOL, both using mini-monovision. |
| Bala (2022)<br>Australia, Canada, Spain, and the United Kingdom | 540 eyes<br>(270 patients) | Comparison of the AcrySof IQ Vivity (DFT015) non-diffractive EDOF IOL with the AcrySof IQ SN60WF aspheric monofocal IOL. |
| Chang (2022)<br>USA | 597 eyes<br>(299 patients) | Comparison of bilateral TECNIS Symfony (ZXR00) diffractive EDOF IOLs with TECNIS 1-piece (ZCB00) aspheric monofocal IOLs, both targeted for emmetropia. |
| Guarro (2022)<br>Spain | 176 eyes<br>(88 patients) | Comparison of the AcrySof IQ Vivity non-diffractive EDOF, AT LARA 829MP and TECNIS Symfony ZXR00 diffractive EDOF IOLs with the AcrySof IQ SN60WF monofocal IOL. |
| McCabe (2022)<br>USA | 435 eyes<br>(218 patients) | Comparison of the AcrySof IQ Vivity (DFT015) non-diffractive EDOF IOL with the AcrySof IQ SN60WF aspheric monofocal IOL. |
| Chang (2023)<br>USA | 544 eyes<br>(272 patients) | Comparison of the TECNIS Synergy (ZFR00V) hybrid multifocal-EDOF IOL with the TECNIS ZCB00 monofocal IOL. |
| Li (2023)<br>China | 99 eyes<br>(99 patients) | Comparison of the AT LISA tri toric 939M/MP trifocal IOL with the AT LISA toric 909M bifocal IOL. |
| Pantanelli (2023)<br>USA | 102 eyes<br>(51 patients) | Comparison of the AcrySof Vivity (DFT015) non-diffractive EDOF IOL with the enVista (MX60E) neutral aspheric monofocal IOL, both targeted for emmetropia. |
| Corbett / Black (2024)<br>Australia and New Zealand | 235 eyes<br>(118 patients) | Comparison of the TECNIS PureSee (ZEN00V) fully refractive EDOF IOL with the TECNIS Eyhance (ICB00) enhanced monofocal IOL. |
| Hida (2024)<br>Brazil | 116 eyes<br>(58 patients) | Comparison of bilateral TECNIS Synergy (ZFR00V) hybrid multifocal-EDOF IOLs with blended TECNIS |
| | | Eyhance (ICB00) enhanced monofocal and TECNIS Synergy implantation. |
| Labiris (2024)<br>Greece | 180 eyes<br>(90 patients) | Comparison of mixed AcrySof IQ PanOptix trifocal and AcrySof IQ Vivity non-diffractive EDOF implantation with bilateral PanOptix or bilateral Vivity implantation. |
| Zeilinger (2025)<br>Austria | 80 eyes<br>(40 patients) | Comparison of the RayOne EMV (RAO200E) advanced monofocal IOL with the AcrySof IQ Vivity (DFT015) non-diffractive EDOF IOL, both using mini-monovision. |
| Palkovits (2026)<br>Austria | 70 eyes<br>(35patients) | Comparison of the Acunex Vario AN6V refractive EDOF IOL with the Acunex AN6 monofocal IOL, both targeted at -0.25 D. |
| Pradeep (2026)<br>India | 140 eyes<br>(70 patients) | Comparison of bilateral AcrySof IQ PanOptix trifocal IOLs with bilateral AcrySof IQ Vivity non-diffractive EDOF IOLs. |
| Skov (2026)<br>Denmark | 360 eyes<br>(180 patients) | Comparison of emmetropia, mini-monovision, and monovision using bilateral AcrySof ACU0T0 or TECNIS GCB00/ZCB00 monofocal IOLs with bilateral AcrySof IQ Vivity non-diffractive EDOF IOLs. |
| Vandekerckhove (2026)<br>Switzerland | 210 eyes<br>(105 patients) | Comparison of Isopure and TECNIS Eyhance enhanced monofocal IOLs with the AcrySof IQ Vivity EDOF IOL using mini-monovision. |

**Supplementary Information 9. Excluded studies and reasons (prospective study, human-only group)**

| Study | Reason for exclusion |
|---|---|
| Hayashi K, Yoshida M, Sasaki H, Hirata A. Binocular visual function of myopic pseudophakic monovision. Japanese Journal of Ophthalmology. 2018;62:357–364. | Intervention did not meet the prespecified pseudophakic monovision definition. |
| Schojai M, Schultz T, Jerke C, Böcker J, Dick HB. Visual performance comparison of 2 extended depth-of-focus intraocular lenses. Journal of Cataract & Refractive Surgery. 2020;46(3):388–393. | Both groups evaluated EDOF-based IOLs. |
| Xun Y, Wan W, Jiang L, Hu K. Crossed versus conventional pseudophakic monovision for high myopic eyes: a prospective, randomized pilot study. BMC Ophthalmology. 2020;20:447. | Included patients with high-myopic cataract. |
| Dick HB, Ang RE, Corbett D, Hoffmann P, Tetz M, Villarrubia A, Palomino C, Castillo-Gomez A, Tsai L, Thomas EK, Janakiraman P. Comparison of 3-month visual outcomes of a new multifocal intraocular lens vs a trifocal intraocular lens. Journal of Cataract & Refractive Surgery. 2022;48(11):1270–1276. | Cataract-only outcome data were not separately extractable. |
| Gil MÁ, Varón C, Cardona G, Buil JA. Far and Near Contrast Sensitivity and Quality of Vision with Six Presbyopia Correcting Intraocular Lenses. Journal of Clinical Medicine. 2022;11(14):4150. | Participants were not randomly allocated. |
| Fernandes MC, Nunomura CY, Messias A. Visual performance and photic disturbances with diffractive and nondiffractive EDOF intraocular lenses using mini-monovision: randomized trial. Journal of Cataract & Refractive Surgery. 2024;50(2):153–159. | Both groups received EDOF IOLs. |
| Vasavada V, Vasavada SA, Shastri L, Nath V, Vasavada AR, Srivastava S. Visual outcomes comparing emmetropia vs mini-monovision after bilateral implantation of a nondiffractive extended vision intraocular lens: randomized trial. Journal of | Cataract-only outcome data were not separately extractable. |
| Cataract & Refractive Surgery. 2024;50(8):799–804. | |
| Han D, Ma X, Gao J, Xu Z, Liu F. Comparison of clinical effects of aspherical monofocal and Tecnis ZMB00 multifocal intraocular lenses in patients with age-related cataract. Journal of the Pakistan Medical Association. 2025;75(8):1229–1234. | Intervention could not be reliably classified: the title identifies the TECNIS ZMB00, whereas the Methods report the multifocal IOL as model 52501TW manufactured by Ophtec. |
| Nagy ZZ, Dorman P, Szalczer S, Kiss H. Prospective Comparison of Clinical Outcomes After Bilateral Implantation of Diffractive Trifocal Extended Depth-of-Focus and Diffractive Trifocal Hydrophobic Intraocular Lenses. Journal of Refractive Surgery. 2025;41(2):e102–e113. | Cataract-only outcome data were not separately extractable. |
| Wanten JC, Bauer NJC, Berendschot TTJM, van den Biggelaar FJHM, Nuijts RMMA. Two novel extended depth-of-focus intraocular lenses targeted for mini-monovision: A prospective randomized controlled trial. American Journal of Ophthalmology. 2025;276:286–296. | Both groups received EDOF IOLs. |
| Wendelstein JA, Vallotton K, Ziörjen A, Müller M, Riaz KM, Pantanelli SM, Langenbucher A, Seiler TG. Reading Speed and Visual Acuity in Photopic and Mesopic Conditions After Bilateral Implantation of Diffractive Multifocal Intraocular Lenses. American Journal of Ophthalmology. 2026;281:516–525. | Both groups received diffractive multifocal IOLs. |

## Supplementary Information 10. GRADE tables (prospective study, human-only group)

### (1) Multifocal vs Monofocal

**Visual acuity**

| No. of studies | Design | Risk of bias | Inconsistency | Indirectness | Imprecision | No. of participants | Effect size (95% CI) | Quality |
|---|---|---|---|---|---|---|---|---|
| Corrected distance visual acuity worse than 6/6 (RR<1 favours multifocal) | | | | | | | | |
| 9 (elMaghraby 1992; Steinert 1992; Percival 1993; Rossetti 1994; Haaskjold 1998; Kamlesh 2001; Leyland 2002; Sen 2004; Chang 2023) | RCT | Serious | Serious | Not serious | Serious | 953 | RR 1.09 (0.75 to 1.58) | Very low |
| Uncorrected distance visual acuity worse than 6/6 (RR<1 favours multifocal) | | | | | | | | |
| 9 (elMaghraby 1992; Steinert 1992; Percival 1993; Rossetti 1994; Haaskjold 1998; Leyland 2002; Sen 2004; Jusufovic 2011; Chang 2023) | RCT | Serious | Serious | Not serious | Not serious | 943 | RR 1.01 (0.89 to 1.14) | Low |
| Uncorrected near visual acuity worse than J3/J4 or an equivalent threshold (RR<1 favours multifocal) | | | | | | | | |
| 9 (elMaghraby 1992; Percival 1993; Rossetti 1994; Haaskjold 1998; Javitt 2000; Leyland 2002; Jusufovic 2011; Ji 2013; Chang 2023) | RCT | Serious | Serious | Not serious | Not serious | 1,043 | RR 0.18 (0.08 to 0.41) | Low |
| Mean corrected distance visual acuity (logMAR) (lower favours multifocal) | | | | | | | | |
| 8 (Leyland 2002; Nijkamp 2004; Harman 2008; Palmer 2008; Rasp 2012; Peng 2012; Law 2020; Messias 2021) | RCT | Serious | Not serious | Not serious | Not serious | 1,062 | MD 0.03 (0.01 to 0.05) | Moderate |
| Mean uncorrected distance visual acuity (logMAR) (lower favours multifocal) | | | | | | | | |
| 9 (Leyland 2002; Nijkamp 2004; Palmer 2008; Harman 2008; Peng 2012; Rasp 2012; Law 2020; Messias 2021; Chang 2023) | RCT | Serious | Not serious | Not serious | Not serious | 1,265 | MD 0.02 (−0.01 to 0.05) | Moderate |
| Mean corrected intermediate visual acuity (logMAR) (lower favours multifocal) | | | | | | | | |
| 4 (Peng 2012; Maxwell 2017; Law 2020; Chang 2023) | RCT | Serious | Serious | Not serious | Serious | 866 | MD −0.12 (−0.29 to 0.05) | Very low |
| Mean uncorrected intermediate visual acuity (logMAR) (lower favours multifocal) | | | | | | | | |
| 3 (Peng 2012; Maxwell | RCT | Serious | Serious | Not serious | Serious | 776 | MD −0.11 (−0.27 to | Very low |
| 2017; Chang 2023) | | | | | | | 0.05) | |
| Mean corrected near visual acuity (logMAR) (lower favours multifocal) | | | | | | | | |
| 9 (Javitt 2000; Leyland 2002; Palmer 2008; Harman 2008; Rasp 2012; Peng 2012; Maxwell 2017; Law 2020; Chang 2023) | RCT | Serious | Serious | Not serious | Not serious | 1,667 | MD −0.15 (−0.34 to 0.05) | Low |
| Mean uncorrected near visual acuity (logMAR) (lower favours multifocal) | | | | | | | | |
| 9 (Javitt 2000; Leyland 2002; Harman 2008; Peng 2012; Rasp 2012; Maxwell 2017; Law 2020; Messias 2021; Chang 2023) | RCT | Serious | Serious | Not serious | Not serious | 1,559 | MD −0.25 (−0.39 to −0.12) | Low |

**Visual function**

| No. of studies | Design | Risk of bias | Inconsistency | Indirectness | Imprecision | No. of participants | Effect size (95% CI) | Quality |
|---|---|---|---|---|---|---|---|---|
| Contrast sensitivity (higher favours multifocal) | | | | | | | | |
| 4 (Rossetti 1994; Leyland 2002; Sen 2004; Harman 2008) | RCT | Serious | Not serious | Not serious | Serious | 288 | SMD −0.19 (−0.57 to 0.19) | Low |
| Visual function (higher favours multifocal) | | | | | | | | |
| 5 (Nijkamp 2004; Sen 2004; Cillino 2008; Zhao 2010; Law 2020) | RCT | Serious | Serious | Not serious | Serious | 563 | SMD 0.79 (−0.36 to 1.94) | Very low |
| Spectacle dependence (any) (RR<1 favours multifocal) | | | | | | | | |
| 13 (Steinert 1992; Percival 1993; Rossetti 1994; Haaskjold 1998; Javitt 2000; Leyland 2002; Harman 2008; Cillino 2008; Zhao 2010; Peng 2012; Maxwell 2017; Law 2020; Chang 2023) | RCT | Serious | Serious | Not serious | Not serious | 1,647 | RR 0.54 (0.40 to 0.74) | Low |
| Patient satisfaction (number of satisfied participants) (RR>1 favours multifocal) | | | | | | | | |
| 3 (Law 2020; Chang 2023; Hida 2024) | RCT | Serious | Serious | Not serious | Serious | 403 | RR 2.41 (0.47 to 12.46) | Very low |
| Patient satisfaction (questionnaire score) (higher favors multifocal) | | | | | | | | |
| 6 (Steinert 1992; Nijkamp 2004; Sen 2004; | RCT | Serious | Very serious | Not serious | Very serious | 643 | SMD 0.26 (−0.37 to 0.89) | Very low |
| Cillino 2008; Zhao 2010; Peng 2012) | | | | | | | | |

**Adverse outcomes**

| No. of studies | Design | Risk of bias | Inconsistency | Indirectness | Imprecision | No. of participants | Effect size (95% CI) | Quality |
|---|---|---|---|---|---|---|---|---|
| Glare (RR<1 favours multifocal) | | | | | | | | |
| 12 (Percival 1993; Rossetti 1994; Haaskjold 1998; Kamlesh 2001; Sen 2004; Cillino 2008; Harman 2008; Maxwell 2017; Chang 2023; Hida 2024) | RCT | Serious | Serious | Not serious | Serious | 1,153 | RR 1.34 (1.06 to 1.83) | Very low |
| Halos (RR<1 favours multifocal) | | | | | | | | |
| 11 (Percival 1993; Rossetti 1994; Haaskjold 1998; Kamlesh 2001; Sen 2004; Cillino 2008; Zhao 2010; Maxwell 2017; Law 2020; Chang 2023; Hida 2024) | RCT | Serious | Serious | Not serious | Not serious | 1,362 | RR 3.70 (1.89 to 7.23) | Low |
| Nd:YAG posterior capsulotomy (RR<1 favours multifocal) | | | | | | | | |
| 1 (Law 2020) | RCT | Serious | N/A | Not serious | Very serious | 180 | RR 2.55 (0.68 to 9.55) | Very low |

## (2) EDOF vs Monofocal

**Visual acuity**

| No. of studies | Design | Risk of bias | Inconsistency | Indirectness | Imprecision | No. of participants | Effect size (95% CI) | Quality |
|---|---|---|---|---|---|---|---|---|
| Corrected distance visual acuity worse than 6/6 (RR<1 favours EDOF) | | | | | | | | |
| 2 (Chang 2022; Black 2024) | RCT | Serious | Serious | Not serious | Very serious | 415 | RR 0.80 (0.21, 3.08) | Very low |
| Uncorrected distance visual acuity worse than 6/6 (RR<1 favours EDOF) | | | | | | | | |
| 2 (Chang 2022; Black 2024) | RCT | Serious | Not serious | Not serious | Serious | 355 | RR 1.15 (0.94, 1.40) | Low |
| Uncorrected near visual acuity worse than J3/J4 or equivalent (RR<1 favours EDOF) | | | | | | | | |
| 1 (Chang 2022) | RCT | Serious | N/A | Not serious | Not serious | 295 | RR 0.11 (0.05, 0.25) | Moderate |
| Mean corrected distance visual acuity (logMAR) (lower favours EDOF) | | | | | | | | |
| 5 (Reinhard 2021; Bala 2022; Pantanelli 2023; Corbett 2024; Black 2024) | RCT | Serious | Not serious | Not serious | Not serious | 764 | MD 0.02 (−0.01, 0.05) | Moderate |
| Mean uncorrected distance visual acuity (logMAR) (lower favours EDOF) | | | | | | | | |
| 6 (Reinhard 2021; McCabe 2022; Bala 2022; Pantanelli 2023; Black 2024; Skov 2026) | RCT | Serious | Not serious | Not serious | Not serious | 893 | MD 0.03 (−0.00, 0.07) | Moderate |
| Mean corrected intermediate visual acuity (logMAR) (lower favours EDOF) | | | | | | | | |
| 4 (Reinhard 2021; Bala 2022; Pantanelli 2023; Corbett 2024) | RCT | Serious | Serious | Not serious | Not serious | 644 | MD −0.15 (−0.28, −0.02) | Low |
| Mean uncorrected intermediate visual acuity (logMAR) (lower favours EDOF) | | | | | | | | |
| 5 (Reinhard 2021; McCabe 2022; Bala 2022; Pantanelli 2023; Skov 2026) | RCT | Serious | Serious | Not serious | Serious | 833 | MD −0.12 (−0.21, −0.02) | Very low |
| Mean corrected near visual acuity (logMAR) (lower favours EDOF) | | | | | | | | |
| 4 (Reinhard 2021; Bala 2022; Pantanelli 2023; Corbett 2024) | RCT | Serious | Serious | Not serious | Serious | 644 | MD −0.13 (−0.24, −0.01) | Very low |
| Mean uncorrected near visual acuity (logMAR) (lower favours EDOF) | | | | | | | | |
| 5 (Reinhard 2021; McCabe 2022; Bala 2022; Pantanelli 2023; Skov 2026) | RCT | Serious | Serious | Not serious | Serious | 833 | MD −0.16 (−0.24, −0.08) | Very low |

**Visual function**

| No. of studies | Design | Risk of bias | Inconsistency | Indirectness | Imprecision | No. of participants | Effect size (95% CI) | Quality |
|---|---|---|---|---|---|---|---|---|
| Contrast sensitivity (higher favours EDOF) | | | | | | | | |
| 1 (Skov 2026) | RCT | Serious | N/A | Not serious | Not serious | 92 | MD −0.01 (−0.06, 0.04) | Moderate |
| Visual function questionnaire score (scale direction not reported) (lower favours EDOF) | | | | | | | | |
| 1 (Skov 2026) | RCT | Serious | N/A | Not serious | Serious | 92 | MD −0.16 (−0.41, 0.09) | Low |
| Spectacle dependence (any) (RR<1 favours EDOF) | | | | | | | | |
| 7 (McCabe 2022; Bala 2022; Chang 2022; Pantanelli 2023; Black 2024; Skov 2026; Palkovits 2026) | RCT | Serious | Serious | Not serious | Serious | 833 | RR 0.72 (0.60, 0.88) | Very low |
| Patient satisfaction (RR>1 favours EDOF) | | | | | | | | |
| 2 (Black 2024; Skov 2026) | RCT | Serious | Serious | Not serious | Serious | 152 | RR 1.17 (0.80, 1.73) | Very low |
| Mean stereoacuity (arcsec) (lower favours EDOF) | | | | | | | | |
| 1 (Skov 2026) | RCT | Serious | N/A | Not serious | Very Serious | 92 | MD 29.0 (−15.4, 73.4) | Very Low |
| Light distortion index (lower favours EDOF) | | | | | | | | |
| 1 (Guarro 2022) | RCT | Serious | N/A | Not serious | Serious | 88 | MD 9.33 (3.81, 14.85) | Low |

**Adverse outcomes**

| No. of studies | Design | Risk of bias | Inconsistency | Indirectness | Imprecision | No. of participants | Effect size (95% CI) | Quality |
|---|---|---|---|---|---|---|---|---|
| Glare (RR<1 favours EDOF) | | | | | | | | |
| 8 (McCabe 2022; Bala 2022; Guarro 2022; Chang 2022; Pantanelli 2023; Corbett 2024; Skov 2026; Palkovits 2026) | RCT | Serious | Not serious | Not serious | Serious | 981 | RR 1.00 (0.78, 1.28) | Low |
| Halos (RR<1 favours EDOF) | | | | | | | | |
| 9 (McCabe 2022; Bala 2022; Guarro 2022; Chang 2022; Pantanelli 2023; Corbett 2024; Black 2024; Skov 2026; Palkovits 2026) | RCT | Serious | Serious | Not serious | Not serious | 1,040 | RR 1.51 (1.03, 2.20) | Low |
| Blurred vision (RR<1 favours EDOF) | | | | | | | | |
| 3 (McCabe 2022; Chang 2022; Pantanelli 2023) | RCT | Serious | Not serious | Not serious | Very serious | 565 | RR 0.50 (0.12, 2.06) | Very low |
| Double vision (RR<1 favours EDOF) | | | | | | | | |
| 1 (McCabe 2022) | RCT | Serious | N/A | Not serious | Very serious | 219 | RR 2.13 (0.20, 23.17) | Very low |
| Hazy vision (RR<1 favours EDOF) | | | | | | | | |
| 2 (McCabe 2022; Pantanelli 2023) | RCT | Serious | Not serious | Not serious | Very serious | 270 | RR 0.78 (0.38, 1.60) | Very low |
| Negative dysphotopsia (RR<1 favours EDOF) | | | | | | | | |
| 1 (McCabe 2022) | RCT | Serious | N/A | Not serious | Very serious | 219 | RR 0.68 (0.27, 1.69) | Very low |
| Starbursts (RR<1 favours EDOF) | | | | | | | | |
| 8 (McCabe 2022; Bala 2022; Chang 2022; Pantanelli 2023; Corbett 2024; Black 2024; Skov 2026; Palkovits 2026) | RCT | Serious | Serious | Not serious | Very serious | 1,055 | RR 0.97 (0.59, 1.60) | Very low |

## (3) Monovsion vs Monofocal

**Visual acuity**

| No. of studies | Design | Risk of bias | Inconsistency | Indirectness | Imprecision | No. of participants | Effect size (95% CI) | Quality |
|---|---|---|---|---|---|---|---|---|
| Mean uncorrected distance visual acuity (logMAR) (lower favours monovision) | | | | | | | | |
| 1 (Skov 2026) | RCT | Serious | N/A | Not serious | Serious | 137 | MD 0.08 (0.04 to 0.11) | Low |
| Mean uncorrected intermediate visual acuity (logMAR) (lower favours monovision) | | | | | | | | |
| 1 (Skov 2026) | RCT | Serious | N/A | Not serious | Serious | 137 | MD −0.07 (−0.11 to −0.02) | Low |
| Mean uncorrected near visual acuity (logMAR) (lower favours monovision) | | | | | | | | |
| 1 (Skov 2026) | RCT | Serious | N/A | Not serious | Not serious | 137 | MD −0.22 (−0.27 to −0.16) | Moderate |

**Visual function**

| No. of studies | Design | Risk of bias | Inconsistency | Indirectness | Imprecision | No. of participants | Effect size (95% CI) | Quality |
|---|---|---|---|---|---|---|---|---|
| Contrast sensitivity (Pelli–Robson logCS) (higher favours monovision) | | | | | | | | |
| 1 (Skov 2026) | RCT | Serious | N/A | Not serious | Not serious | 137 | MD −0.01 (−0.06 to 0.04) | Moderate |
| Visual function (Catquest-7SF) (higher favours monovision) | | | | | | | | |
| 1 (Skov 2026) | RCT | Serious | N/A | Not serious | Not serious | 137 | MD −0.01 (−0.26 to 0.23) | Moderate |
| Spectacle dependence (any) (RR<1 favours monovision) | | | | | | | | |
| 1 (Skov 2026) | RCT | Serious | N/A | Not serious | Not serious | 137 | RR 0.72 (0.62 to 0.84) | Moderate |
| Patient satisfaction (RR>1 favours monovision) | | | | | | | | |
| 1 (Skov 2026) | RCT | Serious | N/A | Not serious | Very serious | 137 | RR 1.00 (0.79 to 1.26) | Very low |

**Adverse events**

| No. of studies | Design | Risk of bias | Inconsistency | Indirectness | Imprecision | No. of participants | Effect size (95% CI) | Quality |
|---|---|---|---|---|---|---|---|---|
| Glare (RR<1 favours monovision) | | | | | | | | |
| 1 (Skov 2026) | RCT | Serious | N/A | Not serious | Very serious | 137 | RR 0.98 (0.74 to 1.30) | Very low |
| Halos (RR<1 favours monovision) | | | | | | | | |
| 1 (Skov 2026) | RCT | Serious | N/A | Not serious | Serious | 137 | RR 1.25 (0.78 to 2.01) | Low |
| Mean aniseikonia (lower favours monovision) | | | | | | | | |
| 1 (Skov 2026) | RCT | Serious | N/A | Not serious | Not serious | 137 | MD 0.66 (0.11 to 1.22) | Moderate |
| Starbursts (RR<1 favours monovision) | | | | | | | | |
| 1 (Skov 2026) | RCT | Serious | N/A | Not serious | Very serious | 137 | RR 0.97 (0.64 to 1.49) | Very low |

## (4) Multifocal vs EDOF

**Visual acuity**

| No. of studies | Design | Risk of bias | Inconsistency | Indirectness | Imprecision | No. of participants | Effect size (95% CI) | Quality |
|---|---|---|---|---|---|---|---|---|
| Corrected distance visual acuity worse than 6/6 (RR<1 favours multifocal) | | | | | | | | |
| 1 (Cochener 2018) | RCT | Serious | N/A | Not serious | Serious | 118 | RR 0.90 (0.28, 2.89) | Low |
| Uncorrected distance visual acuity worse than 6/6 (RR<1 favours multifocal) | | | | | | | | |
| 1 (Cochener 2018) | RCT | Serious | N/A | Not serious | Serious | 147 | RR 1.19 (0.76, 1.85) | Low |
| Uncorrected near visual acuity worse than J3/J4 or equivalent (RR<1 favours multifocal) | | | | | | | | |
| 1 (Cochener 2018) | RCT | Serious | N/A | Not serious | Serious | 118 | RR 0.34 (0.15, 0.77) | Low |
| Mean corrected distance visual acuity (logMAR) (lower favours multifocal) | | | | | | | | |
| 2 (Gil 2020; Akkaya Turhan 2021) | RCT | Serious | Not serious | Not serious | Not serious | 145 | MD 0.03 (-0.01, 0.07) | Moderate |
| Mean uncorrected distance visual acuity (logMAR) (lower favours multifocal) | | | | | | | | |
| 3 (Webers 2020; Sevik 2022; Labiris 2024) | RCT | Serious | Not serious | Not serious | Not serious | 116 | MD -0.03 (-0.08, 0.03) | Moderate |
| Mean uncorrected distance visual acuity (decimal) (higher favours multifocal) | | | | | | | | |
| 1 (Cochener 2018) | RCT | Serious | N/A | Not serious | Not serious | 118 | MD 0.02 (-0.05, 0.10) | Moderate |
| Mean corrected intermediate visual acuity (logMAR) (lower favours multifocal) | | | | | | | | |
| 2 (Gil 2020; Akkaya Turhan 2021) | RCT | Serious | Not serious | Not serious | Not serious | 145 | MD 0.12 (0.09, 0.16) | Moderate |
| Mean uncorrected intermediate visual acuity (logMAR) (lower favours multifocal) | | | | | | | | |
| 3 (Webers 2020; Sevik 2022; Labiris 2024) | RCT | Serious | Not serious | Not serious | Not serious | 116 | MD 0.03 (-0.04, 0.10) | Moderate |
| Mean uncorrected intermediate visual acuity (decimal) (higher favours multifocal) | | | | | | | | |
| 1 (Cochener 2018) | RCT | Serious | N/A | Not serious | Not serious | 118 | MD -0.05 (-0.11, 0.02) | Moderate |
| Mean corrected near visual acuity (logMAR) (lower favours multifocal) | | | | | | | | |
| 2 (Gil 2020; Akkaya Turhan 2021) | RCT | Serious | Very serious | Not serious | Very serious | 145 | MD -0.02 (-0.20, 0.16) | Very low |
| Mean uncorrected near visual acuity (logMAR) (lower favours multifocal) | | | | | | | | |
| 3 (Webers 2020; Sevik 2022; Labiris 2024) | RCT | Serious | Very serious | Not serious | Serious | 116 | MD -0.03 (-0.27, 0.20) | Very low |
| Mean uncorrected near visual acuity (decimal) (higher favours multifocal) | | | | | | | | |
| 1 (Cochener 2018) | RCT | Serious | N/A | Not serious | Serious | 118 | MD 0.09 (0.03, 0.16) | Low |

**Visual function**

| No. of studies | Design | Risk of bias | Inconsistency | Indirectness | Imprecision | No. of participants | Effect size (95% CI) | Quality |
|---|---|---|---|---|---|---|---|---|
| Visual function (unspecified instruments) (higher favours multifocal) | | | | | | | | |
| 3 (Webers 2020; Sevik 2022; Labiris 2024) | RCT | Serious | Not serious | Not serious | Very serious | 116 | SMD 0.00 (-0.80, 0.80) | Very low |
| Patient satisfaction (score, 0–100 scale) (higher favours multifocal) | | | | | | | | |
| 1 (Sevik 2022) | RCT | Serious | N/A | Not serious | Very serious | 29 | MD -1.71 (-10.67, 7.25) | Very low |
| Patient satisfaction (binary) (RR>1 favours multifocal) | | | | | | | | |
| 1 (Webers 2020) | RCT | Serious | N/A | Not serious | Serious | 27 | RR 0.99 (0.80, 1.23) | Low |
| Maximum reading speed (words/min) (higher favours multifocal) | | | | | | | | |
| 1 (Akkaya Turhan 2021) | RCT | Serious | N/A | Not serious | Very serious | 29 | MD -5.00 (-26.87, 16.87) | Very low |
| Reading acuity (logMAR) (lower favours multifocal) | | | | | | | | |
| 1 (Akkaya Turhan 2021) | RCT | Serious | N/A | Not serious | Not serious | 29 | MD 0.01 (-0.04, 0.06) | Moderate |
| Mean stereoacuity, arcseconds (lower favours multifocal) | | | | | | | | |
| 1 (Pradeep 2026) | RCT | Serious | N/A | Not serious | Serious | 70 | MD 29.14 (7.27, 51.01) | Low |

**Adverse outcomes**

| No. of studies | Design | Risk of bias | Inconsistency | Indirectness | Imprecision | No. of participants | Effect size (95% CI) | Quality |
|---|---|---|---|---|---|---|---|---|
| Glare (RR<1 favours multifocal) | | | | | | | | |
| 1 (Webers 2020) | RCT | Serious | Not serious | Not serious | Very serious | 27 | RR 1.08 (0.07–15.51) | Very low |
| Halos (RR<1 favours multifocal) | | | | | | | | |
| 1 (Webers 2020) | RCT | Serious | N/A | Not serious | Serious | 27 | RR 1.48 (0.89, 2.46) | Low |
| Overall photic phenomena (RR < 1 favours multifocal) | | | | | | | | |
| 1 (Sevik 2022) | RCT | Serious | N/A | Not serious | Very serious | 29 | RR 4.69 (0.24, 89.88) | Very low |

### (5) Multifocal vs Monovision

**Visual acuity**

| No. of studies | Design | Risk of bias | Inconsistency | Indirectness | Imprecision | No. of participants | Effect size (95% CI) | Quality |
|---|---|---|---|---|---|---|---|---|
| Mean uncorrected distance visual acuity (logMAR) (lower favours multifocal) | | | | | | | | |
| 2 (Wilkins 2013; Labiris 2024) | RCT | Serious | Not serious | Not serious | Not serious | 246 | MD 0.04 (0.01, 0.07) | Moderate |
| Mean uncorrected intermediate visual acuity (logMAR) (lower favours multifocal) | | | | | | | | |
| 2 (Wilkins 2013; Labiris 2024) | RCT | Serious | Not serious | Not serious | Not serious | 241 | MD 0.06 (0.03, 0.09) | Moderate |
| Mean uncorrected near visual acuity (logMAR) (lower favours multifocal) | | | | | | | | |
| 2 (Wilkins 2013; Labiris 2024) | RCT | Serious | Serious | Not serious | Not serious | 246 | MD −0.00 (−0.09, 0.08) | Low |

**Visual function**

| No. of studies | Design | Risk of bias | Inconsistency | Indirectness | Imprecision | No. of participants | Effect size (95% CI) | Quality |
|---|---|---|---|---|---|---|---|---|
| Contrast sensitivity (Pelli-Robson test) (higher favours multifocal) | | | | | | | | |
| 2 (Wilkins 2013; Labiris 2015) | RCT | Serious | Not serious | Not serious | Not serious | 262 | MD −0.03 (−0.10, 0.04) | Moderate |
| Visual function (lower favours multifocal) | | | | | | | | |
| 1 (Labiris 2024) | RCT | Serious | N/A | Serious | Serious | 60 | MD −3.81 (−8.64, 1.02) | Very low |
| Visual function (VF-14) (higher favours multifocal) | | | | | | | | |
| 1 (Labiris 2015) | RCT | Serious | N/A | Not serious | Not serious | 75 | MD −1.47 (−5.51, 2.57) | Moderate |
| Spectacle dependence (any) (RR<1 favours multifocal) | | | | | | | | |
| 2 (Wilkins 2013; Labiris 2015) | RCT | Serious | Not serious | Not serious | Not serious | 262 | RR 0.40 (0.30, 0.53) | Moderate |

**Adverse events**

| No. of studies | Design | Risk of bias | Inconsistency | Indirectness | Imprecision | No. of participants | Effect size (95% CI) | Quality |
|---|---|---|---|---|---|---|---|---|
| Glare (RR<1 favours multifocal) | | | | | | | | |
| 1 (Wilkins 2013) | RCT | Serious | N/A | Not serious | Not serious | 187 | RR 1.41 (1.14, 1.73) | Moderate |

### (6) EDOF vs Monovision

**Visual acuity**

| No. of studies | Design | Risk of bias | Inconsistency | Indirectness | Imprecision | No. of participants | Effect size (95% CI) | Quality |
|---|---|---|---|---|---|---|---|---|
| Corrected distance visual acuity worse than 6/6 (RR<1 favours EDOF) | | | | | | | | |
| 1 (Vandekerckhove 2026) | RCT | Serious | N/A | Not serious | Serious | 105 | RR 5.22 (1.07, 25.55) | Low |
| Uncorrected distance visual acuity worse than 6/6 (RR<1 favours EDOF) | | | | | | | | |
| 1 (Vandekerckhove 2026) | RCT | Serious | N/A | Not serious | Serious | 105 | RR 1.70 (0.92, 3.11) | Low |
| Uncorrected near visual acuity worse than J3/J4 or equivalent (RR<1 favours EDOF) | | | | | | | | |
| 1 (Vandekerckhove 2026) | RCT | Serious | N/A | Not serious | Serious | 105 | RR 0.25 (0.08, 0.77) | Low |
| Mean corrected distance visual acuity (logMAR) (lower favours EDOF) | | | | | | | | |
| 2 (Zeilinger 2025; Vandekerckhove 2026) | RCT | Serious | Not serious | Not serious | Not serious | 185 | MD 0.04 (0.01, 0.07) | Moderate |
| Mean uncorrected distance visual acuity (logMAR) (lower favours EDOF) | | | | | | | | |
| 5 (Sandoval 2020; Labiris 2024; Zeilinger 2025; Vandekerckhove 2026; Skov 2026) | RCT | Serious | Not serious | Not serious | Not serious | 452 | MD 0.03 (−0.02, 0.08) | Moderate |
| Mean corrected intermediate visual acuity (logMAR) (lower favours EDOF) | | | | | | | | |
| 2 (Zeilinger 2025; Vandekerckhove 2026) | RCT | Serious | Not serious | Not serious | Serious | 185 | MD −0.10 (−0.14, −0.07) | Low |
| Mean uncorrected intermediate visual acuity (logMAR) (lower favours EDOF) | | | | | | | | |
| 5 (Sandoval 2020; Labiris 2024; Zeilinger 2025; Vandekerckhove 2026; Skov 2026) | RCT | Serious | Not serious | Not serious | Not serious | 452 | MD −0.01 (−0.06, 0.04) | Moderate |
| Mean corrected near visual acuity (logMAR) (lower favours EDOF) | | | | | | | | |
| 2 (Zeilinger 2025; Vandekerckhove 2026) | RCT | Serious | Serious | Not serious | Serious | 185 | MD −0.09 (−0.22, 0.04) | Very low |
| Mean uncorrected near visual acuity (logMAR) (lower favours EDOF) | | | | | | | | |
| 5 (Sandoval 2020; Labiris 2024; Zeilinger 2025; Vandekerckhove 2026; Skov 2026) | RCT | Serious | Very serious | Not serious | Very serious | 452 | MD 0.01 (−0.17, 0.19) | Very low |
| Mean stereoacuity (arcsec) (lower favours EDOF) | | | | | | | | |
| 1 (Skov 2026) | RCT | Serious | N/A | Not serious | Serious | 131 | MD 10.39 (−34.01, 54.78) | Low |

**Visual function**

| No. of studies | Design | Risk of bias | Inconsistency | Indirectness | Imprecision | No. of participants | Effect size (95% CI) | Quality |
|---|---|---|---|---|---|---|---|---|
| Contrast sensitivity (higher favours EDOF) | | | | | | | | |
| 1 (Skov 2026) | RCT | Serious | N/A | Not serious | Not serious | 131 | MD 0.001 (−0.044, 0.045) | Moderate |
| Visual function (VF-14) (higher favours EDOF) | | | | | | | | |
| 1 (Labiris 2024) | RCT | Serious | N/A | Not serious | Not serious | 60 | MD −5.59 (−10.20, −0.98) | Moderate |
| Spectacle dependence (any) (RR<1 favours EDOF) | | | | | | | | |
| 2 (Vandekerckhove 2026; Skov 2026) | RCT | Serious | Not serious | Not serious | Serious | 312 | RR 1.07 (0.92, 1.24) | Low |
| Patient satisfaction (RR>1 favours EDOF) | | | | | | | | |
| 1 (Skov 2026) | RCT | Serious | N/A | Not serious | Very serious | 131 | RR 0.97 (0.76, 1.25) | Very low |

**Adverse outcomes**

| No. of studies | Design | Risk of bias | Inconsistency | Indirectness | Imprecision | No. of participants | Effect size (95% CI) | Quality |
|---|---|---|---|---|---|---|---|---|
| Glare (RR<1 favours EDOF) | | | | | | | | |
| 1 (Skov 2026) | RCT | Serious | N/A | Not serious | Very serious | 131 | RR 1.08 (0.82, 1.43) | Very low |
| Halos (RR<1 favours EDOF) | | | | | | | | |
| 1 (Skov 2026) | RCT | Serious | N/A | Not serious | Very serious | 131 | RR 0.91 (0.57, 1.44) | Very low |
| Starbursts (RR<1 favours EDOF) | | | | | | | | |
| 1 (Skov 2026) | RCT | Serious | N/A | Not serious | Serious | 131 | RR 1.29 (0.87, 1.90) | Low |

## (7) Trifocal multifocal IOL vs Bifocal multifocal IOL

**Visual acuity**

| No of studies | Design | Risk of bias | Inconsistency | Indirectness | Imprecision | No of participants | Effect size (95% CI) | Quality |
|---|---|---|---|---|---|---|---|---|
| Mean corrected distance visual acuity (lower favours trifocal) | | | | | | | | |
| 3 (Jonker 2015, Gundersen 2016, Li 2023) | RCT | Serious | Not serious | Not serious | Not serious | 149 | MD 0.00 (-0.07, 0.08) | Moderate |
| Mean uncorrected distance visual acuity (logMAR) (lower favours trifocal) | | | | | | | | |
| 3 (Jonker 2015, Gundersen 2016, Li 2023) | RCT | Serious | Not serious | Not serious | Not serious | 149 | MD -0.01 (-0.10, 0.07) | Moderate |
| Mean corrected intermediate visual acuity (lower favours trifocal) | | | | | | | | |
| 2 (Jonker 2015, Li 2023) | RCT | Serious | Not serious | Not serious | Not serious | 127 | MD -0.03 (-0.07, 0.01) | Moderate |
| Mean corrected near visual acuity (lower favours trifocal) | | | | | | | | |
| 2 (Jonker 2015, Li 2023) | RCT | Serious | Not serious | Not serious | Not serious | 127 | MD 0.01 (-0.03, 0.06) | Moderate |
| Mean uncorrected near visual acuity (lower favours trifocal) | | | | | | | | |
| 2 (Jonker 2015, Li 2023) | RCT | Serious | Not serious | Not serious | Not serious | 127 | MD 0.00 (-0.06, 0.07) | Moderate |

**Visual function**

| No of studies | Design | Risk of bias | Inconsistency | Indirectness | Imprecision | No of participants | Effect size (95% CI) | Quality |
|---|---|---|---|---|---|---|---|---|
| Spectacle dependence (near) (RR<1 favours trifocal) | | | | | | | | |
| 1 (Jonker 2015) | RCT | Serious | N/A | Not serious | Very serious | 28 | RR 0.65 (0.18, 2.38) | Very low |

**Supplementary Information 11. Evidence to recommendations (prospective study, human-only group)**

| | |
|---|---|
| **Relative value of different outcomes** | The guideline development group considered distance vision to be an essential safety and functional outcome. However, the value of visual correcting strategies is determined by whether they provide meaningful improvements in unaided intermediate or near vision, daily visual functioning and spectacle independence without imposing unacceptable optical disturbances. Patient satisfaction, health-related quality of life, ability to perform occupational tasks, and the effects of optical disturbances on night-time activities were therefore considered as important as differences in visual-acuity measurements. Clinical measurements should be interpreted in relation to their perceptible effect on everyday life. |
| **Trade-off between benefits and harms** | The committee agreed that no IOL strategy was consistently preferable for all patients. For distant vision, there was evidence that monofocal lenses provided greater distant visual acuity and were associated with fewer photic phenomena, maybe because optical performance is concentrated at a single focal point. However, patients usually required spectacles for near tasks and sometimes for intermediate activities. Other strategies could also provide good distance vision and spectacles might still be required for demanding tasks such as night driving.<br>For intermediate and near vision, multifocal, especially trifocal lenses generally offered the widest functional range and the greatest reduction in spectacle dependence. EDOF lenses offer a range of vision from distance to intermediate, making them suitable for patients who frequently work on computers. However, they did not reliably provide spectacle independence for small print or prolonged near work.<br>Improvements in visual range were accompanied by potential reductions in optical quality. Designs that generated multiple focal ranges could cause halos, glare, starbursts or reduced contrast sensitivity. These effects might be particularly troublesome for people who drive frequently at night or work in low illumination. The evidence did not support the assumption that all EDOF lenses produce fewer visual disturbances than multifocal lenses, performance varied between individual lens platforms.<br>For binocular function, monovision presented a distinct trade-off. Greater anisometropia |
| | could improve near vision but impair binocular comfort, stereopsis or distance visual quality. Previous successful contact-lens monovision might indicate likely tolerance, caution was warranted when precise depth perception was important for patients occupational or recreational requirement.<br>Patient satisfaction depends on whether the postoperative visual profile matched individual priorities. Some patients might accept halos or reduced contrast in exchange for greater spectacle independence and others might prioritize optical quality and reliable night vision.<br>The committee did not interpret non-significant findings in small, heterogeneous studies as evidence of equivalence. Differences in lens generation, refractive targets, follow-up and outcome measurement prevented a consistent hierarchy among strategies. Lens selection should therefore be individualized through shared decision-making, addressing the distances at which unaided vision is most important, night-driving and occupational requirements, willingness to use spectacles and tolerance of unwanted visual phenomena. Patients should understand that no current strategy guarantees spectacle independence, optimal vision at every distance or complete freedom from visual symptoms. |
| **Consideration of health benefits and resource use** | The committee agreed that the economic consequences of IOL selection have changed substantially as procurement systems and the range of available products have developed. Acquisition price, reimbursement by the healthcare system and the amount ultimately paid by an individual patient are different quantities and should not be considered interchangeable.<br>Historically, IOLs with special optical designs were associated with a substantial price premium over standard monofocal lenses. However, that the absolute price difference has narrowed in many settings as more domestic and international products have entered the market and centralized procurement has reduced acquisition prices. In China, the national volume-based procurement programme for intraocular lenses introduced substantial price reductions across the category. The National Healthcare Security Administration reported an average reduction of approximately 60% following the 2023 national centralized procurement round for intraocular lenses and related high-value medical consumables. The additional acquisition cost associated with a multifocal, EDOF or other advanced-function lens may therefore be considerably smaller than pervious. |
| | Pervious studies suggest that presbyopia-correcting lenses may provide additional QALYs and reduce lifetime spectacle expenditure in selected patients. However, these findings are sensitive to assumptions about postoperative spectacle dependence, lens price premiums and willingness-to-pay thresholds, and some studies were manufacturer sponsored. Their applicability to contemporary Chinese practice should therefore be reassessed using current local prices, reimbursement arrangements and patient out-of-pocket costs.<br>The committee also noted that affordability may affect equitable access to advanced optical functions. Although centralized procurement and market competition may reduce this barrier, higher price should not be interpreted as evidence of greater clinical benefit. Local decisions should instead consider whether the additional patient-important benefits justify the incremental cost, with resource-use estimates updated as procurement prices and reimbursement policies change. |
| **Quality of evidence** | All included comparisons were based on randomised trials and risk of bias was consistently judged to be serious. Certainty was frequently reduced further because of inconsistency, imprecision and small sample sizes. Evidence for monovision, EDOF versus monovision and multifocal versus EDOF was generally derived from relatively few studies, and several patient-reported or adverse outcomes were informed by a single trial. These limitations reduce confidence in any conclusion that one presbyopia-correcting strategy is superior overall. |
| **Other considerations** | Lens selection should also take account of ocular comorbidity, binocular function, refractive predictability, previous experience with monovision and the patient’s ability to adapt to optical phenomena. Counselling should make clear that no strategy guarantees complete spectacle independence or freedom from visual symptoms. Expected performance at distance, intermediate and near should be discussed in functional terms relevant to the individual patient. |

**Supplementary Information 12. Recommendations (prospective study, human-only group)**

| **Question: What is the optimal strategy to facilitate simultaneous distance and near vision following cataract surgery?** | |
|---|---|
| **Recommendations in 2017 NICE guideline (NG77: Cataracts in adults:management)** | **Recommendations updated by human experts** |
| Do not offer multifocal intraocular lenses for people having cataract surgery.<br>Offer monovision for use after cataract surgery to people who have either anisometropia or monovision preoperatively and would like to remain with it. | ● Discuss the choice of intraocular lens with the person before cataract surgery, taking account of their priorities for unaided distance, intermediate and near vision, willingness to use spectacles and tolerance of unwanted visual phenomena<br>● Consider a multifocal intraocular lens for people whose ocular health is suitable and who place a high value on unaided intermediate and near vision and reduced spectacle dependence<br>● Explain that no intraocular lens strategy guarantees spectacle independence, optimal vision at all distances or freedom from visual symptoms, and that spectacles may still be required for demanding tasks |

**Supplementary Information 13. External consensus survey (prospective study)**

**Background**

This survey forms part of an expert preference assessment of candidate recommendations on the use of intraocular lenses (IOLs) in people undergoing cataract surgery.

Three candidate recommendations have been developed for the clinical question:

**What is the optimal strategy to facilitate simultaneous distance and near vision following cataract surgery?**

This survey seeks to assess the clinical acceptability and practical applicability of each recommendation among a broader group of ophthalmic experts.

**Instructions for expert reviewers**

Please evaluate each recommendation using the five-point Likert scale below. Your rating should reflect the extent to which you agree that the recommendation is clinically appropriate, evidence-aligned, and acceptable for implementation in current cataract surgery practice.

| **Score** | **Response option** |
|---|---|
| 1 | Strongly disagree |
| 2 | Disagree |
| 3 | Neutral |
| 4 | Agree |
| 5 | Strongly agree |

**Voting Questions (presented in randomized order)**

**Question 1: Recommendation A**

Please indicate your level of agreement with the following recommendation:

- *Do not offer multifocal intraocular lenses for people having cataract surgery.*
- *Offer monovision for use after cataract surgery to people who have either anisometropia or monovision preoperatively and would like to remain with it.*

Rating:

☐ 1 Strongly disagree

☐ 2 Disagree

☐ 3 Neutral

☐ 4 Agree

☐ 5 Strongly agree

**Question 2: Recommendation B**

Please indicate your level of agreement with the following recommendation:

*Consider multifocal intraocular lenses for people who want greater spectacle independence for distance, intermediate and near vision. When discussing this option:*

- *ensure the person's ocular health is suitable for multifocal intraocular lenses*
- *explain the potential benefits in uncorrected near and intermediate vision compared with standard monofocal intraocular lenses*
- *discuss the increased risk of visual disturbances, such as halos, glare and starbursts*
- *take into account the person's visual needs and daily activities*
- *consider cost implications and the person's willingness to pay.*

Rating:

☐ 1 Strongly disagree

☐ 2 Disagree

☐ 3 Neutral

☐ 4 Agree

☐ 5 Strongly agree

**Question 3: Recommendation C**

Please indicate your level of agreement with the following recommendation:

- *Discuss the choice of intraocular lens with the person before cataract surgery, taking account of their priorities for unaided distance, intermediate and near vision, willingness to use spectacles and tolerance of unwanted visual phenomena*
- *Consider a multifocal intraocular lens for people whose ocular health is suitable and who place a high value on unaided intermediate and near vision and reduced spectacle dependence*
- *Explain that no intraocular lens strategy guarantees spectacle independence, optimal vision at all distances or freedom from visual symptoms, and that spectacles may still be required for demanding tasks*

Rating:

☐ 1 Strongly disagree

☐ 2 Disagree

☐ 3 Neutral

☐ 4 Agree

☐ 5 Strongly agree

**Consensus Interpretation**

Responses will be analyzed descriptively. For each recommendation, the proportion of experts selecting “agree” or “strongly agree” will be calculated. The results will be used to assess whether the revised recommendation is aligned with current clinical evidence and acceptable for implementation in real-world ophthalmic surgical practice.